\documentclass{article}
\usepackage{ijcai26}

\usepackage{times}
\usepackage{soul}
\usepackage{url}
\usepackage[hidelinks]{hyperref}
\usepackage[utf8]{inputenc}
\usepackage[small]{caption}
\usepackage{graphicx}
\usepackage{amsmath}
\usepackage{amsthm}
\usepackage{booktabs}
\usepackage{algorithmic}
\usepackage{multirow}
\usepackage{amssymb}
\usepackage{booktabs}
\usepackage{subfigure}

\usepackage{stmaryrd} 

\usepackage[ruled,vlined,linesnumbered]{algorithm2e}

\usepackage{enumitem}

\usepackage{pifont}
\newcommand{\cmark}{\ding{51}}%
\newcommand{\xmark}{\ding{55}}%

\usepackage{xcolor}
\usepackage{graphicx}

\usepackage{todonotes}
\usepackage{comment}
 
\usepackage{color}

\definecolor{Orange}{rgb}{1,0.5,0}
\definecolor{Red}{rgb}{1,0,0}
\definecolor{Blue}{rgb}{0,0,1}
\definecolor{Green}{rgb}{0,1,0}
\definecolor{LightGreen}{rgb}{0.2,0.8,0.2}

\title{D2ACE: Multi-Label Batch Selection Guided by Dual Dynamics and Adaptive Correlation Enhancement}

\author{
Bin Liu$^1$\thanks{Corresponding author.}
\and
Haoyu Peng$^{1,2}$,
Zhijia Wei$^{1,2}$,
Jiajing Zhang$^1$,
Grigorios Tsoumakas$^3$\\
\affiliations
$^1$ Key Laboratory of DECV, Chongqing University of Posts and Telecommunications, China\\
$^2$ School of Computer Science and Technology, Chongqing University of Posts and Telecommunications\\
$^3$ School of Informatics, Aristotle University of Thessaloniki, Greece \\ 
\emails
liubin@cqupt.edu.cn,
\{s240231185,s250201123,s250331042\}@stu.cqupt.edu.cn,
greg@csd.auth.gr
}

\begin{document}

\maketitle

\begin{abstract}

Batch selection is crucial for improving both training efficiency and predictive performance in deep multi-label classification (MLC). Existing batch selection methods typically rely on a single metric to assess instance importance and use static label weights to distinguish label significance, neglecting the dynamic evolution of metric utility and label significance during training. In addition, the method that explicitly exploits label correlations is largely affected by abundant irrelevant labels and insensitive to local label distributions. To address these issues, we propose D2ACE, a novel multi-label batch selection method guided by Dual Dynamics and Adaptive Correlation Enhancement. D2ACE explicitly captures metric and label-level training dynamics by combining stage-wise Bernoulli mixture sampling, which balances uncertainty and noise-resistant hardness, with dynamic label weighting to recalibrate label priorities at each epoch based on current metric statistics. Furthermore, D2ACE introduces a local context-aware correlation enhancement to focus on relevant labels with instance-adaptive dependencies. Extensive experiments on tabular and image benchmarks demonstrate that D2ACE outperforms existing batch selection approaches across various deep MLC models, achieving stronger predictive performance and more efficient correlation modeling.



\end{abstract}

\section{Introduction}

By modeling multiple outputs simultaneously, multi-label classification (MLC) methods are able to capture complex real-world phenomena where multiple concepts often co-occur and label dependencies are intrinsic~\cite{shou2023concurrent}. This flexibility has led to the extensive application of MLC methods in domains such as text classification~\cite{chai2024compositional,lin2023effective}, image annotation~\cite{guo2023texts}, and protein function prediction \cite{ML_protein_task}.


Recently, deep learning architectures for MLC have achieved significant progress. Deep MLC models automatically extract semantically rich features from input data and predict multiple labels for each instance simultaneously~\cite{liu2021emerging}. Using techniques such as prompt-tuning~\cite{guo2023texts}, attention mechanisms~\cite{C-Tran}, and graph neural networks~\cite{CLIF}, these models capture complex feature–label and label–label dependencies, improving both accuracy and generalization.

\begin{figure}[!t]
  \centering
  \subfigure[Hard-Imb (epoch 20)]{
    \includegraphics[width=0.46\columnwidth]{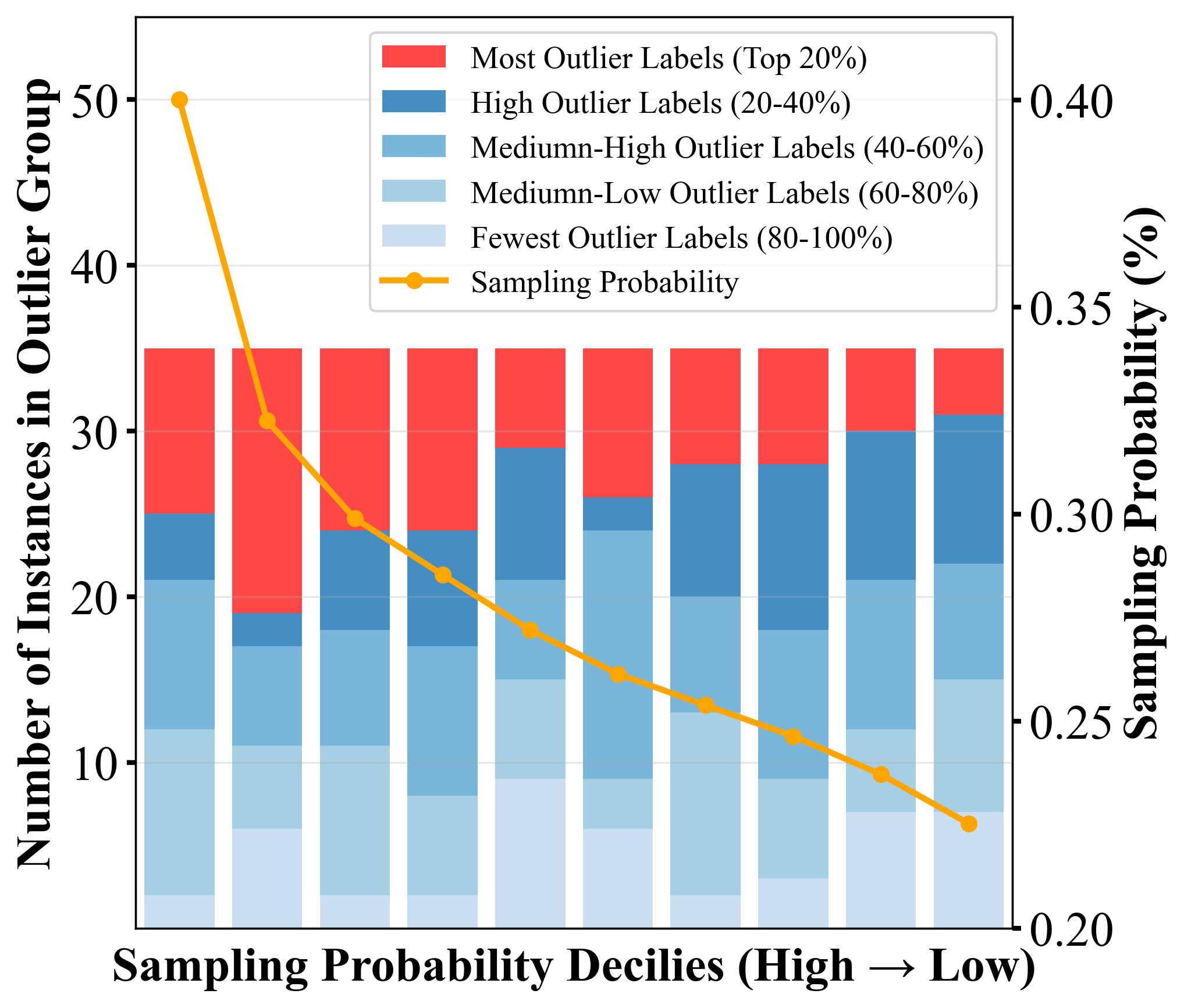}
  }
  \hspace{0.02\columnwidth}
  \subfigure[ML-Unc (epoch 80)]{
    \includegraphics[width=0.46\columnwidth]{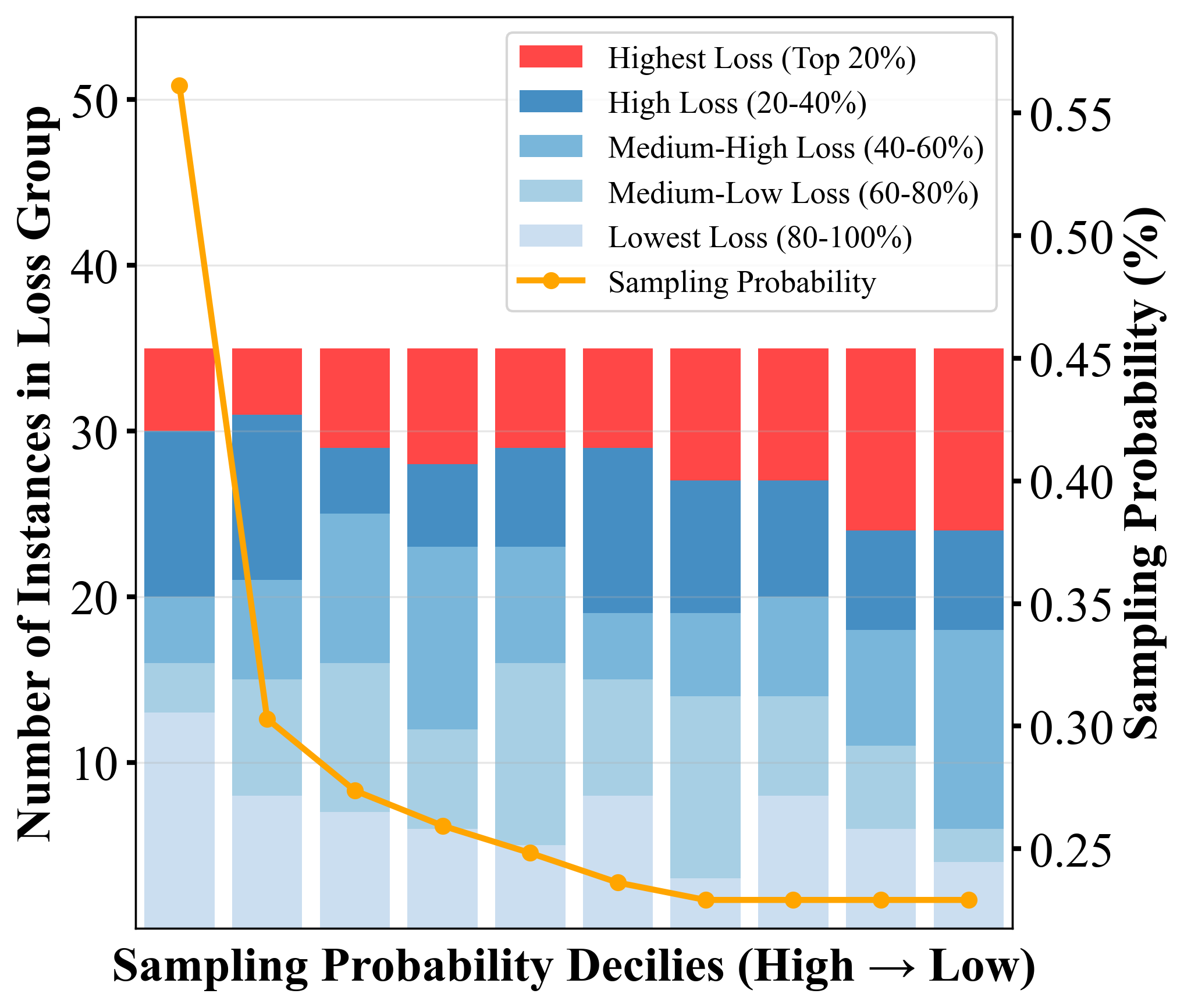}
  }
  \caption{Distribution of instance properties across sampling probability deciles for Hard-Imb and ML-Unc using CLIF base model on the CAL500 dataset. The x-axis shows deciles of instances sorted by descending sampling probability, and color bars indicate counts from each outlier/loss group within each decile. 
  }
  \label{fig:early_epoch_outlier}
\end{figure}

Batch selection plays a key role in improving both training efficiency and prediction accuracy in deep learning. Instance importance is typically measured using hardness and uncertainty. Hardness-based methods assess the learning difficulty of instances, guiding the model to focus on easy, hard, or near-boundary samples \cite{online2015,DIHCL_NIPS20,Ada-boundary2020}, whereas uncertainty-based methods prioritize instances with unstable or ambiguous predictions \cite{active2017,recent2020}. While existing multi-label batch selection methods rely on either hardness \cite{hard_imbal_2024ECML} or uncertainty \cite{MLuncertain2025},  both metrics have clear limitations when applied in isolation (see Figure \ref{fig:early_epoch_outlier}). 
Hardness-based selection (Hard-Imb) can overemphasize noisy or outlier samples during early training, increasing the risk of overfitting.
In contrast, uncertainty-based selection (ML-Unc) ignores genuinely hard (high-loss) instances that exhibit low uncertainty but remain misclassified in later training stages. Current multi-label batch selection methods lack a systematic approach to harmonize the complementary strengths of these two metrics.

\begin{figure}[!t]
  \centering
  \subfigure[Balance]{
    \includegraphics[width=0.46\columnwidth]{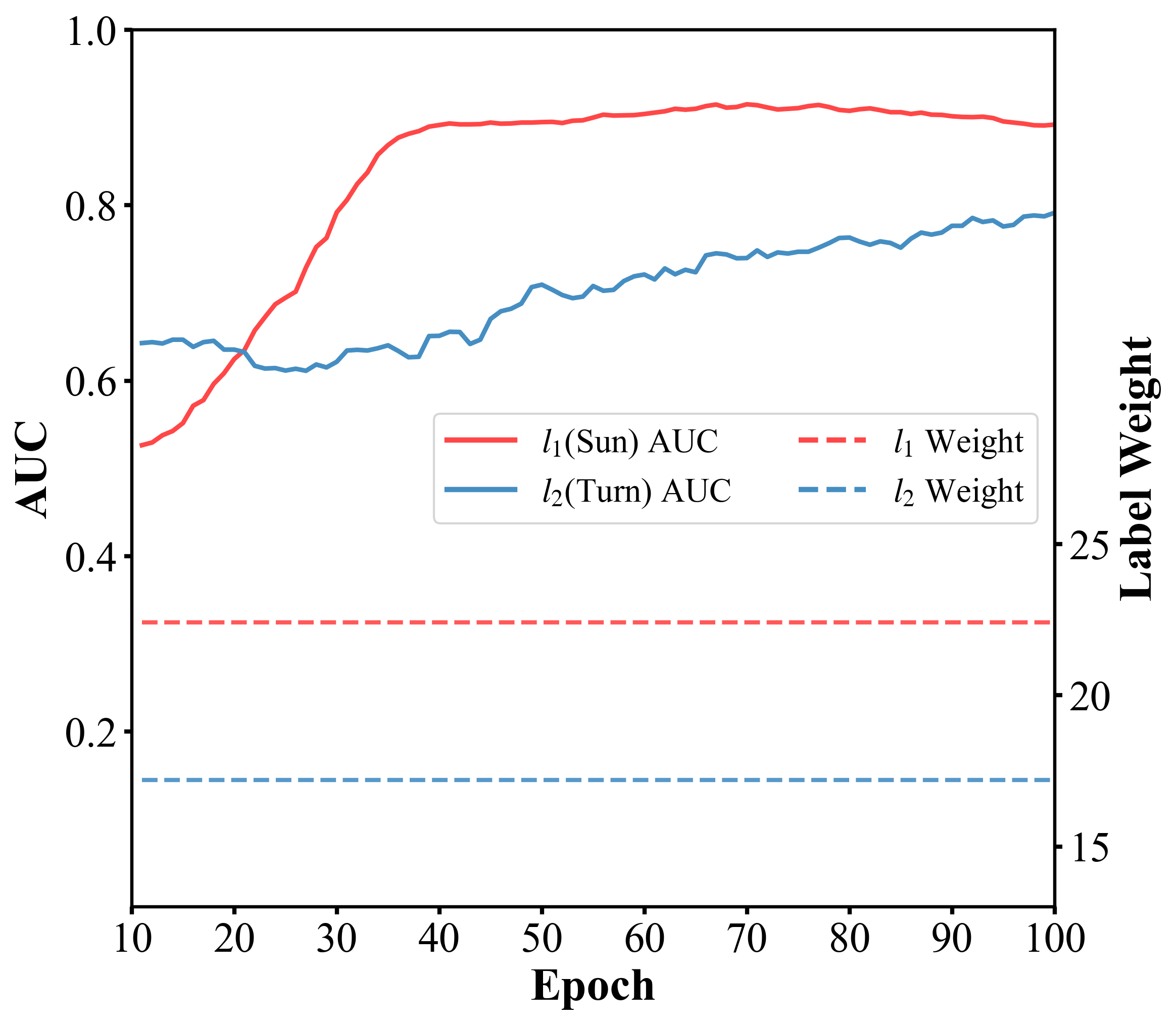}
  }
  \hspace{0.02\columnwidth}
  \subfigure[Hard-Imb]{
    \includegraphics[width=0.46\columnwidth]{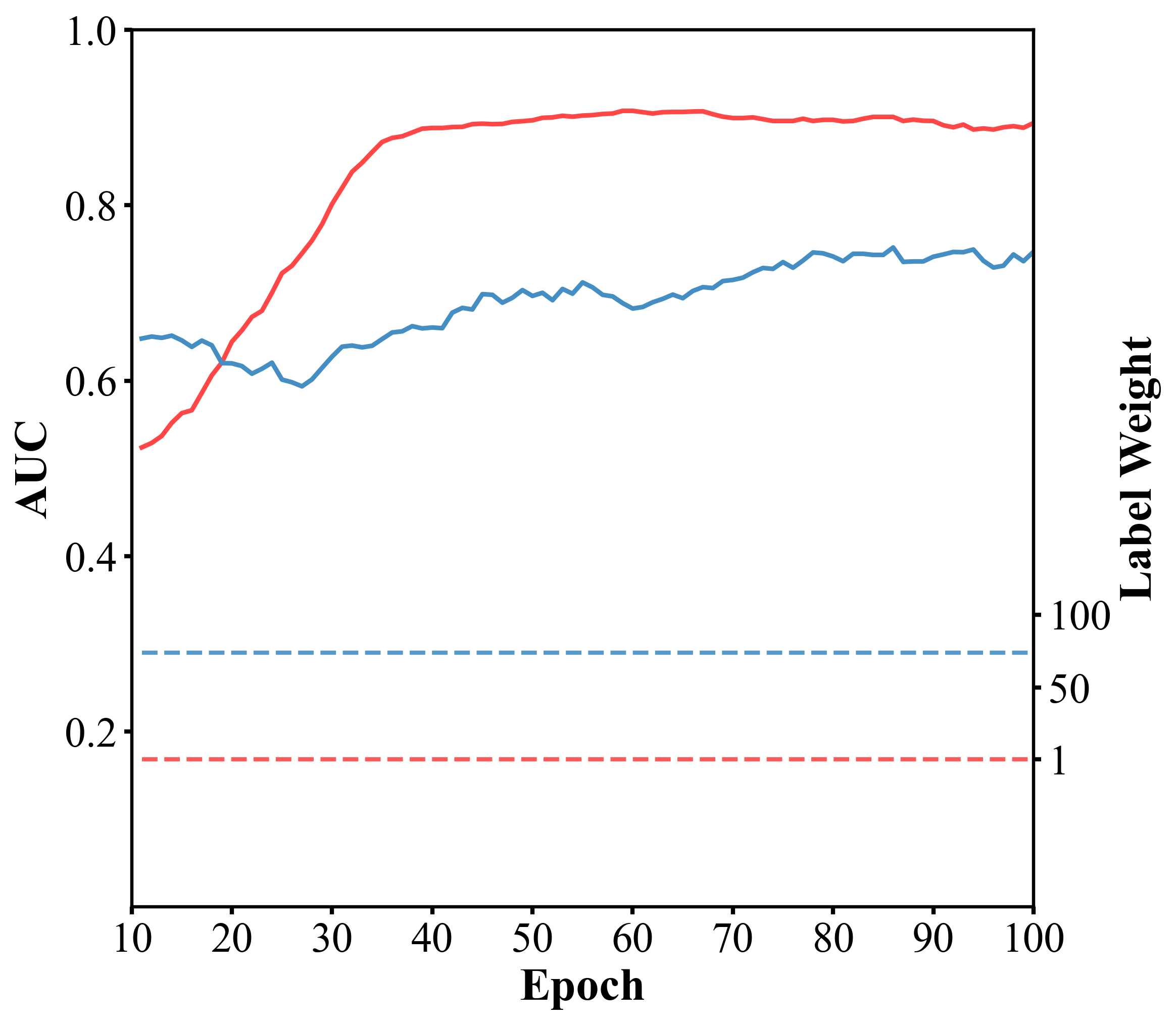}
  }
  \caption{AUC of two labels from the Corel5k dataset and their corresponding label weights given by Balance and Hard-Imb using CLIF base model.}
  \label{fig:label_weight}
\end{figure}

In MLC, different labels can exhibit distinct and evolving importance throughout training, as evidenced by their AUC trajectories. 
As shown in Figure \ref{fig:label_weight}, although label $l_1$ has a lower AUC than $l_2$ in the early stages, its AUC increases more rapidly as training progresses. However, existing multi-label batch selection methods, such as Balance \cite{mllbalance} and Hard-Imb \cite{hard_imbal_2024ECML}, rely on fixed label priorities and predefined static weights, failing to adapt to the evolving label learning dynamics.


Moreover, label correlations play a crucial role in capturing complex output structures and improving performance in multi-label models. ML-Unc \cite{MLuncertain2025} pioneers the incorporation of label correlations into batch selection by prioritizing instances that exhibit synergistic uncertain behavior across multiple labels.
However, it propagates correlation-enhanced uncertainty uniformly over all labels, implicitly assuming equal contributions from relevant and irrelevant labels as well as the same label dependencies across instances.
The inherent sparsity of multi-label datasets, where each instance is associated with only a few relevant labels~\cite{MCE_imbalancedML,oversampling_nn,AEMLO}, causes ML-Unc to disproportionately benefit irrelevant labels over more informative relevant ones.
Furthermore, the globally estimated correlation-based metric score overlooks local variations in co-occurrence, as globally correlated labels may not co-exist in every local region.

To address the above issues, we propose D2ACE, a novel multi-label batch selection strategy that considers training dynamics from both metric and label perspectives while effectively leveraging label correlations.
First, D2ACE jointly integrates uncertainty and noise-resistant-hardness through a stage-wise Bernoulli mixture sampling scheme, which dynamically shifts the sampling focus from uncertain to hard instances as training progresses. This stage-adaptive design exploits the complementary strengths of the two metrics, enabling the selection of informative instances at different learning stages.
Second, D2ACE introduces a dynamic label weighting scheme that recalibrates label priority at each epoch based on current metric statistics to capture the evolving importance of labels during training.
Third, D2ACE develops a local context-aware label correlation enhancement that explicitly focuses on relevant labels to prevent the dominance of irrelevant ones and incorporates local label context to exploit instance-adaptive dependencies, thereby adaptively amplifying informative metric scores and improving the efficiency of correlation mining. 

Our main contributions are summarized as follows:
\begin{itemize}
    \item \textbf{Training Dynamics}: We propose a unified multi-label batch selection framework D2ACE that concurrently addresses metric dynamics by adaptively balancing uncertainty and hardness through stage-wise Bernoulli mixture sampling while tackling label dynamics via epoch-wise recalibration of label priorities.
    \item \textbf{Correlation Mining}: We develop a novel label correlation enhancement strategy that estimates relevance-filtered label correlations and propagates correlations based on local label context, facilitating efficient metric enhancement in an instance-adaptive manner.  
    \item \textbf{Empirical Validation}: Our method achieves the largest performance improvement compared with state-of-the-art single- and multi-label batch selection methods, demonstrating consistent superiority across diverse deep multi-label models on tabular and image datasets. 
\end{itemize}

\begin{table*}[!h]
\label{table:batchselectmethod}
\centering
\small
\setlength{\tabcolsep}{4pt} 
\begin{tabular}{ccccccc}
\toprule
\multicolumn{2}{c}{Method} & Metric & 
Metric dynamics & Label dynamics &  Label correlation & Local context \\
\midrule
\multirow{3}{*}{\begin{tabular}{@{}c@{}}single\\-label\end{tabular}} & 
Active \cite{active2017} & uncertainty & - & - & - & -  \\
& Recent \cite{recent2020} & uncertainty & - & - & - & -  \\
& DIHCL \cite{DIHCL_NIPS20} & hardness & - & - & - & -  \\
\midrule
\multirow{4}{*}{\begin{tabular}{@{}c@{}}multi\\-label\end{tabular}} & 
Balance \cite{mllbalance} & imbalance & - & \xmark & \xmark & \xmark  \\
& Hard-Imb \cite{hard_imbal_2024ECML} & hardness \& imbalance & \xmark & \xmark & \xmark & \cmark  \\
& ML-Unc \cite{MLuncertain2025} & uncertainty & - & \xmark & \cmark & \xmark \\
& \textbf{D2ACE (Ours)}  & hardness \& uncertainty & \cmark & \cmark & \cmark & \cmark \\
\bottomrule
\end{tabular}
\caption{The summary of single-label and multi-label batch selection approaches.}
\label{ta:methods_summary}
\end{table*}


\section{Related Work}
\subsection{Multi-Label Classification}


Traditional methods for classifying multi-label tabular data fall into two main categories: algorithm adaptation, which extends conventional machine learning models to directly handle multi-label outputs \cite{MLkNN,ML_deep_fortest}, and problem transformation, which converts the MLC task into multiple single-label classification problems via various label space transformations \cite{new_BR,RAkEL,New_CC}.
Recently, deep learning models have advanced MLC by effectively capturing feature–label and label–label dependencies~\cite{deepsurvey2024}.
CLIF~\cite{CLIF} integrates a graph autoencoder to encode label semantics to guide informative label-specific feature extraction. 
HOT-VAE \cite{HotVae} employs attention mechanisms to adaptively exploit high-order label dependencies.
DELA~\cite{DELA} identifies label-specific non-informative features and learns robust classifiers through stochastic feature perturbation.
PACA~\cite{PACA} formulates an end-to-end probabilistic framework based on
prototype-based latent metric spaces with label correlation regularization.
FLEM~\cite{FLEM} integrates label enhancement into model training to jointly optimize label importance recovery and predictive learning.


Multi-label image classification methods typically employ pre-trained ResNet \cite{ResNet} backbones to extract visual features from raw images, upon which multi-label prediction is performed with label dependency modeling.
ML-GCN~\cite{MLGCN} captures global label correlations by generating inter-dependent classifiers from semantic label representations using graph convolutional networks, enabling shared parameter learning across labels.
\textit{C-Tran} \cite{C-Tran} is a transformer-based method that models label dependencies via self-attention and reconstruction of masked label embeddings conditioned on visual features, without relying on predefined label graphs.
HST~\cite{HST} exploits both intra-image co-occurrence and cross-image semantic similarity through heterogeneous semantic transfer.
TAI++~\cite{TAIpp} leverages pseudo-visual prompts and a dual-adapter co-learning strategy to transfer visual knowledge from text to image representations.



\subsection{Batch Selection}

Single-label batch selection methods typically prioritize instances based on either uncertainty or learning difficulty.
To measure uncertainty, Active~\cite{active2017} estimates prediction variance over the training process, while Recent~\cite{recent2020} focuses on the entropy of recent predictions.
In contrast, DIHCL~\cite{DIHCL_NIPS20} characterizes sample hardness by tracking loss variation and prediction flips between consecutive epochs, and employs an exponential moving average to progressively reduce the influence of earlier training stages.

Multi-label batch selection methods extend these metrics by operating on sample–label pairs and explicitly accounting for label-specific characteristics. The Balance method~\cite{mllbalance} constructs mini-batches to match desired label distributions based on global label imbalance. 
Hard-Imb integrates static local label imbalance with label-wise loss values to identify hard instances.
ML-Unc~\cite{MLuncertain2025} proposes an uncertainty metric that combines current prediction reliability with fine-grained changes in recent outputs, and further exploits global label correlations to emphasize instances exhibiting synergistic uncertainty across more labels. 

As summarized in Table \ref{ta:methods_summary}, D2ACE is distinguished from existing batch selection methods by the joint consideration of hardness and uncertainty in a stage-aware manner, dynamic label weighting, and local context-aware label correlation enhancement.

\section{Method}

\subsection{Preliminary}
Let $\mathcal{D} = \{(\mathbf{x}_i, \mathbf{y}_i)\}_{i=1}^{n}$ be a multi-label dataset consisting of $n$ instances, where the $i$-th instance consists of a feature vector $\mathbf{x}_i \in \mathbb{R}^d$ and a corresponding binary label vector $\mathbf{y}_i = [y_{i1}, y_{i2}, \ldots, y_{iq}] \in \{0,1\}^q.$ 
Given a label set $\mathcal{L} = \{l_1, l_2, \ldots, l_q\}$, $y_{ij}=1$ indicates that $\mathbf{x}_i$ is relevant to label $l_j$, and $y_{ij}=0$ otherwise. 
The objective of multi-label classification is to learn a mapping function $\mathbb{R}^d \rightarrow \{0,1\}^q$ that predicts the label subset associated with unseen instances. 
To accommodate memory and computational limitations during training, deep multi-label model parameters are updated iteratively using mini-batches $\mathcal{B} = \{(\mathbf{x}_{k}, \mathbf{y}_{k})\}_{k=1}^{b} \subseteq \mathcal{D}.$




\subsection{Importance Metrics}


Focusing exclusively on either uncertain or hard samples can be suboptimal, as it may overlook persistently misclassified instances in later training stages or lead to overfitting noisy outliers in early epochs. To address this issue, D2ACE employs the uncertainty metric from \cite{MLuncertain2025} and further develops a hardness metric that integrates prediction stability with loss to mitigate the effect of noise.

\subsubsection{Uncertainty Metric}

We adopt the uncertainty metric of~\cite{MLuncertain2025} that jointly accounts for both the ambiguity of the current model prediction and the temporal instability of recent predictions.
Let $\hat{y}_{ij}^t \in [0,1]$ denote the predicted probability that instance $\mathbf{x}_i$ is associated with label $l_j$ at epoch $t$. The uncertainty of this prediction is assessed using binary entropy:
\begin{equation}
e_{ij}^t = -\left(\hat{y}_{ij}^t \log_2\hat{y}_{ij}^t + (1-\hat{y}_{ij}^t)\log_2(1-\hat{y}_{ij}^t)\right),
\label{eq:current_uncertain}
\end{equation}
which attains its maximum when $\hat{y}_{ij}^t = 0.5$.
To capture the temporal variation of model predictions, the metric measures their fluctuation over $n_t$ most recent epochs by averaging the absolute differences of predicted probabilities at adjacent epochs:
\begin{equation}
d_{ij}^{\,t} = \frac{1}{n_t-1}\sum_{k=1}^{n_t-1} \left| \hat{y}_{ij}^{\,t-k+1} - \hat{y}_{ij}^{\,t-k} \right|,
\label{eq:historical_uncertain}
\end{equation}
where larger values reflect higher fluctuations. 

The overall uncertainty of instance $\mathbf{x}_i$ with respect to label $l_j$ at epoch $t$ is obtained by aggregating Eq.\eqref{eq:current_uncertain} and Eq.\eqref{eq:historical_uncertain}: 
\begin{equation}
u_{ij}^{\,t} = \lambda_1 e_{ij}^t + (1-\lambda_1) d_{ij}^{\,t},
\label{eq:u_metric}
\end{equation}
where $\lambda_1 \in [0,1]$ balances the contribution of current prediction uncertainty and temporal instability. 
Finally, the uncertainty scores ${u_{ij}^{\,t}}$ for all training instances and labels collectively form the uncertainty matrix $\mathbf{U}^t \in \mathbb{R}^{n \times q}$.

\subsubsection{Hardness Metric}

The instantaneous training loss (e.g., binary cross-entropy) is a commonly used indicator for assessing instance difficulty~\cite{DIHCL_NIPS20,hard_imbal_2024ECML}. 
However, relying solely on model loss incurs the risk of overfitting to noisy instances. Prior studies found that predictions for noisy (outlier) samples tend to oscillate frequently during training, whereas genuinely hard examples are more likely to be misclassified consistently over long periods~\cite{toneva2018,maini2022characterizing}. Motivated by this observation, we disentangle hard samples from noisy ones by incorporating the prediction flip into the hardness metric.

Let $\tilde{y}_{ij}^{\,t} \in \{0,1\}$ be the binary relevance prediction of instance $\mathbf{x}_i$ for label $l_j$ at epoch $t$. The prediction flip indicator between two consecutive epochs is defined as $f_{ij}^{\,t} = \left| \tilde{y}_{ij}^{\,t} - \tilde{y}_{ij}^{\,t-1} \right|$, which captures whether the predicted relevance state of $l_j$ for $\mathbf{x}_i$ changes across two successive epochs. 
To accumulate flip information and capture long-term prediction stability, we apply an exponential moving average (EMA)~\cite{DIHCL_NIPS20}:
\begin{equation}
\bar{f}_{ij}^{\,t} = \lambda_2 f_{ij}^{\,t} + (1-\lambda_2)\bar{f}_{ij}^{\,t-1},
\label{eq:ema_flip}
\end{equation}
where $\lambda_2 \in (0,1]$ is the smoothing parameter. Larger value of $\bar{f}_{ij}^{\,t}$ indicates higher long-term prediction instability, suggesting a higher likelihood that the corresponding label $y_{ij}$ is noisy rather than intrinsically difficult.
Then, we define the hardness metric by integrating the current loss and prediction stability:
\begin{equation}
h_{ij}^{\,t} = \ell_{ij}^t\big(1 - \bar{f}_{ij}^{\,t}\big),
\label{eq:hard}
\end{equation}
where $\ell_{ij}^t$ denotes the loss of instance $\mathbf{x}_i$ with respect to label $l_j$ at epoch $t$.
This metric prioritizes sample-label pairs that consistently exhibit high loss with stable incorrect predictions, while suppressing those dominated by frequent prediction flips that are more likely caused by label noise. 
The hardness scores ${h_{ij}^{\,t}}$ over all instances and labels constitute the hardness matrix $\mathbf{H}^t \in \mathbb{R}^{n \times q}$.

\subsection{Dynamic Label Weight}

To adaptively capture the evolving importance (e.g., uncertainty and hardness) of various labels during training, D2ACE introduces  \textit{dynamic label weights} that assess the priority of each label across epochs.
Given a metric matrix $\mathbf{A} \in \{\mathbf{U}, \mathbf{H}\}$ representing either uncertainty or hardness at the current epoch\footnote{For simplicity, when referring to variables at the current epoch $t$, the superscript is omitted if there is no ambiguity.}, we first compute label-wise (column-wise) means and standard deviations:
\begin{align}
\boldsymbol{\mu}_A = \frac{1}{n}\mathbf{1}_n^\top \mathbf{A}, \quad
\boldsymbol{\sigma}_A = \sqrt{\frac{1}{n} \mathbf{1}_n^\top (\mathbf{A} \odot \mathbf{A}) - \boldsymbol{\mu}_A \odot \boldsymbol{\mu}_A},
\label{eq:label_weight1}
\end{align}
where $\mathbf{1}_n$ is a row vector of ones and $\odot$ denotes element-wise multiplication. 
Labels with higher mean values consistently exhibit stronger metric values across many instances, indicating overall importance for training, while labels with higher variance highlight the presence of instances with particularly large values that should be prioritized.
We then combine these two factors using an element-wise exponential transformation to obtain label weights:
\begin{align}
\mathbf{v}_A = \exp\!\Big(\tfrac{1}{2}(\boldsymbol{\mu}_A + \boldsymbol{\sigma}_A)\Big),
\label{eq:label_weight2}
\end{align}
and scale the metric matrix column-wise to highlight important labels:
\begin{align}
\mathbf{Q}_A = \mathbf{A} \, \mathrm{diag}(\mathbf{v}_A).
\label{eq:label_weight3}
\end{align}
Finally, instance weights that reflect dynamic label-specific importance are obtained via row-wise aggregating the weighted matrix, i.e., $\boldsymbol{\delta}_A = \mathbf{Q}_A \mathbf{1}_q^\top \in \mathbb{R}^{n}$.

\subsection{Local Context-Aware Label Correlation Enhancement} 
\label{sec:local-corr-prop}

To exploit label dependencies, we propose a  \textit{local context-aware label correlation enhancement} strategy that selectively amplifies informative metric scores in an instance-adaptive manner.

Let $\mathbf{Y} = [\mathbf{y}_1^\top, \dots, \mathbf{y}_n^\top]^\top \in \{0,1\}^{n \times q}$ be the ground-truth label matrix of the training set.
To prevent irrelevant labels from dominating correlation construction, we first apply relevance filtering by constructing a label-masked metric matrix
$\mathbf{M}_A = \mathbf{Y} \odot \mathbf{A}$.
This operation explicitly focuses on correlations among relevant labels (i.e., $y_{ij}=1$), thereby reducing the interference from abundant irrelevant (negative) labels and improving the ability to capture informative dependencies for positive label prediction~\cite{ICME2024positive}.
Based on the masked metric, we estimate global label correlations using cosine similarity:
\begin{equation}
    \mathbf{C}_A = \mathbf{\bar{M}}_A^\top \mathbf{\bar{M}}_A, 
\label{eq:C_A}
\end{equation}
where $\mathbf{\bar{M}}_A = \mathbf{M}_A \text{diag}(||[\mathbf{M}_A]_{:,1}||_2, \cdots, ||[\mathbf{M}_A]_{:,q}||_2)$ and $[\mathbf{M}_A]_{:,j}$ is the $j$-th column of $\mathbf{M}_A$. A larger value of $[C_A]_{ij}$ indicates stronger synchronous patterns of metric values between labels $l_i$ and $l_j$ at the dataset level.

To further incorporate \emph{local label context} into global correlation modeling, we introduce a local label appearance matrix $\mathbf{Z} \in \{0,1\}^{n \times q}$:
\begin{equation}
Z_{ij} = \mathbb{I} \Big( \sum_{\mathbf{x}_m \in \mathcal{N}(\mathbf{x}_i)}y_{mj} \ge 1 \Big),
\end{equation}
where $\mathbb{I} \left(\cdot\right)$ is the indicator function, and $\mathcal{N}(\mathbf{x}_i)$ identifies the local region of $\mathbf{x}_i$, which can be defined as $\mathbf{x}_i$ itself or its $K$-nearest neighbors. An entry $Z_{ij}=1$ indicates at least one instance in $\mathcal{N}(\mathbf{x}_i)$ associating with label $l_j$. 
We then integrate global dependency and local context to perform instance-adaptive correlation enhancement:
\begin{equation}
\mathbf{R}_A = \mathbf{M}_A \odot \big( \mathbf{Z}_A \mathbf{C}_A \big),
\label{eq:Q_A}
\end{equation}
where only metric scores associated with relevant labels are amplified, and the degree of increase is determined by both global label correlations and their local presence around each instance.
Our correlation enhancement focuses correlation estimation on relevant labels and explicitly conditions correlation propagation on local label context, addressing irrelevant label correlation dominance and neglect of instance-specific information in ML-Unc (see Appendix A.1 for a visual example).
Finally, we obtain instance-level weights that capture local context-aware label correlations via row-wise summarization, i.e., $\boldsymbol{\phi}_A = \mathbf{R}_A \mathbf{1}_q^\top \in \mathbb{R}^{n}$. 

\textit{Remarks}: ML-Unc can be regarded as a degenerate variant of our label correlation enhancement, which does not consider label relevance filtering and defines the local region as the entire dataset, i.e., $\mathcal{N}(\mathbf{x}_i) = \mathcal{D}$ (see Appendix A.2 for details).

\subsection{Sampling Probability}






At each epoch $t$, we first compute the \emph{final instance weight} by integrating label dynamic importance and correlation-enhanced information:
\begin{equation}
\mathbf{w}_A = \boldsymbol{\delta}_A + \boldsymbol{\phi}_A,
\end{equation}
where $\boldsymbol{\delta}_A$ captures label dynamics and $\boldsymbol{\phi}_A$ encodes label correlation-aware contributions.

To convert instance weights into sampling probabilities, we adopt the quantization index strategy \cite{recent2020,hard_imbal_2024ECML}.
Following the exponential decay strategy that captures the weight changes across epochs, the weight $[w_A]_i \in [0,1]$ of $\mathbf{x}_i$ is first mapped to a quantization index:
\begin{equation}
Q([w_A]_i) = \left\lceil 1 - \frac{[w_A]_i}{\Delta} \right\rceil, 
\end{equation}
where $\Delta=\frac{1}{n}$ is the quantization step size. 
The sampling probability of $\mathbf{x}_i$ at the current epoch $t$ is defined as:
\begin{equation}
P(\mathbf{x}_i|\mathbf{w}_A,t) = \frac{1/\exp(\log(s^{(t)})/n)^{Q([w_A]_i)}}{\sum_{j=1}^{n} 1/\exp(\log(s^{(t)})/n)^{Q([w_A]_j)}},
\label{eq:probability}
\end{equation}
which favors samples with larger weights. Besides, the selection pressure $s^{(t)}$ is exponentially decayed to prevent excessive exploitation in later training stages (see Appendix D1 for details). 

D2ACE considers two types of complementary instance weights, namely uncertainty weights $\mathbf{w}_U$ and hardness weights $\mathbf{w}_H$, and obtains two corresponding types of sampling distributions based on Eq.\eqref{eq:probability}.
To adaptively balance their contributions during training, we introduce an epoch-wise Bernoulli random variable
 $\beta^t \sim \mathrm{Bernoulli}(p_\beta^t)$ to determine whether uncertainty-based ($\beta^t=1$) or hardness-based ($\beta^t=0$) sampling is used for each instance. The joint sampling probability of instance $\mathbf{x}_i$ is defined as:
\begin{equation}
\begin{aligned}
P(\mathbf{x}_i, \beta_t) & = P(\beta_t)\, P(\mathbf{x}_i \mid \beta_t) \\
& = \begin{cases}
p_{\beta}^t \, P(\mathbf{x}_i \mid w_U, t), & \beta_t = 1, \\[4pt]
(1 - p_{\beta}^t) \, P(\mathbf{x}_i \mid w_H, t), & \beta_t = 0.
\end{cases}
\end{aligned}
\label{eq:eachprob}
\end{equation}
Marginalizing over $\beta^t$ yields the \textit{Bernoulli mixture of two metrics-aware sampling} distributions:
\begin{equation}
 P(\mathbf{x}_i \mid t) = p_\beta^t \, P(\mathbf{x}_i \mid \mathbf{w}_U, t) + (1 - p_\beta^t) \, P(\mathbf{x}_i \mid \mathbf{w}_H, t)
 \label{eq:sampling_prob}
\end{equation}
The mixing coefficient $p_\beta^t$ linearly decays across epochs as follows:
\begin{equation}
p_{\beta}^{t}
= p_{\beta}^{t_{\text{start}}}
+ \frac{t - t_{\text{start}}}{t_{\text{end}} - t_{\text{start}}}
\left(
p_{\beta}^{t_{\text{end}}} - p_{\beta}^{t_{\text{start}}}
\right), 
\label{eq:linear_schedule_p}
\end{equation}
where $p_{\beta}^{t_{\text{start}}}$ and $p_{\beta}^{t_{\text{end}}}$ 
denote the mixing probabilities at the initial and final epochs, respectively.
We set  $p_{\beta}^{t_{\text{start}}}>p_{\beta}^{t_{\text{end}}}$, which progressively shifts the sampling focus from uncertain to hard instances. Specifically, early epochs emphasize uncertain instances to explore ambiguous regions and prevent premature convergence, while later epochs prioritize persistently misclassified hard instances to refine challenging decision boundaries and correct remaining systematic errors after initial learning.


\subsection{Algorithm}

Algorithm~\ref{alg:alogrithm} presents the training procedure of a deep multi-label model with D2ACE batch selection. After a warm-up stage with randomly sampling, mini-batches are selected via a Bernoulli mixture of uncertainty- and hardness-based distributions, controlled by the epoch-wise mixing coefficient $p_\beta^t$.
The \textbf{computational complexity} of our method is $O(n^2d + T(nqn_t+\sigma(\mathbf{Y})^{2}/n + \sigma(\mathbf{Z})q)$, where $\sigma(\cdot)$ denotes the number of nonzero elements. It benefits from label sparsity and is more efficient than ML-Unc that scales quadratically with $q$ (see Appendix~B for details).
The Algorithm~\ref{alg:alogrithm} is \textbf{guaranteed to converge} in the same sense as generic Adam \cite{Adam_convergency_CVPR19}. A formal analysis and proof are provided in Appendix~C.

\begin{algorithm}[!t]
\small
\KwIn{Training set: $\mathcal{D}$, \# epochs: $T$, \# warm-up epochs $T_w$, batch size: $b$, Bernoulli mixing coefficient in each epoch $\{p_\beta^{t}\}_{t=T_w+1}^T$, initialized model parameters $\Theta$}
\KwOut{Trained model $\Theta$}


\For{$t=1$ to \textit{T}}{
    \For{$k=1$ to $n/b$}{
        \eIf{$t < T_w$}{ 
            $\mathcal{B} \leftarrow$ Sample a batch randomly;
        }{ 
            $\mathbf{U} \leftarrow$ Update uncertainty by Eq.\eqref{eq:current_uncertain}-\eqref{eq:u_metric} \;
            $\mathbf{H} \leftarrow$ Update hardness by Eq.\eqref{eq:ema_flip}-\eqref{eq:hard} \;
            $\boldsymbol{\delta}_H,\boldsymbol{\delta}_U \leftarrow$ Update label dynamic weights by Eq.\eqref{eq:label_weight1}-\eqref{eq:label_weight3} \;
            $\boldsymbol{\phi}_H, \boldsymbol{\phi}_U \leftarrow$ Update label correlation enhanced weights by Eq.\eqref{eq:C_A}-\eqref{eq:Q_A} \;
            $\mathbf{w}_U, \mathbf{w}_H \leftarrow \boldsymbol{\delta}_U + \boldsymbol{\phi}_U, \boldsymbol{\delta}_H+ \boldsymbol{\phi}_H$ \;
            $\mathcal{B} \leftarrow$ Sample a batch based on Eq.\eqref{eq:eachprob};
        }
    Forward with $\mathcal{B}$\ and compute loss \;
    Backward and update $\Theta$\ by Adam optimizer;
    }
}
\caption{Training by D2ACE Batch Selection}
\label{alg:alogrithm}
\end{algorithm}


        

\section{Experiments}
\subsection{Experiment Setup}


\textbf{Datasets.} 
Table \ref{tab:datasets} shows 15 multi-label datasets spanning various domains and input formats used in the experiments.
The tabular multi-label datasets are from the MULAN repository~\cite{mulan}, where each instance is represented as a feature vector. 
VOC2007~\cite{VOC2007} and MS-COCO~\cite{MS-COCO} are two multi-label image datasets, which consist of raw images.


\begin{table}[t]
\centering
\resizebox{0.48\textwidth}{!}{
    \begin{tabular}{ccccccccccc}
        \toprule
        \multicolumn{2}{c}{Dataset} & $n$ & $d$ & $q$ & \textit{Card} & \textit{Dens}  & Domain \\ 
        \midrule
        \multirow{12}{*}{tabular} & cal500 & 502 & 68 & 174 & 26.04 & 0.15 & music \\
        & birds & 645 & 260 & 19 & 1.01 & 0.05 & audio \\
        & enron & 1702 & 1001 & 53 & 3.38 & 0.06 & text \\
        & scene & 2407 & 294 & 6 & 1.07 & 0.18 & vision \\
        & yeast & 2417 & 103 & 14 & 4.24 & 0.30 & biology \\
        & Corel5k & 5000 & 499 & 374 & 3.52 & 0.01 & vision \\
        & rcv1subset1 & 6000 & 944 & 101 & 2.88 & 0.03 & text \\
        & rcv1subset2 & 6000 & 944 & 101 & 2.63 & 0.03 & text \\
        & rcv1subset3 & 6000 & 944 & 101 & 2.61 & 0.03 & text \\
        & bibtex & 7395 & 1836 & 159 & 2.40 & 0.02 & text \\
        & yahoo-Arts & 7484 & 2314 & 25 & 1.67 & 0.07 & text \\
        & yahoo-Business & 11214 & 2192 & 28 & 1.47 & 0.06 & text \\
        & mediamill & 43907 & 120 & 101 & 4.38 & 0.04 & vision \\ \midrule
       \multirow{2}{*}{image} &  VOC2007 & 9963 & - & 20 & 1.45 & 0.07 & vision \\
       & MS-COCO & 123287 & - & 80 & 2.91 & 0.03 & vision \\
        \bottomrule
    \end{tabular}
}
\caption{Multi-label datasets.}
\label{tab:datasets}
\end{table}

\textbf{Baselines.} We compare D2ACE with seven batch selection methods, including Random, Active~\cite{active2017}, Recent~\cite{recent2020}, DIHCL~\cite{DIHCL_NIPS20}, Balance~\cite{mllbalance}, Hard-Imb~\cite{hard_imbal_2024ECML}, and ML-Unc~\cite{MLuncertain2025}. 
The Random baseline uses the default mini-batch sampling implemented in PyTorch. For Active, Recent, and DIHCL that are tailored for single-label data, we adapt them to the multi-label scenario by computing the instance weight as the aggregation of label-wise importance scores. 
For tabular datasets, we evaluate all batch selection methods on CLIF~\cite{CLIF}, DELA~\cite{DELA}, and PACA~\cite{PACA} base models.
For image datasets, we use HST~\cite{HST} and ML-GCN~\cite{MLGCN} with a ResNet-101 backbone pre-trained on ImageNet \cite{Imagenet} and standard preprocessing following~\cite{HST}.
The hyperparameters of batch selection baselines and base deep multi-label models follow their respective original implementations. The specific configurations for these and our method are listed in Appendix D.


\textbf{Evaluation.} 
For tabular datasets, we evaluate performance using Macro-F1, Macro-AUC, and Ranking-Loss~\cite{MLLreview_2013} that cover both classification and label ranking aspects. Results are reported as the average over stratified five-fold cross-validation~\cite{stratification}.
For image datasets, we use the Mean Average Precision (mAP) as the evaluation metric and report average results on the pre-defined split test set over five runs with different random seeds. 
In each run or fold, we record the test performance at the epoch that achieves the best validation performance.

\textbf{Implementation.} All experiments were conducted on a machine with an NVIDIA RTX 4090 GPU. 
The source code is available at \url{https://github.com/iunanyou/D2ACE}.

\subsection{Results}



\textbf{Tabular Data Results.} As shown in Table \ref{ta:avgRank_statisticalTest}, D2ACE is the best method across three base models and three evaluation metrics, and significantly outperforms the baselines in 61 out of 63 cases according to the Wilcoxon signed-rank test.
The success of our method stems from jointly modeling stage-wise metric utility and dynamic label importance, as well as enhancing instance selection by exploiting label correlations under local label context.
Based on detailed results presented in Table \ref{ta:CLIF_METRIC} and Appendix E, our method demonstrates more prominent advantages on large-scale datasets. 
Multi-label batch selection methods that explicitly account for label characteristics generally outperform single-label approaches, yet remain inferior to our method.
ML-Unc propagates label correlations dominated by irrelevant labels and relies solely on uncertainty, which may neglect extremely hard instances in later training.
Hard-Imb emphasizes hardness, making it prone to overfitting noisy samples in early training, while both Hard-Imb and Balance treat label imbalance as a static priority throughout training.
In Macro-F1, Balance, DIHCL, and Recent achieve the second-best performance with different base models. This is because Balance explicitly enforces label-balanced mini-batches, and DIHCL and Recent assess instance importance based on binary label predictions, which are more closely aligned with the F1 metric than prediction probabilities used in ML-Unc and Active or loss-based criteria adopted by Hard-Imb.



\begin{table}[!t]
\centering
\resizebox{0.5\textwidth}{!}{
\begin{tabular}{cccccccccc}
\toprule
Metric & Base & Random & Active & Recent & DIHCL & Balance & Hard-Imb & ML-Unc & \textbf{D2ACE} \\
\midrule
\multirow{3}{*}{\begin{tabular}{@{}c@{}}Macro\\-AUC\end{tabular}}
& CLIF   & $7.08^{*}$ & $5.54^{*}$ & $5.77^{*}$ & $\textcolor{blue}{3.08}^{*}$ & $4.15^{*}$ & $5.23^{*}$ & $\textcolor{LightGreen}{3.54}^{\dagger}$ & $\textbf{\textcolor{red}{1.31}}$ \\
& DELA   & $6.46^{*}$ & $4.85^{*}$ & $5.38^{-}$  & $5.31^{*}$ & $4.92^{*}$ & $\textcolor{LightGreen}{3.92}^{*}$ & $\textcolor{blue}{2.92}^{*}$ & $\textbf{\textcolor{red}{2.00}}$ \\
& PACA   & $5.31^{\dagger}$  & $4.69^{*}$ & $4.69^{*}$ & $5.31^{*}$ & $\textcolor{LightGreen}{4.38}^{\dagger}$  & $5.23^{*}$ & $\textcolor{blue}{4.00}$  & $\textbf{\textcolor{red}{2.15}}$ \\
\midrule
\multirow{3}{*}{\begin{tabular}{@{}c@{}}Macro\\-F1\end{tabular}}
& CLIF   & $6.62^{*}$ & $5.08^{*}$ & $4.69^{*}$ & $\textcolor{blue}{3.00}^{*}$ & $5.31^{*}$ & $5.31^{*}$ & $\textcolor{LightGreen}{4.00}^{\dagger}$ & $\textbf{\textcolor{red}{1.69}}$ \\
& DELA   & $6.46^{*}$ & $4.77^{*}$ & $\textcolor{blue}{3.23}$  & $5.92^{*}$ & $5.38^{*}$ & $\textcolor{LightGreen}{4.00}^{*}$ & $4.23^{*}$ & $\textbf{\textcolor{red}{1.92}}$ \\
& PACA   & $5.54^{*}$ & $5.54^{*}$ & $5.69^{*}$ & $5.38^{*}$ & $\textcolor{blue}{3.77}^{*}$ & $\textcolor{LightGreen}{4.00}^{*}$ & $4.31^{\dagger}$ & $\textbf{\textcolor{red}{1.69}}$ \\
\midrule
\multirow{3}{*}{\begin{tabular}{@{}c@{}}Ranking\\-Loss\end{tabular}}
& CLIF   & $6.85^{*}$ & $5.77^{*}$ & $4.85^{*}$ & $\textcolor{LightGreen}{3.54}^{*}$ & $4.31^{*}$ & $4.92^{*}$ & $\textcolor{blue}{3.00}^{*}$ & $\textbf{\textcolor{red}{1.69}}$ \\
& DELA   & $5.77^{*}$ & $4.85^{*}$ & $5.46^{*}$ & $4.92^{*}$ & $4.77^{*}$ & $\textcolor{blue}{3.62}^{*}$ & $\textcolor{LightGreen}{4.46}^{*}$ & $\textbf{\textcolor{red}{1.46}}$ \\
& PACA   & $5.69^{\dagger}$  & $5.31^{*}$ & $5.54^{*}$ & $4.92^{*}$ & $4.77^{\dagger}$  & $\textcolor{blue}{3.38}^{\dagger}$  & $\textcolor{LightGreen}{3.85}^{\dagger}$  & $\textbf{\textcolor{red}{2.08}}$ \\
\bottomrule
\end{tabular}
}
\caption{
Average ranks and Wilcoxon signed-rank test results on multi-label tabular datasets. The \textcolor{red}{\textbf{best}}, \textcolor{blue}{second}, and \textcolor{LightGreen}{third} ranks are highlighted. $\dagger$ ($*$) denotes that our method significantly outperforms baseline at $p<0.05$ ($p<0.01$) levels.} 
\label{ta:avgRank_statisticalTest}
\end{table}

\textbf{Image Data Results.} Table~\ref{ta:IMAGE_MAP} reports the mAP results of various batch selection methods under HST and ML-GCN base models on two multi-label image datasets. D2ACE consistently achieves the best performance across both base models, with more pronounced improvements on the larger MS-COCO dataset. 
DIHCL ranks second overall, primarily due to its temporally smoothed hardness estimation, which 
is more robust than relying solely on the loss at the current epoch (e.g., Hard-Imb) for  complex image classification tasks.
ML-Unc follows next, as it incorporates recent uncertainty variations together with label correlations. 

\begin{table*}[!t]
\centering
\resizebox{\textwidth}{!}{
\begin{tabular}{ccccccccc|cccccccc}
\toprule
\multirow{2}{*}{Dataset}
& \multicolumn{8}{c|}{Macro-AUC $\uparrow$}
& \multicolumn{8}{c}{Ranking-Loss $\downarrow$} \\
\cmidrule(lr){2-9} \cmidrule(lr){10-17}
& Random & Active & Recent & DIHCL & Balance & Hard-Imb & ML-Unc & \textbf{D2ACE} & Random & Active & Recent & DIHCL & Balance & Hard-Imb & ML-Unc & \textbf{D2ACE} \\
\midrule
cal500
& 0.5849(4) & 0.5826(6) & 0.5818(7) & 0.5869(3) & 0.5827(5) & 0.5807(8) & \textbf{\textcolor{Red}{0.5884(1)}} & 0.5879(2)
& 0.2318(6) & 0.2313(4) & 0.2317(5) & \textbf{\textcolor{Red}{0.2300(1)}} & 0.2310(2) & 0.2318(6) & 0.2311(3) & 0.2318(6) \\
birds
& 0.8026(6) & 0.8037(5) & 0.8006(8) & 0.8025(7) & 0.8046(3) & 0.8045(4) & 0.8050(2) & \textbf{\textcolor{Red}{0.8090(1)}}
& 0.1777(7) & 0.1710(2) & 0.1822(8) & 0.1773(6) & 0.1728(4) & 0.1739(5) & \textbf{\textcolor{Red}{0.1705(1)}} & 0.1720(3) \\
enron
& 0.7704(7) & 0.7751(4) & 0.7748(5) & 0.7729(6) & 0.7757(2) & 0.7695(8) & 0.7753(3) & \textbf{\textcolor{Red}{0.7766(1)}}
& 0.1205(5) & 0.1213(7) & 0.1209(6) & \textbf{\textcolor{Red}{0.1146(1)}} & 0.1242(8) & 0.1166(2) & 0.1186(4) & 0.1170(3) \\
scene
& 0.9481(8) & 0.9496(6) & 0.9489(7) & \textbf{\textcolor{Red}{0.9519(1)}} & 0.9504(3) & 0.9503(5) & 0.9504(3) & 0.9510(2)
& 0.0671(7) & 0.0669(6) & 0.0665(5) & 0.0650(2) & 0.0661(3) & 0.0671(7) & 0.0664(4) & \textbf{\textcolor{Red}{0.0629(1)}} \\
yeast
& 0.7226(7) & 0.7236(3) & 0.7230(5) & 0.7217(8) & 0.7234(4) & 0.7229(6) & 0.7242(2) & \textbf{\textcolor{Red}{0.7305(1)}}
& 0.1654(5) & 0.1648(4) & 0.1670(7) & 0.1682(8) & 0.1646(3) & 0.1658(6) & 0.1645(2) & \textbf{\textcolor{Red}{0.1616(1)}} \\
Corel5k
& 0.7582(8) & 0.7603(7) & 0.7614(5) & 0.7629(2) & 0.7619(4) & 0.7606(6) & 0.7621(3) & \textbf{\textcolor{Red}{0.7647(1)}}
& 0.1590(8) & 0.1584(5) & 0.1564(2) & 0.1589(6) & 0.1576(4) & 0.1589(6) & 0.1568(3) & \textbf{\textcolor{Red}{0.1544(1)}} \\
rcvsubset1
& 0.9243(8) & 0.9257(4) & 0.9256(5) & 0.9269(2) & 0.9250(7) & 0.9259(3) & 0.9254(6) & \textbf{\textcolor{Red}{0.9271(1)}}
& 0.0668(8) & 0.0664(6) & 0.0655(3) & 0.0641(2) & 0.0667(7) & 0.0662(5) & 0.0655(3) & \textbf{\textcolor{Red}{0.0633(1)}} \\
rcvsubset2
& 0.9211(8) & 0.9221(6) & 0.9222(5) & 0.9232(2) & 0.9223(3) & 0.9217(7) & 0.9223(3) & \textbf{\textcolor{Red}{0.9238(1)}}
& 0.0719(8) & 0.0709(7) & 0.0694(2) & 0.0696(4) & 0.0696(4) & 0.0698(6) & 0.0694(2) & \textbf{\textcolor{Red}{0.0675(1)}} \\
rcvsubset3
& 0.9168(8) & 0.9176(6) & 0.9188(4) & \textbf{\textcolor{Red}{0.9220(1)}} & 0.9185(5) & 0.9172(7) & 0.9205(3) & 0.9211(2)
& 0.0738(6) & 0.0743(8) & 0.0739(7) & 0.0705(2) & 0.0728(3) & 0.0736(5) & 0.0733(4) & \textbf{\textcolor{Red}{0.0697(1)}} \\
bibtex
& 0.9031(8) & 0.9036(4) & 0.9034(6) & 0.9045(2) & 0.9035(5) & 0.9037(3) & 0.9034(6) & \textbf{\textcolor{Red}{0.9061(1)}}
& 0.0879(8) & 0.0862(7) & 0.0861(6) & 0.0857(3) & 0.0858(5) & 0.0857(3) & 0.0855(2) & \textbf{\textcolor{Red}{0.0839(1)}} \\
yahoo-Arts1
& 0.7441(7) & 0.7438(8) & 0.7479(5) & 0.7495(2) & 0.7465(6) & 0.7482(4) & 0.7486(3) & \textbf{\textcolor{Red}{0.7512(1)}}
& 0.1578(8) & 0.1572(7) & 0.1569(5) & 0.1565(4) & 0.1564(3) & 0.1569(5) & 0.1563(2) & \textbf{\textcolor{Red}{0.1528(1)}} \\
yahoo-Business1
& 0.8063(7) & 0.8070(5) & 0.8047(8) & \textbf{\textcolor{Red}{0.8115(1)}} & 0.8076(3) & 0.8064(6) & 0.8072(4) & \textbf{\textcolor{Red}{0.8115(1)}}
& 0.1339(6) & 0.1334(4) & 0.1336(5) & 0.1332(2) & 0.1346(8) & 0.1332(2) & 0.1343(7) & \textbf{\textcolor{Red}{0.1267(1)}} \\
mediamill
& 0.8025(6) & 0.8012(8) & 0.8027(5) & 0.8043(3) & 0.8028(4) & \textbf{\textcolor{Red}{0.8129(1)}} & 0.8018(7) & 0.8108(2)
& 0.0441(7) & 0.0442(8) & 0.0438(2) & 0.0439(5) & 0.0438(2) & 0.0440(6) & 0.0438(2) & \textbf{\textcolor{Red}{0.0429(1)}} \\
\midrule
Avg(Rank)
& 7.08 & 5.54 & 5.77 & 3.08 & 4.15 & 5.23 & 3.54 & \textbf{\textcolor{Red}{1.31}}
& 6.85 & 5.77 & 4.85 & 3.54 & 4.31 & 4.92 & 3.00 & \textbf{\textcolor{Red}{1.69}}\\
\bottomrule
\end{tabular}}
\caption{The Macro-AUC and Ranking-Loss results of CLIF using various batch selection methods on multi-label tabular datasets.}
\label{ta:CLIF_METRIC}
\end{table*}

\begin{table*}[!t]
\centering
\resizebox{\textwidth}{!}{
\begin{tabular}{ccccccccc|cccccccc}
\toprule
\multirow{2}{*}{Dataset} & \multicolumn{8}{c|}{HST} & \multicolumn{8}{c}{ML-GCN} \\ 
\cmidrule(lr){2-9} \cmidrule(lr){10-17}
& Random & Active & Recent & DIHCL & Balance & Hard-Imb & ML-Unc & \textbf{D2ACE} & Random & Active & Recent & DIHCL & Balance & Hard-Imb & ML-Unc & \textbf{D2ACE} \\ 
\midrule
VOC2007 & 0.9300(5) & 0.9294(8) & 0.9308(4) & 0.9310(3) & 0.9298(7) & 0.9300(5) & 0.9314(2) & $\textbf{\textcolor{red}{0.9320(1)}}$
& 0.9293(6) & 0.9285(8) & 0.9295(5) & 0.9308(2) & 0.9289(7) & 0.9298(4) & 0.9305(3) & $\textbf{\textcolor{red}{0.9314(1)}}$ \\
MS-COCO & 0.8012(8) & 0.8054(7) & 0.8127(5) & 0.8186(2) & 0.8065(6) & 0.8182(3) & 0.8179(4) & $\textbf{\textcolor{red}{0.8252(1)}}$
& 0.7962(8) & 0.8025(6) & 0.8074(4) & 0.8103(2) & 0.8021(7) & 0.8055(5) & 0.8085(3) & $\textbf{\textcolor{red}{0.8142(1)}}$ \\
\midrule
Avg(Rank) & 6.50 & 7.50 & 4.50 & 2.50 & 6.50 & 4.00 & 3.00 & $\textbf{\textcolor{red}{1.00}}$
& 7.00 & 7.00 & 4.50 & 2.00 & 7.00 & 4.50 & 3.00 & $\textbf{\textcolor{red}{1.00}}$ \\
\bottomrule
\end{tabular}}
\caption{The mAP results of HST and ML-GCN using various batch selection methods on multi-label image datasets.}  
\label{ta:IMAGE_MAP}
\end{table*}

\begin{figure}[!t]
  \centering
  \includegraphics[width=1\linewidth]{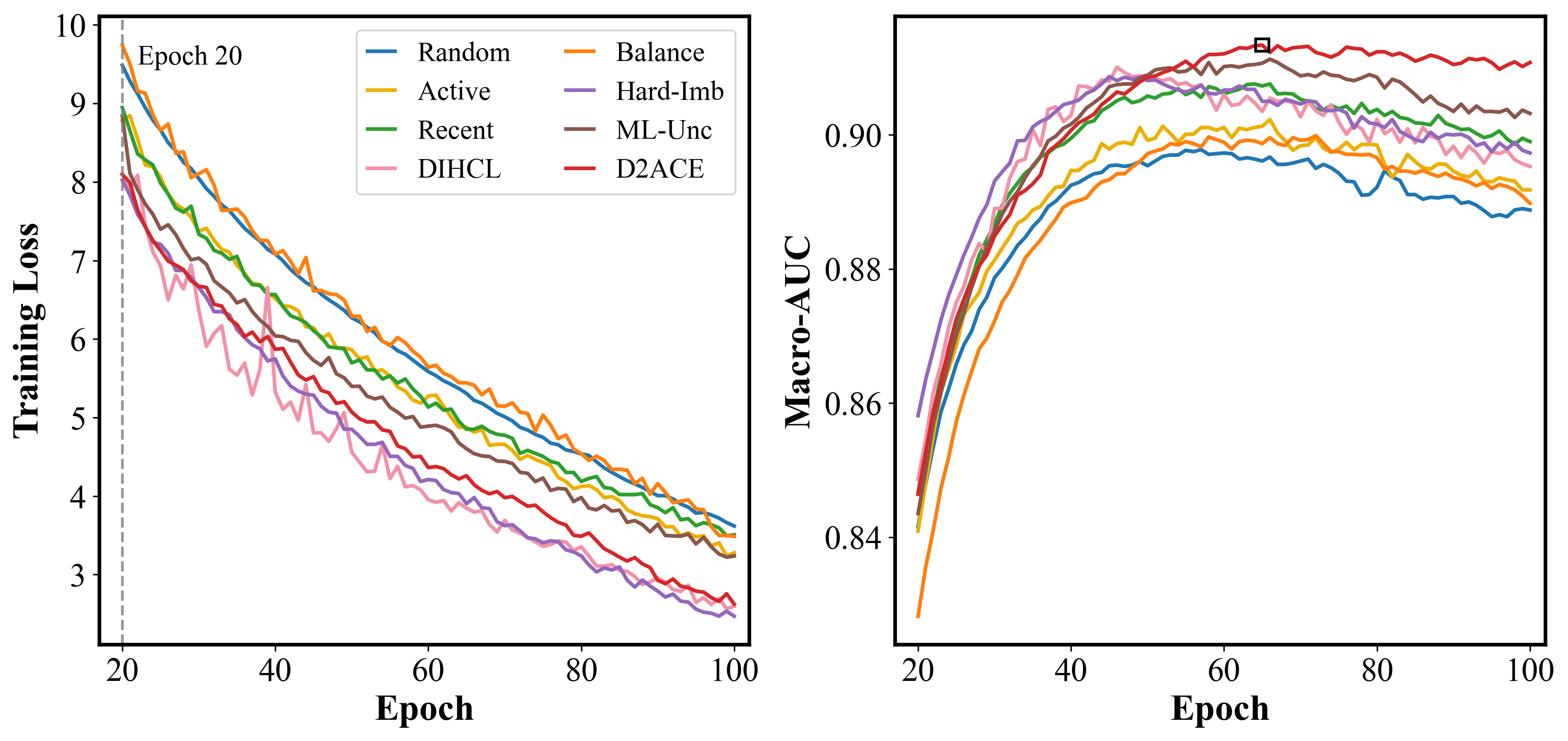}
  \caption{The convergence curves and Macro-AUC on validation set of CLIF using various batch selection methods on the bibtex dataset.}
  \label{fig:convergency}
\end{figure}

\subsection{Analysis}

\textbf{Convergency.} As shown in Figure \ref{fig:convergency}, hardness-based methods, including DIHCL and Hard-Imb, converge fastest, as they prioritize high-loss instances during training. However, their validation performance declines sharply after 50 epochs, indicating a tendency to overfit outliers.
In contrast, D2ACE converges faster than uncertainty-based approaches (ML-Unc, Recent, and Active), especially in later training stages, due to its gradual shift in emphasis toward difficult instances. Moreover, it achieves the best validation performance near epoch 60 (marked by the black square), surpassing all baselines, and remains stable thereafter.
Uncertainty-based methods exhibit mild performance degradation after reaching their peak validation performance.
Balance converges at a speed similar to Random but with larger fluctuations, mainly due to random filtering of negative samples.

\begin{figure}[!t]
  \centering
  \includegraphics[width=0.7\linewidth]{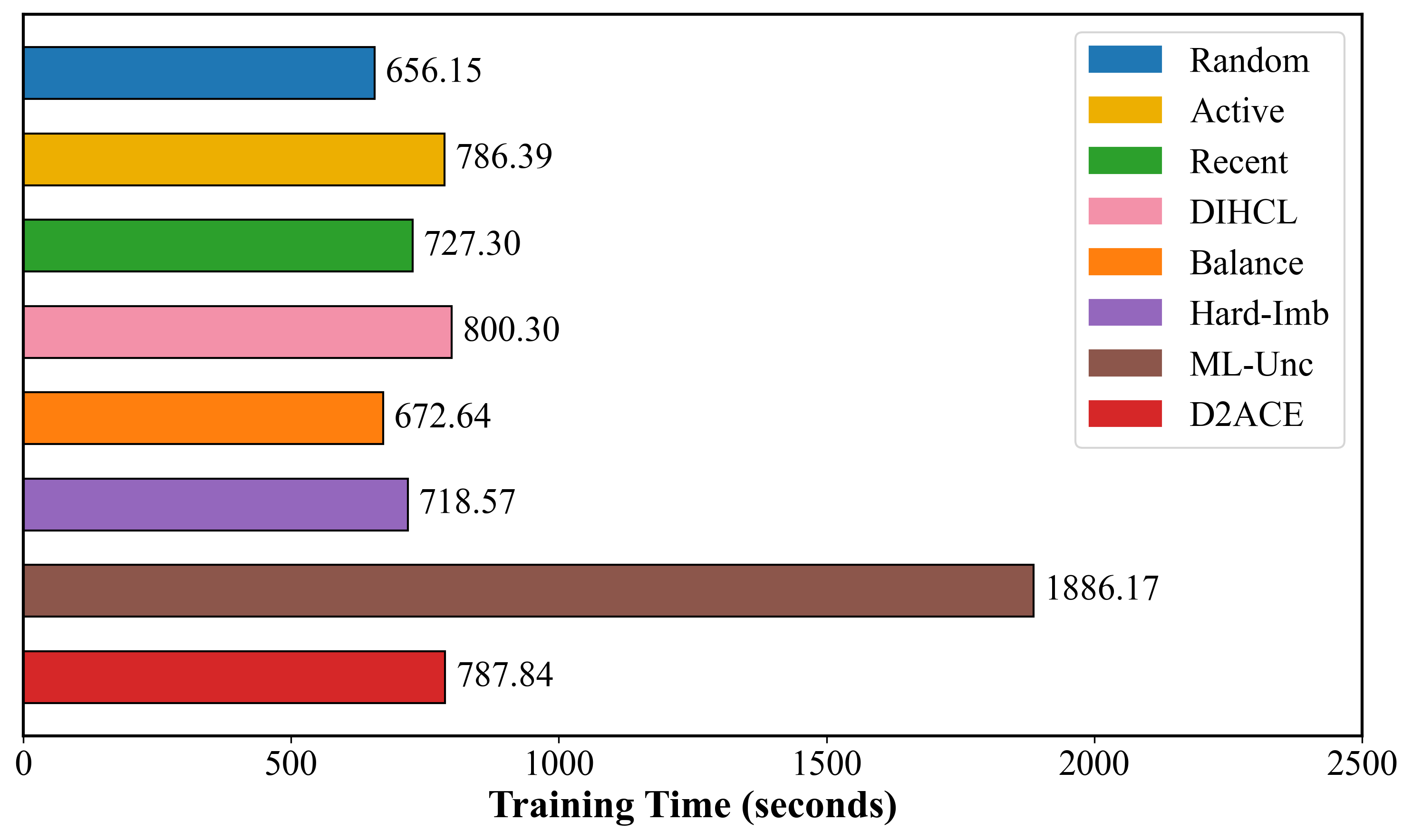}
  \caption{Training time of CLIF using various batch selection methods on the bibtex dataset (100 epochs).}
  \label{fig:time}
\end{figure}

\textbf{Efficiency.} As shown in Figure \ref{fig:time}, Random is the most time-efficient method, followed by Balance which only requires the computation of label statistics. 
D2ACE leverages sparse matrix operations to model label correlations, attaining comparable training time to label dependency-agnostic methods (Active, DIHCL) while exhibiting higher efficiency than the correlation-based ML-Unc.

\begin{figure}[!t]
  \centering
  \includegraphics[width=0.7\linewidth]{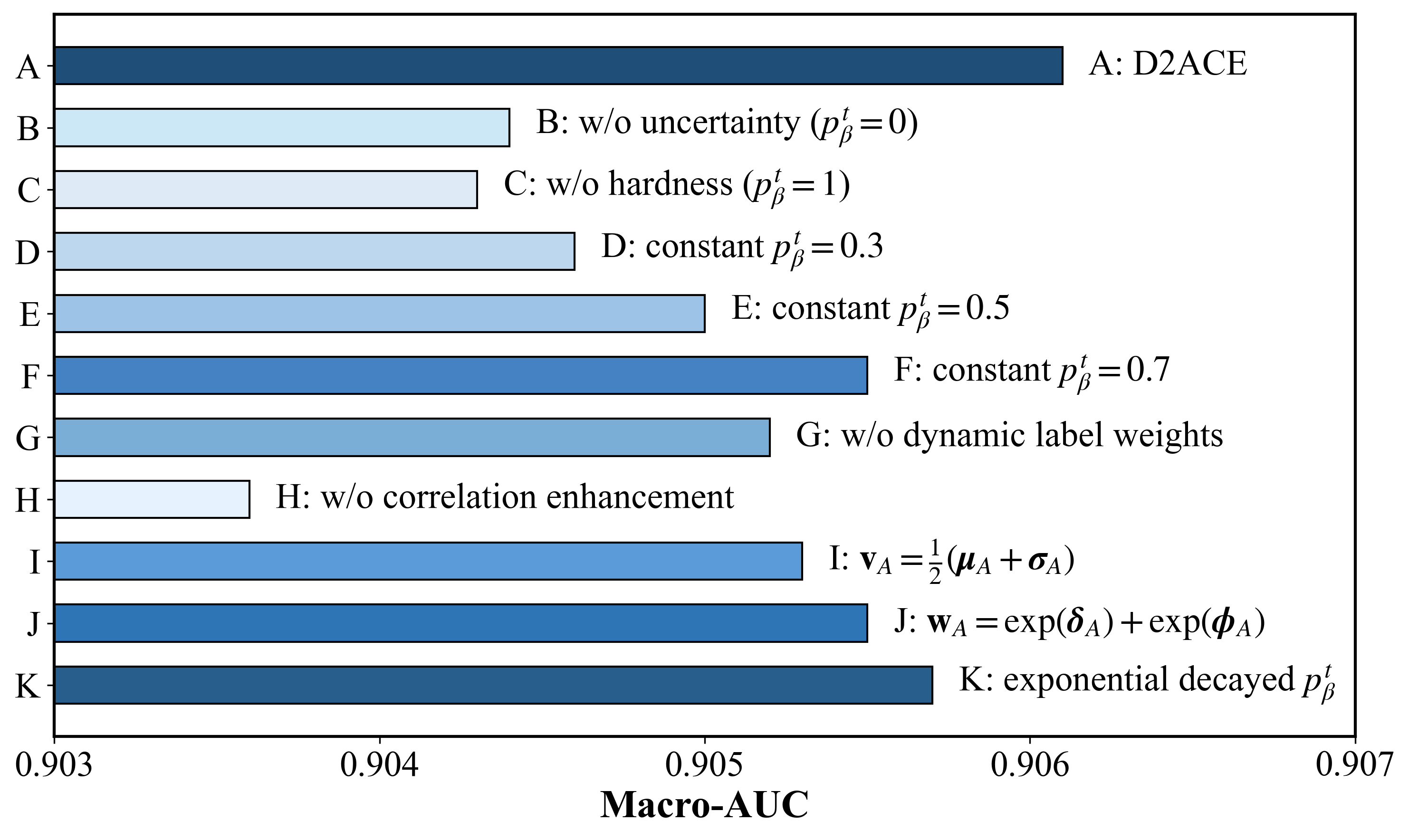}
  \caption{Ablation study of D2ACE with CLIF on the bibtex dataset.}
  \label{fig:ablation}
\end{figure}

\textbf{Ablation Study.}  Figure~\ref{fig:ablation} presents the ablation results of D2ACE under different variants, each removing or modifying a key component. 
Relying solely on either hardness (B) or uncertainty (C), as well as fixing the metric priority (D–F), leads to consistent performance degradation, demonstrating both the complementarity of the two metrics and the importance of stage-wise adjustment of metric emphasis.
Removing dynamic label weighting (G) also reduces performance, highlighting the necessity of capturing evolving label significance during training.
Most notably, excluding label correlation enhancement (H) causes the largest performance decline, which indicates the critical role of leveraging label correlations under local context.
A linear form for positive label weights $\mathbf{v}_A$ (I) is less effective than the exponential form in amplifying informative signals. Conversely, exponential transformation of instance weights $\mathbf{w}A$ yields overly uniform sampling probabilities. Exponential decay for $p_\beta^t$ (K) is less suitable, as it slows the shift from exploration to exploitation.

Further empirical analyses concerning parameter sensitivity, along with case studies on stage-wise sampling and label dynamic weighting, are detailed in Appendix F.

\section{Conclusion}
Batch selection in deep multi-label classification is challenging due to evolving metric utility, dynamic label importance, as well as sparse and instance-specific label correlations. To address these, we propose D2ACE, a novel batch selection framework that does not affect the convergence of generic Adam optimizers.
It handles metric dynamics by combining uncertainty and noise-resistant hardness through stage-wise Bernoulli mixture sampling, and captures label dynamics with epoch-wise weighting that adjusts label priorities based on current metrics. It also employs local context-aware correlation enhancement to focus on relevant labels and exploit instance-specific dependencies.
Extensive experiments on both tabular and image benchmarks demonstrate that D2ACE consistently outperforms state-of-the-art batch selection methods across various deep MLC models, while being more computationally efficient than the label correlation-based baseline.


\section*{Acknowledgments}
This work was supported in part by the National Natural Science Foundation of China (62302074, 62402077), Natural Science Foundation of Chongqing (CSTB2025NSCQ-GPX1269, CSTB2023NSCQ-MSX0613), the Science and Technology Research Program of Chongqing Municipal Education Commission (KJQN202300631).

\bibliographystyle{named}
\bibliography{ijcai26.bib}

@inproceedings{chai2024compositional,
  title={Compositional generalization for multi-label text classification: A data-augmentation approach},
  author={Chai, Yuyang and Li, Zhuang and Liu, Jiahui and Chen, Lei and Li, Fei and Ji, Donghong and Teng, Chong},
  booktitle={Proceedings of the AAAI Conference on Artificial Intelligence},
  pages={17727--17735},
  year={2024}
}

@inproceedings{lin2023effective,
  title={An effective deployment of contrastive learning in multi-label text classification},
  author={Lin, Nankai and Qin, Guanqiu and Wang, Gang and Zhou, Dong and Yang, Aimin},
  booktitle={Findings of the Association for Computational Linguistics},
  pages={8730--8744},
  year={2023}
}

@inproceedings{guo2023texts,
  title={Texts as images in prompt tuning for multi-label image recognition},
  author={Guo, Zixian and Dong, Bowen and Ji, Zhilong and Bai, Jinfeng and Guo, Yiwen and Zuo, Wangmeng},
  booktitle={Proceedings of the IEEE/CVF Conference on Computer Vision and Pattern Recognition},
  pages={2808--2817},
  year={2023}
}

@article{liu2021emerging,
  title={The emerging trends of multi-label learning},
  author={Liu, Weiwei and Wang, Haobo and Shen, Xiaobo and Tsang, Ivor W},
  journal={IEEE Transactions on Pattern Analysis and Machine Intelligence},
  pages={7955--7974},
  year={2021}
}

@inproceedings{shou2023concurrent,
  title={Concurrent multi-label prediction in event streams},
  author={Shou, Xiao and Gao, Tian and Subramanian, Dharmashankar and Bhattacharjya, Debarun and Bennett, Kristin P},
  booktitle={Proceedings of the AAAI Conference on Artificial Intelligence},
  pages={9820--9828},
  year={2023}
}

@inproceedings{recent2020,
  title={Carpe diem, seize the samples uncertain "at the moment" for adaptive batch selection},
  author={Song, Hwanjun and Kim, Minseok and Kim, Sundong and Lee, Jae-Gil},
  booktitle={Proceedings of the ACM International Conference on Information \& Knowledge Management},
  pages={1385--1394},
  year={2020}
}

@inproceedings{active2017,
  title={Active bias: Training more accurate neural networks by emphasizing high variance samples},
  author={Chang, Haw-Shiuan and Learned-Miller, Erik and McCallum, Andrew},
  booktitle={Proceedings of the International Conference on Advances in Neural Information Processing Systems},
  year={2017}
}

@inproceedings{MLuncertain2025,
  title={Batch selection for multi-label classification guided by uncertainty and dynamic label correlations},
  author={Zhou, Ao and Liu, Bin and Wang, Jin and Tsoumakas, Grigorios},
  booktitle={Proceedings of the AAAI Conference on Artificial Intelligence},
  pages={22902--22909},
  year={2025}
}

@inproceedings{DIHCL_NIPS20,
  title={Curriculum learning by dynamic instance hardness},
  author={Zhou, Tianyi and Wang, Shengjie and Bilmes, Jeffrey},
  booktitle={Proceedings of the International Conference on Advances in Neural Information Processing Systems},
  pages={8602--8613},
  year={2020}
}

@article{online2015,
  title={Online batch selection for faster training of neural networks},
  author={Loshchilov, Ilya and Hutter, Frank},
  journal={arXiv preprint arXiv:1511.06343},
  year={2015}
}

@inproceedings{hard_imbal_2024ECML,
  title={Multi-label adaptive batch selection by highlighting hard and imbalanced samples},
  author={Zhou, Ao and Liu, Bin and Peng, Zhaoyang and Wang, Jin and Tsoumakas, Grigorios},
  booktitle={Proceedings of the Joint European Conference on Machine Learning and Knowledge Discovery in Databases},
  pages={265--281},
  year={2024}
}

@inproceedings{maini2022characterizing,
  title={Characterizing datapoints via second-split forgetting},
  author={Maini, Pratyush and Garg, Saurabh and Lipton, Zachary and Kolter, J Zico},
  booktitle={Proceedings of the International Conference on Advances in Neural Information Processing Systems},
  pages={30044--30057},
  year={2022}
}

@article{deepsurvey2024,
  title={Deep learning for multi-label learning: A comprehensive survey},
  author={Tarekegn, Adane Nega and Ullah, Mohib and Cheikh, Faouzi Alaya},
  journal={arXiv preprint arXiv:2401.16549},
  year={2024}
}

@article{toneva2018,
  title={An empirical study of example forgetting during deep neural network learning},
  author={Toneva, Mariya and Sordoni, Alessandro and Combes, Remi Tachet des and Trischler, Adam and Bengio, Yoshua and Gordon, Geoffrey J},
  journal={arXiv preprint arXiv:1812.05159},
  year={2018}
}

@inproceedings{Adam_convergency_CVPR19,
  author={Fangyu Zou and Li Shen and Zequn Jie and Weizhong Zhang and Wei Liu},
  title={A sufficient condition for convergences of Adam and RMSProp},
  booktitle={Proceedings of the IEEE/CVF Conference on Computer Vision and Pattern Recognition},
  pages={11127--11135},
  year={2019}
}

@inproceedings{ICME2024positive,
  title={Positive label is all you need for multi-label classification},
  author={Yuan, Zhixiang and Zhang, Kaixin and Huang, Tao},
  booktitle={Proceedings of the IEEE International Conference on Multimedia and Expo (ICME)},
  pages={1--6},
  year={2024}
}

@article{Ada-boundary2020,
  title={Ada-boundary: Accelerating DNN training via adaptive boundary batch selection},
  author={Song, Hwanjun and Kim, Sundong and Kim, Minseok and Lee, Jae-Gil},
  journal={Machine Learning},
  volume={109},
  number={9},
  pages={1837--1853},
  year={2020}
}

@inproceedings{mllbalance,
  title={Doing the best we can with what we have: Multi-label balancing with selective learning for attribute prediction},
  author={Hand, Emily and Castillo, Carlos and Chellappa, Rama},
  booktitle={Proceedings of the AAAI Conference on Artificial Intelligence},
  pages={6878--6885},
  year={2018}
}

@article{CLIF,
  title={Collaborative learning of label semantics and deep label-specific features for multi-label classification},
  author={Hang, Jun-Yi and Zhang, Min-Ling},
  journal={IEEE Transactions on Pattern Analysis and Machine Intelligence},
  volume={44},
  number={12},
  pages={9860--9871},
  year={2021}
}

@article{DELA,
  title={Dual perspective of label-specific feature learning for multi-label classification},
  author={Hang, Jun-Yi and Zhang, Min-Ling},
  journal={ACM Transactions on Knowledge Discovery from Data},
  volume={19},
  number={1},
  pages={1--30},
  year={2024}
}

@inproceedings{PACA,
  title={End-to-end probabilistic label-specific feature learning for multi-label classification},
  author={Hang, Jun-Yi and Zhang, Min-Ling and Feng, Yanghe and Song, Xiaocheng},
  booktitle={Proceedings of the AAAI Conference on Artificial Intelligence},
  pages={6847--6855},
  year={2022}
}

@inproceedings{HotVae,
  title={Hot-vae: Learning high-order label correlation for multi-label classification via attention-based variational autoencoders},
  author={Zhao, Wenting and Kong, Shufeng and Bai, Junwen and Fink, Daniel and Gomes, Carla},
  booktitle={Proceedings of the AAAI Conference on Artificial Intelligence},
  pages={15016--15024},
  year={2021}
}

@inproceedings{MLGCN,
  title={Multi-label image recognition with graph convolutional networks},
  author={Chen, Zhao-Min and Wei, Xiu-Shen and Wang, Peng and Guo, Yanwen},
  booktitle={Proceedings of the IEEE/CVF Conference on Computer Vision and Pattern Recognition},
  pages={5177--5186},
  year={2019}
}

@article{HST,
  title={Heterogeneous semantic transfer for multi-label recognition with partial labels},
  author={Chen, Tianshui and Pu, Tao and Liu, Lingbo and Shi, Yukai and Yang, Zhijing and Lin, Liang},
  journal={International Journal of Computer Vision},
  volume={132},
  number={12},
  pages={6091--6106},
  year={2024}
}

@inproceedings{MS-COCO,
  title={Microsoft coco: Common objects in context},
  author={Lin, Tsung-Yi and Maire, Michael and Belongie, Serge and Hays, James and Perona, Pietro and Ramanan, Deva and Doll{\'a}r, Piotr and Zitnick, C Lawrence},
  booktitle={Proceedings of the European Conference on Computer Vision},
  pages={740--755},
  year={2014}
}

@article{VOC2007,
  title={The pascal visual object classes (voc) challenge},
  author={Everingham, Mark and Van Gool, Luc and Williams, Christopher KI and Winn, John and Zisserman, Andrew},
  journal={International Journal of Computer Vision},
  volume={88},
  number={2},
  pages={303--338},
  year={2010}
}

@inproceedings{stratification,
  title={On the stratification of multi-label data},
  author={Sechidis, Konstantinos and Tsoumakas, Grigorios and Vlahavas, Ioannis},
  booktitle={Proceedings of the  Joint European Conference on Machine Learning and Knowledge Discovery in Databases},
  pages={145--158},
  year={2011}
}

@article{new_BR,
  title={Binary relevance for multi-label learning: An overview},
  author={Zhang, Min-Ling and Li, Yu-Kun and Liu, Xu-Ying and Geng, Xin},
  journal={Frontiers of Computer Science},
  volume={12},
  number={2},
  pages={191--202},
  year={2018}
}

@inproceedings{New_CC,
  title={AdaBoost.C2: Boosting classifiers chains for multi-label classification},
  author={Li, Jiaxuan and Zhu, Xiaoyan and Wang, Jiayin},
  booktitle={Proceedings of the AAAI Conference on Artificial Intelligence},
  pages={8580--8587},
  year={2023}
}

@article{RAkEL,
  title={Random k-labelsets for multilabel classification},
  author={Tsoumakas, Grigorios and Katakis, Ioannis and Vlahavas, Ioannis},
  journal={IEEE Transactions on Knowledge and Data Engineering},
  volume={23},
  number={7},
  pages={1079--1089},
  year={2010}
}

@article{MLkNN,
  author={Zhang, Min-Ling and Zhou, Zhi-Hua},
  title={ML-KNN: A lazy learning approach to multi-label learning},
  journal={Pattern Recognition},
  volume={40},
  number={7},
  pages={2038--2048},
  year={2007}
}

@inproceedings{ML_deep_fortest,
  author={Liang Yang and Xi-Zhu Wu and Yuan Jiang and Zhi-Hua Zhou},
  title={Multi-label learning with deep forest},
  booktitle={Proceedings of the European Conference on Artificial Intelligence},
  pages={1634--1641},
  year={2020}
}

@inproceedings{Imagenet,
  title={Imagenet: A large-scale hierarchical image database},
  author={Deng, Jia and Dong, Wei and Socher, Richard and Li, Li-Jia and Li, Kai and Fei-Fei, Li},
  booktitle={Proceedings of the IEEE Conference on Computer Vision and Pattern Recognition},
  pages={248--255},
  year={2009}
}

@inproceedings{C-Tran,
  title={General multi-label image classification with transformers},
  author={Lanchantin, Jack and Wang, Tianlu and Ordonez, Vicente and Qi, Yanjun},
  booktitle={Proceedings of the IEEE/CVF Conference on Computer Vision and Pattern Recognition},
  pages={16478--16488},
  year={2021}
}

@inproceedings{ResNet,
  title={Deep residual learning for image recognition},
  author={He, Kaiming and Zhang, Xiangyu and Ren, Shaoqing and Sun, Jian},
  booktitle={Proceedings of the IEEE Conference on Computer Vision and Pattern Recognition},
  pages={770--778},
  year={2016}
}

@article{mulan,
  title={Mulan: A java library for multi-label learning},
  author={Tsoumakas, Grigorios and Spyromitros-Xioufis, Eleftherios and Vilcek, Jozef and Vlahavas, Ioannis},
  journal={The Journal of Machine Learning Research},
  volume={12},  
  pages={2411--2414},
  year={2011}
}

@article{MLLreview_2013,
  title = "A review on multi-label learning algorithms",
  author = "Zhang, Min-Ling and Zhou, Zhi-Hua",
  journal = "IEEE Transactions on Knowledge and Data Engineering",
  volume={26},
  number={8},
  pages = "1819--1837",
  year = "2013"
}

@article{MCE_imbalancedML,
  author       = {Zhihan Ning and
                  Zhixing Jiang and
                  David Zhang},
  title        = {Exploiting meta-learned confidences for imbalanced multilabel learning},
  journal      = {IEEE Transactions on Neural Networks and Learning Systems},
  pages        = {10242--10256},
  year         = {2025}
}

@article{ML_protein_task,
  title={Deep learning model for protein multi-label subcellular localization and function prediction based on multi-task collaborative training},
  author={Bai, Peihao and Li, Guanghui and Luo, Jiawei and Liang, Cheng},
  journal={Briefings in Bioinformatics},
  volume={25},
  number={6},
  pages={bbae568},
  year={2024},
}

@article{oversampling_nn,
  title={Oversampling multi-label data based on natural neighbor and label correlation},
  author={Liu, Bin and Zhou, Ao and Wei, Bingkun and Wang, Jin and Tsoumakas, Grigorios},
  journal={Expert Systems with Applications},
  volume={259},
  pages={125257},
  year={2025},
}

@inproceedings{AEMLO,
  title={AEMLO: Autoencoder-guided multi-label oversampling},
  author={Zhou, Ao and Liu, Bin and Wang, Jin and Sun, Kaiwei and Liu, Kelin},
  booktitle={Joint European Conference on Machine Learning and Knowledge Discovery in Databases},
  pages={107--124},
  year={2024},
  organization={Springer}
}

@inproceedings{TAIpp,
  title     = {Tai++: Text as image for multi-label image classification by co-learning transferable prompt},
  author    = {Wu, Xiangyu and Jiang, Qing-Yuan and Yang, Yang and Wu, Yi-Feng and Chen, Qing-Guo and Lu, Jianfeng},
  booktitle = {Proceedings of the Thirty-Third International Joint Conference on Artificial Intelligence},
  pages = {5226--5234},
  year  = {2024}
}

@article{FLEM,
  title={Interactive fusion label enhancement for multi-label learning},
  author={Zhao, Xingyu and An, Yuexuan and Xu, Ning and Qi, Lei and Geng, Xin},
  journal={ACM Transactions on Knowledge Discovery from Data},
  volume={19},
  number={7},
  pages={1--23},
  year={2025},
}

\end{document}


\maketitle

\appendix

\renewcommand{\thetable}{A\arabic{table}}  
\renewcommand{\thefigure}{A\arabic{figure}}  
\setcounter{table}{0}  
\setcounter{figure}{0} 

\section{Supplementary Details of D2ACE}

\subsection{Instance-Adaptive Dependencies}

\begin{figure}[h]
  \centering
  \includegraphics[width=0.8\linewidth]{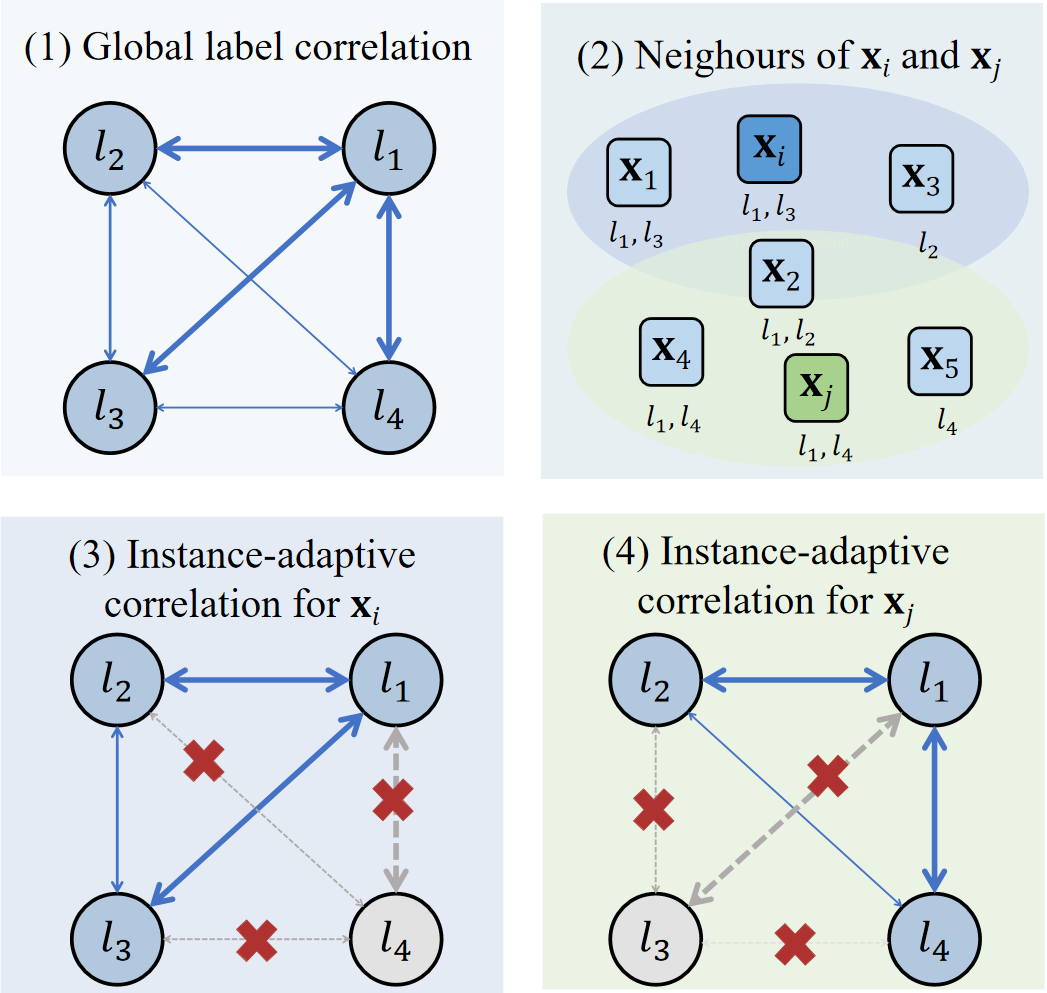}
  \caption{Global vs. instance-adaptive label correlations.}
  \label{fig:local-aware_label_correlation}
\end{figure}

We provide Fig.\ref{fig:local-aware_label_correlation} to compare global and instance-adaptive label dependencies. While global correlations are shared across instances, our method activates only those supported by each instance’s local neighborhood. As a result, different instances ($\mathbf{x}_i$ and $\mathbf{x}_j$) yield distinct dependency structures by filtering out locally irrelevant correlations.

\subsection{Analytical Comparison with ML-Unc}

In ML-Unc \cite{MLuncertain2025}, the label correlation enhanced uncertainty matrix is defined as:
\begin{equation}
    \mathbf{R}_{U} = \mathbf{U}\mathbf{C}_{U},
\end{equation}
where $\mathbf{C}_{U}$ denotes the mutual-information-based label correlation matrix estimated from $\mathbf{U}$ without label masking.
Element-wise, this computation can be written as:
\begin{equation}
    [\mathbf{R}_{U}]_{ij} = \sum_{k=1}^{q} U_{ik} C_{kj}.
\end{equation}
The uncertainty weight for the $i$-th sample is obtained by summing over all labels:
\begin{equation}
\begin{aligned}
    [\mathbf{w}_{U}]_{i}
    & = \sum_{j=1}^{q} [\mathbf{R}_{U}]_{ij} \\
    & = \sum_{j=1}^{q} \sum_{k=1}^{q} U_{ik} C_{kj} \\
    & = \sum_{k=1}^{q} U_{ik} \sum_{j=1}^{q} C_{kj} \\
    & = \sum_{j=1}^{q} U_{ij} \left( \sum_{k=1}^{q} C_{jk} \right)  \quad\quad\text{\# swapping indices $j$ and $k$.}
\label{eq:w_u_i_MLuncertain}
\end{aligned}
\end{equation}

In contrast, D2ACE defines the label correlation-aware uncertainty matrix as:
\begin{equation}
    \mathbf{R}_{U} = (\mathbf{U} \odot \mathbf{Y}) \odot \big( \mathbf{Z}\mathbf{C}_{U} \big),
\end{equation}
which explicitly incorporates both a label mask $\mathbf{Y}$ and a local-region indicator $\mathbf{Z}$.
The element-wise formulation is given by
\begin{equation}
\begin{aligned}
    [\mathbf{R}_{U}]_{ij}
    & = [\mathbf{M}_{U}]_{ij} \sum_{k=1}^{q} Z_{ik} C_{kj} \\
    & = (Y_{ij} U_{ij}) \sum_{k=1}^{q} Z_{ik} C_{kj}.
\end{aligned}
\end{equation}
Accordingly, the uncertainty-based weight for sample $\mathbf{x}_i$ is computed as:
\begin{equation}
\begin{aligned}
    [\mathbf{w}_{U}]_{i}
    & = \sum_{j=1}^{q} [\mathbf{R}_{U}]_{ij} \\
    & = \sum_{j=1}^{q} \sum_{k=1}^{q} (Y_{ij} U_{ij}) (Z_{ik} C_{kj}) \\
    & = \sum_{j=1}^{q} (Y_{ij} U_{ij})
        \left( \sum_{k=1}^{q} Z_{ik} C_{kj} \right).
\label{eq:w_u_i_ours}
\end{aligned}
\end{equation}

Comparing Eq.~\eqref{eq:w_u_i_MLuncertain} and Eq.~\eqref{eq:w_u_i_ours}, ML-Unc can be viewed as a special case of our framework in which the label mask $\mathbf{Y}$ is omitted and the local region degenerates to the entire dataset, i.e.,
$\mathcal{N}(\mathbf{x}_i) = \mathcal{D}$ and $\mathbf{Z} = \mathbf{1}_{n \times q}$.

\section{Computational Complexity Analysis \label{sec:complexity}}

Let \(n\) denote the number of training samples, \(q\) the number of labels, \(K\) the number of nearest neighbors, \(T\) the total number of training epochs, and \(n_t\) the sliding-window size.
The computational complexities of existing and our batch selection methods are summarized in Table \ref{tab:computational_complexity}. Below, we provide a detailed explanation.


\textbf{Random}: $O(T)$

\textbf{Active}~\cite{active2017}: $O(T^2nq)$
\begin{itemize}
    \item Computing prediction variance at epoch $t$: $O(nqt)$
\end{itemize}

\textbf{Recent}~\cite{recent2020}: $O(Tnqn_t)$
\begin{itemize}
    \item Computing predictive uncertainty at epoch $t$: $O(nqn_t)$
\end{itemize}

\textbf{DIHCL}~\cite{DIHCL_NIPS20}: $O(T^2nq)$
\begin{itemize}
    \item Computing dLoss or prediction flip at epoch $t$: $O(nq)$
    \item Normalizing weight by previous learning rate at epoch $t$: $O(nqt)$
\end{itemize}

\textbf{Balance}~\cite{mllbalance}:  $O(Tnq)$
\begin{itemize}
    \item Computing target and batch distributions at epoch $t$: $O(nq)$
\end{itemize}

\textbf{Hard-Imb}~\cite{hard_imbal_2024ECML}: $O(Tnq + n^2d+nkq)$
\begin{itemize}
    \item Computing weights (only once): $O(n^2d+nkq)$
    \begin{itemize}
        \item Computing imbalance ratio: $O(nq)$
        \item Retrieving $K$NNs of all instances: $O(n^2d)$
        \item Computing local imbalance matrix: $O(nkq)$
        \item Computing class imbalance aware weights: $O(nq)$
    \end{itemize}
    \item Complexity at epoch $t$: $O(nq)$
    \begin{itemize}
        \item Sum per-label losses: $O(nq)$
        \item Weight multiplying loss: $O(n)$
    \end{itemize}
\end{itemize}

\textbf{ML-Unc}~\cite{MLuncertain2025}: $O(T(nqn_t+nq^2))$
\begin{itemize}
    \item Complexity at epoch $t$: $O(nqn_t+nq^2))$
    \begin{itemize}
        \item Computing binary entropy of current prediction: $O(nq)$
        \item Computing label uncertainty: $O(nqn_t)$
        \item Computing mutual-information-based label correlations: $O(nq^{2})$
        \item Computing label correlation enhanced uncertain matrix: $O(nq^{2})$
    \end{itemize}
\end{itemize}

\textbf{D2ACE} (Ours): $O\left(n^2d + T\left(nqn_t+\sigma(\mathbf{Y})^{2}/n + \sigma(\mathbf{Z})q \right)\right)$
\begin{itemize}
    \item Retrieving $K$NNs of all instances (only once): $O(n^2d)$
    \item Complexity at epoch $t$: $O\left(nqn_t+\sigma(\mathbf{Y})^{2}/n + \sigma(\mathbf{Z})q \right)$, where $\sigma(\cdot)$ denotes the number of nonzero elements
    \begin{itemize}
       \item Computing uncertain matrix $\mathbf{U}$: $O(nqn_t)$
       \item Computing hard matrix $\mathbf{H}$: $O(nq)$
       \item Computing dynamic label weight: $O(nq)$
       \item Computing label correlations $\mathbf{R}_A$: $O\left(\sigma(\mathbf{Y})^{2}/n + \sigma(\mathbf{Z})q \right)$
    \end{itemize}
\end{itemize}

\begin{table}[h]
\centering
\footnotesize
\resizebox{0.46\textwidth}{!}{
\begin{tabular}{cc}
\toprule
Method & Overall Complexity \\
\midrule
Random & $O(T)$ \\
\midrule
Active & $O(T^2nq)$ \\
Recent & $O(Tnqn_t)$ \\
DIHCL & $O(T^2nq)$ \\
\midrule
Balance & $O(Tnq)$ \\
Hard-Imb & $O(Tnq+n^2d+nkq)$ \\
ML-Unc & $O(T(nqn_t+nq^2))$ \\
\midrule
\textbf{D2ACE} & $O(n^2d + T(nqn_t+\sigma(\mathbf{Y})^{2}/n + \sigma(\mathbf{Z})q)$ \\
\bottomrule
\end{tabular}
}
\caption{Computational complexity of batch selection methods.}
\label{tab:computational_complexity}
\end{table}

Random is the most efficient method, as it introduces no additional computation beyond epoch-wise iteration. Balance, Recent, and Hard-Imb exhibit near-linear scaling with respect to the number of instances and labels during training, apart from a one-time preprocessing cost in Hard-Imb.
In contrast, Active and DIHCL incur a quadratic dependence on the number of training epochs ($O(T^2nq)$), stemming from their reliance on accumulated historical predictions or cumulative normalization across epochs, which leads to increased computational overhead as training progresses.
ML-Unc is the most computationally expensive method, since its mutual-information-based label correlation modeling introduces a dominant $O(nq^2)$ cost at each epoch, making it poorly scalable when the label space is large.
By comparison, D2ACE explicitly models label correlations while avoiding dense $q^2$ operations by exploiting label sparsity. As a result, its per-epoch complexity depends on $\sigma(\mathbf{Y})$ and $\sigma(\mathbf{Z})$ rather than $q^2$. This leads to training times comparable to correlation-ignored methods (Active and DIHCL), while being substantially more efficient than ML-Unc, demonstrating a favorable trade-off between computational efficiency and expressive correlation modeling. Consequently, our approach is better suited for large-scale multi-label datasets.


\section{Convergence Guarantee}


\begin{lemma} [Strict Positivity of Bernoulli Mixture Sampling]
\label{lemma:strict_positive}
The Bernoulli mixture of two metrics-aware sampling probability, defined as:
\begin{equation}
 P(\mathbf{x}_i \mid t) = p_\beta^t \, P(\mathbf{x}_i \mid \mathbf{w}_U, t) + (1 - p_\beta^t) \, P(\mathbf{x}_i \mid \mathbf{w}_H, t), 
 \label{eq:sampling_prob}
\end{equation}
is strictly positive for all $\mathbf{x}_i \in D$ and all epochs $t$
\[
P(\mathbf{x}_i \mid t) > 0,  \quad \forall i, t.
\]
\end{lemma}

\begin{proof}

We first consider any weight vector $\mathbf{w}_A \in \{\mathbf{w}_U, \mathbf{w}_H\}$.
The corresponding quantization index
\[
Q([w_A]_i) = \left\lceil 1 - \frac{[w_A]_i}{\Delta} \right\rceil
\]
satisfies
\[
0 \le Q([w_A]_i) \le \left\lceil \frac{1}{\Delta} \right\rceil,
\]
which implies that $Q([w_A]_i)$ is finite for all instances.

The associated weight-aware sampling probability is given by
\[
P(\mathbf{x}_i \mid \mathbf{w}_A, t) =
\frac{1/\exp(\log(s^{(t)})/n)^{Q([w_A]_i)}}{
\sum_{j=1}^{n} 1/\exp(\log(s^{(t)})/n)^{Q([w_A]_j)} }.
\]
Since all terms in both the numerator and denominator are strictly positive,
it follows that
\[
P(\mathbf{x}_i \mid \mathbf{w}_A, t) > 0, \quad \forall i, t.
\]

Next, note that the Bernoulli mixing coefficient satisfies
$0 \le p_\beta^t \le 1$ for all epochs $t$.
Therefore, the mixture sampling probability
\[
P(\mathbf{x}_i \mid t)
= p_\beta^t \, P(\mathbf{x}_i \mid \mathbf{w}_U, t)
+ (1-p_\beta^t) \, P(\mathbf{x}_i \mid \mathbf{w}_H, t)
\]
is a strictly positive convex combination of two strictly positive terms, which completes the proof.
\end{proof}

\begin{lemma}[Unbiasedness of Mini-Batch Gradient]
At epoch $t$, let $\ell^t = \frac{1}{n} \sum_{i=1}^{n} \ell^t_i$
denote the empirical objective, where
$\ell_i^t = \sum_{j=1}^{q} \ell_{ij}^t$ is the loss associated with instance
$\mathbf{x}_i$.
A mini-batch $B^t = \{\mathbf{x}_{i_1}, \ldots, \mathbf{x}_{i_b}\}$ of size $b$
is sampled \emph{i.i.d.} according to the sampling probability defined in Eq.~\eqref{eq:sampling_prob}.
The stochastic gradient estimator
\[
\widetilde{\mathbf{G}}^t
= \frac{1}{b} \sum_{\mathbf{x}_{i_k} \in B^t}
\frac{1}{n \, P(\mathbf{x}_{i_k} \mid t)} \nabla \ell_{i_k}^t
\]
is an unbiased estimator of the full empirical gradient
\[
\mathbf{G}^t = \nabla \ell^t.
\]
\end{lemma}

\begin{proof}

By Lemma~\ref{lemma:strict_positive}, the sampling probability satisfies
\[
P(\mathbf{x}_i \mid t) > 0 \quad \forall i,t,
\]
ensuring that the estimator $\widetilde{\mathbf{G}}^t$ is well defined.

Since the mini-batch samples are drawn independently, the joint probability of observing $B^t$ is
\[
P_{B}(B^t \mid t) = \prod_{k=1}^b P(\mathbf{x}_{i_k} \mid t).
\]

Taking expectation with respect to the mini-batch sampling distribution, we obtain
\begin{equation}
    \begin{aligned}
\mathbb{E}_{B^t}\!\left[ \widetilde{\mathbf{G}}^t \right]
&= \mathbb{E}_{B^t}
\left[
\frac{1}{b} \sum_{k=1}^{b}
\frac{1}{n \, P(\mathbf{x}_{i_k} \mid t)} \nabla \ell_{i_k}^t
\right] \\
&= \frac{1}{b} \sum_{k=1}^{b}
\mathbb{E}_{B^t}
\left[
\frac{1}{n \, P(\mathbf{x}_{i_k} \mid t)} \nabla \ell_{i_k}^t
\right]. \\
& \quad \text{\# linearity of expectation} \\
\label{eq:expection_G}
    \end{aligned}
\end{equation}
For any fixed index $k$, the expectation can be expanded explicitly as
\begin{equation}
    \begin{aligned}
    & \mathbb{E}_{B^t} \left[ \frac{1}{n \, P(\mathbf{x}_{i_k} \mid t)} \nabla \ell_{i_k}^t \right] \\
    & = \sum_{i_1=1}^n \cdots \sum_{i_b=1}^n \left( \prod_{l=1}^b P(\mathbf{x}_{i_l} \mid t) \right)  \frac{1}{n \, P(\mathbf{x}_{i_k} \mid t)} \nabla l_{i_k}^t \\
    &= \sum_{i_1=1}^n \cdots \sum_{i_b=1}^n \left( P(\mathbf{x}_{i_k} \mid t) \prod_{l \neq k} P(\mathbf{x}_{i_l} \mid t) \right) \frac{1}{n \, P(\mathbf{x}_{i_k} \mid t)} \nabla l_{i_k}^t \\
    &= \sum_{i_1=1}^n \cdots \sum_{i_b=1}^n \left( \prod_{l \neq k} P(\mathbf{x}_{i_l} \mid t) \right) \frac{1}{n} \nabla l_{i_k}^t \\
    &= \left(  \sum_{i_1=1}^n \cdots \sum_{i_{k-1}=1}^n \sum_{i_{k+1}=1}^n \cdots \sum_{i_{b}=1}^n \prod_{l \neq k} P(\mathbf{x}_{i_l} \mid t) \right) \\ 
    & \quad  \left( \sum_{i_k=1}^n \frac{1}{n} \nabla l_{i_k}^t \right) \\
    & = \left( \prod_{l \neq k} \sum_{i_l=1}^n P(\mathbf{x}_{i_l} \mid t) \right) \left( \sum_{i_k=1}^n \frac{1}{n} \nabla l_{i_k}^t \right) \\
    &= 1 \times \left( \frac{1}{n} \sum_{i_k=1}^n \nabla l_{i_k}^t \right)   \\
     & \quad \text{\# sum of sampling probability of all instances is 1}\\
    & = \frac{1}{n} \sum_{i=1}^n \nabla l_i^t \quad\quad \text{\# relabeling the index}.
    \label{eq:expection_G_2}
    \end{aligned}
\end{equation}


Substituting Eq.\eqref{eq:expection_G_2} back to Eq.\eqref{eq:expection_G} yields
\begin{align*}
\mathbb{E}_{B^t}\!\left[ \widetilde{\mathbf{G}}^t \right]
&= \frac{1}{b} \sum_{k=1}^{b}
\left( \frac{1}{n} \sum_{i=1}^{n} \nabla \ell_i^t \right) \\
&= \frac{1}{n} \sum_{i=1}^{n} \nabla \ell_i^t
= \nabla \ell^t .
\end{align*}
Therefore, $\widetilde{\mathbf{G}}^t$ is an unbiased estimator of the full empirical gradient
$\mathbf{G}^t$.

\end{proof}

\begin{lemma}[Uniform Boundedness of Mini-Batch Gradient Estimator (Adam-Safe)]
\label{lemma:bounded_detailed}
Assume that the stochastic gradients have uniformly bounded second moments, i.e.,
\[
\|\ell_i^t\|^2 \le G, \quad \forall t.
\]
According to Lemma \ref{lemma:strict_positive}, there exists a constant $p_{\min} > 0$ such that
\[
P(\mathbf{x}_i \mid t) \ge p_{\min}, \quad \forall i,t.
\]
Given the mini-batch gradient estimator
\[
\widetilde{\mathbf{G}}^t
= \frac{1}{|B^t|} \sum_{\mathbf{x}_i \in B^t}
\frac{1}{n P(\mathbf{x}_i \mid t)} \nabla \ell_i^t,
\]
where samples in $B^t$ are drawn i.i.d. according to $P(\mathbf{x}_i \mid t)$, 
there exists a constant $\widetilde{G} > 0$, independent of $t$, such that
\[
\mathbb{E}\big[\|\widetilde{\mathbf{G}}^t\|^2\big] \le \widetilde{G}.
\]
\end{lemma}

\begin{proof}
We begin by computing the second moment of the estimator:
\begin{equation}
\begin{aligned}
& \mathbb{E}_{B^t}\!\left[ \|\widetilde{\mathbf{G}}^t\|^2 \right] \\
& = \mathbb{E}_{B^t}
\Bigg[
\Big\| \frac{1}{b} \sum_{k=1}^{b} \frac{1}{n P(\mathbf{x}_{i_k} \mid t)} \nabla \ell_{i_k}^t \Big\|^2
\Bigg] \\
& = \frac{1}{b^2}\mathbb{E}_{B^t}
\Bigg[
\Big\|\sum_{k=1}^{b} \frac{1}{n P(\mathbf{x}_{i_k} \mid t)} \nabla \ell_{i_k}^t \Big\|^2
\Bigg] \\
\label{eq:expection_G2}
\end{aligned}
\end{equation}
Expanding the squared norm gives
\begin{equation}
\begin{aligned}
& \Bigg\| \sum_{k=1}^b \frac{1}{n P(\mathbf{x}_{i_k} \mid t)} \nabla \ell_{i_k}^t \Big\|^2 \\
&= \sum_{k=1}^b \Big\| \frac{1}{n P(\mathbf{x}_{i_k} \mid t)} \nabla \ell_{i_k}^t \Big\|^2 \\
& \quad + \sum_{k \neq k'} 
\Big\langle \frac{\nabla \ell_{i_k}^t}{n P(\mathbf{x}_{i_k} \mid t)}, \frac{\nabla \ell_{i_{k'}}^t}{n P(\mathbf{x}_{i_{k'}} \mid t)} \Big\rangle.
\label{eq:expand_sq_norm}
\end{aligned}
\end{equation}
Applying the linearity of expectation yields the separation into diagonal and
cross terms
\begin{equation}
\begin{aligned}
& \mathbb{E}\!\left[\|\widetilde{\mathbf{G}}^t\|^2\right] =
\frac{1}{b^2}\sum_{k=1}^{b}
\mathbb{E}\!\left[
\Big\|
\frac{1}{nP(\mathbf{x}_{i_k}\mid t)}\nabla \ell_{i_k}^t
\Big\|^2
\right]  \\
& \quad\quad +
\frac{1}{b^2}\sum_{k\neq k'}
\mathbb{E}\!\left[
\Big\langle
\frac{\nabla \ell_{i_k}^t}{nP(\mathbf{x}_{i_k}\mid t)},
\frac{\nabla \ell_{i_{k'}}^t}{nP(\mathbf{x}_{i_{k'}}\mid t)}
\Big\rangle
\right].
\end{aligned}
\label{eq:expection_G2_2}
\end{equation}
The diagonal terms correspond to the second moments of individual scaled gradients, while the cross terms arise from inner products between gradients of different mini-batch samples.

Due to the i.i.d.\ sampling assumption, for any fixed $k$, each diagonal term in Eq.\eqref{eq:expection_G2_2} is reduced to
\begin{equation}
\begin{aligned}
& \mathbb{E}_{B^t}
\Bigg[ \Big\| \frac{1}{n P(\mathbf{x}_{i_k} \mid t)} \nabla \ell_{i_k}^t \Big\|^2 \Bigg] \\
& = \sum_{i_1=1}^n \cdots \sum_{i_b=1}^n \left( \prod_{l=1}^b P(\mathbf{x}_{i_l} \mid t) \right)  \Big\| \frac{1}{n P(\mathbf{x}_{i_k} \mid t)} \nabla \ell_{i_k}^t \Big\|^2 \\
&= \sum_{i_1=1}^n \cdots \sum_{i_b=1}^n \left( P(\mathbf{x}_{i_k} \mid t) \prod_{l \neq k} P(\mathbf{x}_{i_l} \mid t) \right)  \\
&  \quad \frac{1}{n^2 P(\mathbf{x}_{i_k} \mid t)^2} \Big\|  \nabla \ell_{i_k}^t \Big\|^2 \\
& = \sum_{i_1=1}^n \cdots \sum_{i_b=1}^n \left( \prod_{l \neq k} P(\mathbf{x}_{i_l} \mid t) \right)  \sum_{i_k=1}^n \frac{1}{n^2 P(\mathbf{x}_{i_k} \mid t)} \Big\|  \nabla \ell_{i_k}^t \Big\|^2 \\
& = \sum_{i=1}^n \frac{1}{n^2 P(\mathbf{x}_{i} \mid t)} \Big\|  \nabla \ell_{i}^t \Big\|^2 \quad\text{\# same as steps 3-8 in Eq.\eqref{eq:expection_G_2}} 
\label{eq:expection_norm2_for_k}
\end{aligned}
\end{equation}
Summing over $k=1,\ldots,b$, we obtain
\begin{equation}
\begin{aligned}
& \frac{1}{b^2} \sum_{k=1}^b \mathbb{E}_{B^t}
\Bigg[ \Big\| \frac{1}{n P(\mathbf{x}_{i_k} \mid t)} \nabla \ell_{i_k}^t \Big\|^2 \Bigg] \\
& = \frac{1}{b^2} \sum_{k=1}^b \sum_{i=1}^n \frac{1}{n^2 P(\mathbf{x}_{i} \mid t)} \Big\|  \nabla \ell_{i}^t \Big\|^2 \\
& = \frac{1}{b} \sum_{i=1}^n \frac {\Big\| \nabla \ell_{i}^t \Big\|^2} {n^2 P(\mathbf{x}_{i} \mid t)}
\label{eq:expection_norm2}
\end{aligned}
\end{equation}

Due to the i.i.d.\ sampling assumption, for any $k \neq k'$ pair, each cross  term in Eq.\eqref{eq:expection_G2_2} is reduced to
\begin{equation}
\begin{aligned}
& \mathbb{E}_{B^t} 
\Bigg[ \Big\langle \frac{\nabla \ell_{i_k}^t}{n P(\mathbf{x}_{i_k} \mid t)}, \frac{\nabla \ell_{i_{k'}}^t}{n P(\mathbf{x}_{i_{k'}} \mid t)} \Big\rangle \Bigg] \\ 
& =  \Big\langle \mathbb{E}_{B^t} 
\Bigg[ \frac{\nabla \ell_{i_k}^t}{n P(\mathbf{x}_{i_k} \mid t)} \Bigg], \mathbb{E}_{B^t} 
\Bigg[ \frac{\nabla \ell_{i_{k'}}^t}{n P(\mathbf{x}_{i_{k'}} \mid t)} \Bigg]\Big\rangle  \\ 
& = \Big\langle \frac{1}{n} \sum_{i=1}^n \nabla l_i^t,  \frac{1}{n} \sum_{i'=1}^n \nabla l_i'^t \Big\rangle \quad \text{\# according to Eq.\eqref{eq:expection_G_2}} \\
& = \Big\| \frac{1}{n} \sum_{i=1}^n \nabla l_i^t \Big\|^2.
\label{eq:cross_inner_product_for_kk}
\end{aligned}
\end{equation}
Since there are $b(b-1)$ such pairs, we obtain
\begin{equation}
\begin{aligned}
& \frac{1}{b^2} \sum_{k \neq k'} \mathbb{E}_{B^t} 
\Bigg[ \Big\langle \frac{\nabla \ell_{i_k}^t}{n P(\mathbf{x}_{i_k} \mid t)}, \frac{\nabla \ell_{i_{k'}}^t}{n P(\mathbf{x}_{i_{k'}} \mid t)} \Big\rangle \Bigg] \\
& = \frac{1}{b^2} \sum_{k \neq k'} \Big\| \frac{1}{n} \sum_{i=1}^n \nabla l_i^t \Big\|^2  \\ 
& = \frac{b-1}{b}  \Big\| \frac{1}{n} \sum_{i=1}^n \nabla l_i^t \Big\|^2
\label{eq:cross_inner_product}
\end{aligned}
\end{equation}

Combining Eq.\eqref{eq:expection_norm2} and Eq.\eqref{eq:cross_inner_product} leads to
\begin{equation}
\begin{aligned}
& \mathbb{E}_{B^t}\!\left[ \|\widetilde{\mathbf{G}}^t\|^2 \right]  \\
& \quad = \frac{1}{b} \sum_{i=1}^n \frac{\|\nabla \ell_i^t\|^2}{n^2 P(\mathbf{x}_i \mid t)}  
+ \frac{b-1}{b} \Big\| \frac{1}{n} \sum_{i=1}^n \nabla \ell_i^t \Big\|^2.
\label{eq:expection_G2_3}
\end{aligned}
\end{equation}
Using the boundedness assumption $\|\nabla \ell_i^t\|^2 \le G$ and the lower bound $P(\mathbf{x}_i \mid t) \ge p_{\min}$,
each term in Eq.~\eqref{eq:expection_G2_3} can be upper bounded explicitly.
\begin{equation}
\frac{1}{b} \sum_{i=1}^n \frac{\|\nabla \ell_i^t\|^2}{n^2 P(\mathbf{x}_i \mid t)}  \le 
\frac{1}{b} \sum_{i=1}^n \frac{G}{n^2 p_{min}} =
\frac{G}{bnp_{min}} 
\end{equation}
\begin{equation}
\frac{b-1}{b} \Big\| \frac{1}{n} \sum_{i=1}^n \nabla \ell_i^t \Big\|^2 
\le \frac{b-1}{b} \sum_{i=1}^n \big\| \nabla \ell_i^t \big\|^2 
\le \frac{(b-1)G}{b} 
\label{eq:2nd_term_bound}
\end{equation}
where the 1st $\le$ in Eq.\eqref{eq:2nd_term_bound} is derived due to Jensen's inequality. 
Putting the bounds together yields
\begin{equation}
\begin{aligned}
    \mathbb{E}_{B^t}\!\left[ \|\widetilde{\mathbf{G}}^t\|^2 \right] & \le 
    \frac{G}{bnp_{min}} + \frac{(b-1)G}{b}  \\
    & = G\left( 1-\frac{1}{b}+\frac{1}{bnp_{min}} \right) \\
    & \le \frac{G}{np_{min}} \quad \text{\# $np_{min}\le 1$ and $b\ge1$} \\
    & \triangleq  \widetilde{G}>0,
\end{aligned}
\end{equation}
where $\widetilde{G} > 0$ is a constant independent of $t$.
This completes the proof.
\end{proof}

We next demonstrate that the proposed mini-batch sampling strategy does not affect the convergence behavior of Generic Adam.
Specifically, we verify that the stochastic gradient estimator induced by our sampling scheme satisfies all assumptions required in the convergence analysis in \cite{Adam_convergency_CVPR19}: 
\begin{enumerate}[label=\textbf{(A\arabic*)}]
    \item The objective function is lower-bounded.
    \item The gradient of $\ell$ is $L$-Lipschitz continuous.
    \item The stochastic gradient is an unbiased estimate.
    \item The second-order moment of the stochastic gradient is uniformly upper-bounded.
\end{enumerate}
Lemma~2 shows that the mini-batch gradient estimator $\widetilde{\mathbf{G}}^t$ is an unbiased estimator of the full empirical gradient, i.e., $\mathbb{E}[\widetilde{\mathbf{G}}^t] = \nabla \ell_t$, thereby verifying Assumption~\textbf{(A3)}.
Moreover, Lemma~\ref{lemma:bounded_detailed} demonstrates that the second moment of $\widetilde{\mathbf{G}}^t$ is uniformly bounded by a constant independent of the iteration index $t$, which ensures that Assumption~\textbf{(A4)} holds.
Since Assumptions~\textbf{(A1)}–\textbf{(A2)} concern only the objective function and are unaffected by the sampling strategy,  all assumptions required in \cite{Adam_convergency_CVPR19} are satisfied under the proposed sampling scheme. Consequently, \textbf{the convergence guarantees for generic Adam derived in \cite{Adam_convergency_CVPR19} remain valid for the proposed D2ACE}.

Let $\{\alpha_t\}$, $\{\beta_t\}$, and $\{\theta_t\}$ denote the stepsize, first-moment decay, and second-moment decay parameters of Generic Adam,
respectively. According to Corollary 9 in \cite{Adam_convergency_CVPR19}, generic Adam converges provided that the following sufficient conditions are satisfied:
\begin{enumerate}
    \item \textbf{(Bounded momentum parameter)}
    \[
    \beta_t \le \beta < 1, \quad \forall t.
    \]
    \item \textbf{(Monotone variance smoothing)}
    \[
    0 < \theta_t < 1, \quad \text{and } \theta_t \text{ is non-decreasing in } t.
    \]
    \item \textbf{(Almost non-increasing effective stepsize)}
    \[
    \chi_t = \frac{\alpha_t}{\sqrt{1-\theta_t}}
    \quad \text{is almost non-increasing}.
    \]
    \item \textbf{(Vanishing cumulative stepsize ratio)}
    \[
    \frac{\sum_{t=1}^{T} \alpha_t \sqrt{1-\theta_t}}{T \alpha_T} = o(1)
    \quad \text{as } T \to \infty.
    \]
\end{enumerate}

Furthermore, since the convergence rate analysis in \cite{Adam_convergency_CVPR19} relies exclusively on Assumptions~\textbf{(A1)}–\textbf{(A4)}
and the above parameter conditions, the non-asymptotic convergence rate of generic Adam still holds by using D2ACE to sample batches. Corollary~10 in \cite{Adam_convergency_CVPR19} shows that for the parameter choices
\[
\alpha_t = \frac{\eta}{t^s}, \quad s \in (0,1),
\]
\[
\theta_t =
\begin{cases}
1 - \dfrac{\alpha}{K^r}, & t < K, \\[6pt]
1 - \dfrac{\alpha}{t^r}, & t \ge K,
\end{cases}
\]
where $0 < r \le 2s < 2$ ensures that the effective stepsize $\chi_t$ is non-increasing, generic Adam achieves the following
non-asymptotic convergence rate:
\[
\mathbb{E}\!\left[\|\nabla \ell^t_i\|^2\right]
\le
\begin{cases}
\mathcal{O}\!\left(T^{-r/2}\right), & \text{if } \dfrac{r}{2} + s < 1, \\[6pt]
\mathcal{O}\!\left(\dfrac{\log T}{T^{1-s}}\right), & \text{if } \dfrac{r}{2} + s = 1, \\[10pt]
\mathcal{O}\!\left(T^{-(1-s)}\right), & \text{if } \dfrac{r}{2} + s > 1.
\end{cases}
\]

\section{Hyperparameter Setting and Implementation Details}


\subsection{Batch Selection Methods}
For all batch selection methods, except for Random, the warm-up epochs are set to $T_w = 10$ for tabular data and $T_w = 5$ for image data. 
Except for DIHCL, all methods use the quantization index strategy to project instance weights to sampling probabilities. To prevent excessive exploitation in later training stages, the selection pressure $s^{(t)}$ is exponentially decayed from an initial value $s^{(T_w+1)}$ at the first epoch after warm-up to 1 at the last epoch:
\begin{equation}
s^{(t)} = s^{(T_w+1)} 
\left( \exp \left( \frac{\log(1 / s^{(T_w+1)})}{T -(T_w+1)} \right) \right)^{t-(T_w+1)}
\end{equation}
where the initial selection pressure $s^{T_w+1}=100$.

The hyperparameter settings for each batch selection method are as follows:
\begin{itemize}
    \item \textbf{Random}: randomly shuffling and sampling mini-batches without replacement.
    \item \textbf{Active}~\cite{active2017}: estimating prediction variance plus its confidence interval.
    \item \textbf{Recent}~\cite{recent2020}: sliding window size $n_t=5$
    \item \textbf{DIHCL}~\cite{DIHCL_NIPS20}: hardness metric chosen from loss change (dLoss) and prediction flip (Flip), discount factor $\gamma = 0.95$,  using softmax-based Exp3 sampling.
    \item \textbf{Balance}~\cite{mllbalance}: each label is sampled toward a balanced distribution (positive:negative \( = 1:1 \)).
    \item \textbf{Hard-Imb}~\cite{hard_imbal_2024ECML}: number of nearest neighbors for computing local imbalance $K=5$, using the binary cross entropy loss for hardness estimation.
    \item \textbf{ML-Unc}~\cite{MLuncertain2025}: sliding window size $n_t=5$, trade-off parameter $\lambda_1=0.5$.
    \item \textbf{D2ACE (ours)}: sliding window size $n_t=5$, trade-off parameter $\lambda_1=0.5$, using binary cross entropy loss, smoothing factor $\lambda_2=0.7$, the number of nearest neighbors $K \in \{4,5,6,7,8\}$, 
    Bernoulli sampling probability starts at \( p_{\beta}^{\text{start}} = 0.7 \) and ends at \( p_{\beta}^{\text{end}} = 0.3 \), with transition epochs \( t^{\text{start}} = 30 \) and \( t^{\text{end}} = 70 \) for tabular datasets (\( t^{\text{start}} = 5 \) and \( t^{\text{end}} = 15 \) for image datasets).
\end{itemize}

\subsection{Base Deep Multi-Label Models}

All tabular deep multi-label models use the same optimizer and learning rate schedule to ensure a fair comparison. Specifically, we adopt the Adam optimizer with an initial learning rate of $\mathit{lr}=1\times10^{-4}$, momentum coefficients $\beta_1=0.9$ and $\beta_2=0.999$, and weight decay set to $1\times10^{-4}$. A linear warm-up strategy is applied over the first 10 epochs, during which the learning rate increases from $0$ to $1\times10^{-4}$, followed by a constant learning rate of $1\times10^{-4}$ for the remaining epochs. The batch size is fixed to 128, and all models are trained for 100 epochs.

The hyperparameter settings for each model are as follows:
\begin{itemize}
    \item \textbf{CLIF}~\cite{CLIF}: number of GIN layers $=2$; label embedding dimension $d_{e}\in\{64,128,256\}$; dimensionality of label-specific features $=512$; negative slope of LeakyReLU $=0.1$; trade-off parameter $\lambda\in\{10^{-5},10^{-4},\dots,10\}$.
    
    \item \textbf{DELA}~\cite{DELA}: binary concrete gate temperature parameter $\tau=\frac{2}{3}$, Monte Carlo sampling steps $L=1$; trade-off parameter $\lambda\in\{10^{-5},10^{-4},\dots,10\}$.
    
    \item \textbf{PACA}~\cite{PACA}: number of hidden units in the autoregressive network $d_{\tau}=16$, Monte Carlo sampling steps $L=1$, trade-off parameters $\alpha\in\{1,2,5,10,20,50\}$ and $\gamma\in\{10^{-4},10^{-3},10^{-2},10^{-1},1,10\}$.
\end{itemize}

All image models adopt the data preprocessing pipeline described in \cite{HST}. Specifically, images are first resized to a scale of 512 using bicubic interpolation. To introduce multi-scale variation, we randomly crop regions with sizes selected from ${512, 448, 384, 320, 256}$, followed by resizing to a fixed resolution of $448 \times 448$. Random horizontal flipping is applied with a probability of 0.5. Finally, images are normalized using ImageNet statistics, with mean $[0.485, 0.456, 0.406]$ and standard deviation $[0.229, 0.224, 0.225]$.
All image deep multi-label models also use the same optimizer and learning rate schedule to ensure a fair comparison. 
Specifically, we adopt the Adam optimizer with an initial learning rate of $\mathit{lr}=1\times10^{-5}$, momentum coefficients $\beta_1=0.9$ and $\beta_2=0.999$, and weight decay set to $5\times10^{-4}$. The batch size is fixed to 32, and all models are trained for 20 epochs.

The hyperparameter settings for each model are as follows:
\begin{itemize}
    \item \textbf{HST}~\cite{HST}: 
    $\theta_{intra} = 1.0$, $\theta_{cross} = 1.0$; 
    intra co-occurrence weight = 1.0, inter distance weight = 1.0, prototype number = 50.
    \item \textbf{ML-GCN}~\cite{MLGCN}: 
    LeakyReLU with negative slope 0.2, 2 layers GCN with output dimensions of 1024 and 2048, correlation matrix parameters $\tau = 0.4$ and $p = 0.2$.
    
    
\end{itemize}



\section{Supplementary Results}


Tables \ref{ta:Macro-AUC}-\ref{ta:Ranking-loss} detail the performance of three base models using various batch selection methods on thirteen tabular datasets, measured by Macro-AUC, Macro-F1, and Ranking-Loss, respectively. Each entry shows the metric value with its rank, and the best results are highlighted. Table \ref{ta:statistical test} further provides statistical validation through Wilcoxon signed-ranks tests at the 0.05 significance level, indicating whether D2ACE significantly outperforms (``win") or ties with other approaches for each base model-evaluation metric combination, with corresponding p-values in brackets.

\begin{table*}[!t]
\centering
\resizebox{\textwidth}{!}{
\begin{tabular}{ccccccccc|cccccccc|cccccccc}
\toprule
\multirow{2}{*}{Dataset} 
& \multicolumn{8}{c}{CLIF} 
& \multicolumn{8}{c}{DELA} 
& \multicolumn{8}{c}{PACA} \\ 
\cmidrule(lr){2-9} \cmidrule(lr){10-17} \cmidrule(lr){18-25}
& Random & Active & Recent & DIHCL & Balance & Hard-Imb & ML-Unc & \textbf{D2ACE} & Random & Active & Recent & DIHCL & Balance & Hard-Imb & ML-Unc & \textbf{D2ACE} & Random & Active & Recent & DIHCL & Balance & Hard-Imb & ML-Unc  & \textbf{D2ACE} \\ 
\midrule
cal500
& 0.5849(4) & 0.5826(6) & 0.5818(7) & 0.5869(3) & 0.5827(5) & 0.5807(8) & \textcolor{red}{\textbf{0.5884(1)}} & 0.5879(2)
& 0.5513(5) & 0.5520(3) & 0.5492(7) & 0.5487(8) & 0.5503(6) & 0.5517(4) & 0.5531(2) & \textcolor{red}{\textbf{0.5539(1)}}
& 0.5576(7) & 0.5629(4) & 0.5649(2) & 0.5592(6) & 0.5606(5) & 0.5574(8) & 0.5640(3) & \textcolor{red}{\textbf{0.5657(1)}} \\

birds
& 0.8026(6) & 0.8037(5) & 0.8006(8) & 0.8025(7) & 0.8046(3) & 0.8045(4) & 0.8050(2) & \textcolor{red}{\textbf{0.8090(1)}}
& 0.6670(6) & 0.6763(2) & 0.6662(8) & 0.6667(7) & \textcolor{red}{\textbf{0.6782(1)}} & 0.6683(5) & 0.6705(4) & 0.6751(3)
& 0.6522(7) & 0.6603(5) & 0.6584(6) & 0.6522(7) & 0.6633(2) & 0.6625(3) & 0.6621(4) & \textcolor{red}{\textbf{0.6649(1)}} \\

enron
& 0.7704(7) & 0.7751(4) & 0.7748(5) & 0.7729(6) & 0.7757(2) & 0.7695(8) & 0.7753(3) & \textcolor{red}{\textbf{0.7766(1)}}
& 0.7572(7) & 0.7627(4) & 0.7538(8) & 0.7600(5) & 0.7630(3) & 0.7599(6) & 0.7664(2) & \textcolor{red}{\textbf{0.7667(1)}}
& 0.7617(3) & 0.7506(8) & 0.7573(6) & 0.7608(5) & \textcolor{red}{\textbf{0.7674(1)}} & 0.7611(4) & 0.7652(2) & 0.7560(7) \\

scene
& 0.9481(8) & 0.9496(6) & 0.9489(7) & \textcolor{red}{\textbf{0.9519(1)}} & 0.9504(3) & 0.9503(5) & 0.9504(3) & 0.9510(2)
& 0.9495(7) & 0.9508(4) & 0.9510(3) & 0.9502(5) & 0.9497(6) & 0.9491(8) & 0.9513(2) & \textcolor{red}{\textbf{0.9515(1)}}
& 0.9516(5) & 0.9514(6) & 0.9521(2) & 0.9503(8) & \textcolor{red}{\textbf{0.9529(1)}} & 0.9507(7) & 0.9520(3) & 0.9520(3) \\

yeast
& 0.7226(7) & 0.7236(3) & 0.7230(5) & 0.7217(8) & 0.7234(4) & 0.7229(6) & 0.7242(2) & \textcolor{red}{\textbf{0.7305(1)}}
& 0.7028(6) & 0.7007(8) & 0.7048(2) & 0.7033(5) & \textcolor{red}{\textbf{0.7081(1)}} & 0.7017(7) & 0.7048(2) & 0.7047(4)
& 0.7090(8) & 0.7150(2) & 0.7129(5) & 0.7122(7) & 0.7150(2) & 0.7127(6) & 0.7143(4) & \textcolor{red}{\textbf{0.7158(1)}} \\

Corel5k
& 0.7582(8) & 0.7603(7) & 0.7614(5) & 0.7629(2) & 0.7619(4) & 0.7606(6) & 0.7621(3) & \textcolor{red}{\textbf{0.7647(1)}}
& 0.7512(5) & 0.7524(3) & 0.7491(8) & 0.7508(6) & 0.7494(7) & 0.7517(4) & 0.7539(2) & \textcolor{red}{\textbf{0.7542(1)}}
& 0.7396(8) & 0.7460(3) & 0.7458(4) & 0.7436(7) & 0.7485(2) & 0.7450(5) & 0.7449(6) & \textcolor{red}{\textbf{0.7507(1)}} \\

rcvsubset1
& 0.9243(8) & 0.9257(4) & 0.9256(5) & 0.9269(2) & 0.9250(7) & 0.9259(3) & 0.9254(6) & \textcolor{red}{\textbf{0.9271(1)}}
& 0.9208(6) & 0.9209(5) & 0.9211(4) & 0.9200(8) & 0.9205(7) & \textcolor{red}{\textbf{0.9226(1)}} & 0.9216(3) & \textcolor{red}{\textbf{0.9226(1)}}
& 0.9321(8) & 0.9322(7) & 0.9348(2) & 0.9337(3) & 0.9331(6) & 0.9333(5) & 0.9335(4) & \textcolor{red}{\textbf{0.9351(1)}} \\

rcvsubset2
& 0.9211(8) & 0.9221(6) & 0.9222(5) & 0.9232(2) & 0.9223(3) & 0.9217(7) & 0.9223(3) & \textcolor{red}{\textbf{0.9238(1)}}
& 0.9180(8) & 0.9196(5) & 0.9188(7) & \textcolor{red}{\textbf{0.9219(1)}} & 0.9191(6) & 0.9211(3) & 0.9206(4) & 0.9217(2)
& 0.9318(2) & 0.9290(6) & 0.9277(8) & 0.9306(3) & 0.9284(7) & 0.9299(5) & 0.9300(4) & \textcolor{red}{\textbf{0.9346(1)}} \\

rcvsubset3
& 0.9168(8) & 0.9176(6) & 0.9188(4) & \textcolor{red}{\textbf{0.9220(1)}} & 0.9185(5) & 0.9172(7) & 0.9205(3) & 0.9211(2)
& 0.9129(6) & 0.9132(5) & 0.9113(8) & 0.9143(2) & 0.9123(7) & 0.9135(4) & 0.9137(3) & \textcolor{red}{\textbf{0.9147(1)}}
& 0.9263(8) & 0.9301(3) & 0.9299(4) & 0.9305(2) & 0.9278(6) & 0.9279(5) & 0.9272(7) & \textcolor{red}{\textbf{0.9326(1)}} \\

bibtex
& 0.9031(8) & 0.9036(4) & 0.9034(6) & 0.9045(2) & 0.9035(5) & 0.9037(3) & 0.9034(6) & \textcolor{red}{\textbf{0.9061(1)}}
& 0.9061(8) & 0.9065(7) & 0.9075(4) & 0.9084(2) & 0.9076(3) & \textcolor{red}{\textbf{0.9085(1)}} & 0.9074(5) & 0.9074(5)
& \textcolor{red}{\textbf{0.9031(1)}} & 0.9003(5) & 0.8998(6) & 0.8995(7) & 0.8991(8) & 0.9013(3) & 0.9014(2) & 0.9005(4) \\

yahoo-Arts1
& 0.7441(7) & 0.7438(8) & 0.7479(5) & 0.7495(2) & 0.7465(6) & 0.7482(4) & 0.7486(3) & \textcolor{red}{\textbf{0.7512(1)}}
& 0.7315(7) & 0.7371(5) & 0.7387(3) & 0.7277(8) & 0.7352(6) & \textcolor{red}{\textbf{0.7407(1)}} & 0.7373(4) & 0.7394(2)
& \textcolor{red}{\textbf{0.7358(1)}} & 0.7219(8) & 0.7236(7) & 0.7310(3) & 0.7260(6) & 0.7282(4) & 0.7264(5) & 0.7319(2) \\

yahoo-Business1
& 0.8063(7) & 0.8070(5) & 0.8047(8) & \textcolor{red}{\textbf{0.8115(1)}} & 0.8076(3) & 0.8064(6) & 0.8072(4) & \textcolor{red}{\textbf{0.8115(1)}}
& 0.7870(7) & 0.7891(5) & 0.7945(3) & 0.7905(4) & 0.7860(8) & 0.7890(6) & \textcolor{red}{\textbf{0.7974(1)}} & 0.7958(2)
& 0.7688(7) & 0.7755(2) & 0.7753(3) & 0.7693(6) & 0.7643(8) & 0.7712(5) & \textcolor{red}{\textbf{0.7795(1)}} & 0.7749(4) \\

mediamill
& 0.8025(6) & 0.8012(8) & 0.8027(5) & 0.8043(3) & 0.8028(4) & \textcolor{red}{\textbf{0.8129(1)}} & 0.8018(7) & 0.8108(2)
& 0.8610(6) & 0.8606(7) & 0.8611(5) & 0.8568(8) & 0.8619(3) & \textcolor{red}{\textbf{0.8637(1)}} & 0.8617(4) & 0.8624(2)
& 0.8689(4) & 0.8699(2) & 0.8681(6) & 0.8682(5) & 0.8694(3) & 0.8673(8) & 0.8677(7) & \textcolor{red}{\textbf{0.8701(1)}} \\

\midrule
Avg(Rank)
& 7.08 & 5.54 & 5.77 & 3.08 & 4.15 & 5.23 & 3.54 & \textcolor{red}{\textbf{1.31}}
& 6.46 & 4.85 & 5.38 & 5.31 & 4.92 & 3.92 & 2.92 & \textcolor{red}{\textbf{2.00}}
& 5.31 & 4.69 & 4.69 & 5.31 & 4.38 & 5.23 & 4.00 & \textcolor{red}{\textbf{2.15}} \\
\bottomrule
\end{tabular}}
\caption{The Macro-AUC results of three base models using various batch selection methods on multi-label tabular datasets.
} 
\label{ta:Macro-AUC}
\end{table*}


\begin{table*}[!t]
\centering
\resizebox{\textwidth}{!}{
\begin{tabular}{ccccccccc|cccccccc|cccccccc}
\toprule
\multirow{2}{*}{Dataset} & \multicolumn{8}{c}{CLIF} & \multicolumn{8}{c}{DELA} & \multicolumn{8}{c}{PACA} \\ 
\cmidrule(lr){2-9} \cmidrule(lr){10-17} \cmidrule(lr){18-25}
& Random & Active & Recent & DIHCL & Balance & Hard-Imb & ML-Unc & \textbf{D2ACE} & Random & Active & Recent & DIHCL & Balance & Hard-Imb & ML-Unc & \textbf{D2ACE} & Random & Active & Recent & DIHCL & Balance & Hard-Imb & ML-Unc & \textbf{D2ACE} \\
\midrule
cal500 & 0.0996(5) & 0.1057(2) & 0.0977(6) & 0.0977(6) & 0.1018(3) & 0.0967(8) & 0.1005(4) & \textcolor{red}{\textbf{0.1161(1)}}
& 0.0590(8) & 0.0596(6) & 0.0627(2) & 0.0601(5) & 0.0607(4) & 0.0622(3) & 0.0593(7) & \textcolor{red}{\textbf{0.0658(1)}}
& 0.0709(6) & 0.0678(7) & 0.0640(8) & 0.0718(3) & 0.0714(5) & 0.0758(2) & 0.0715(4) & \textcolor{red}{\textbf{0.0761(1)}} \\

birds & 0.3251(5) & 0.3290(4) & 0.3133(8) & 0.3298(3) & 0.3247(6) & \textcolor{red}{\textbf{0.3611(1)}} & 0.3244(7) & 0.3560(2)
& 0.0700(2) & 0.0645(6) & \textcolor{red}{\textbf{0.0719(1)}} & 0.0652(5) & 0.0596(7) & 0.0659(4) & 0.0585(8) & 0.0698(3)
& 0.0158(8) & 0.0179(7) & 0.0285(3) & 0.0221(6) & 0.0223(5) & 0.0306(2) & 0.0252(4) & \textcolor{red}{\textbf{0.0372(1)}} \\

enron & 0.3011(7) & 0.3057(3) & 0.3015(5) & \textcolor{red}{\textbf{0.3104(1)}} & 0.2982(8) & 0.3013(6) & 0.3022(4) & 0.3068(2)
& 0.2370(5) & 0.2364(7) & \textcolor{red}{\textbf{0.2432(1)}} & 0.2364(7) & 0.2368(6) & 0.2402(4) & 0.2403(3) & 0.2407(2)
& 0.2832(5) & 0.2820(7) & 0.2819(8) & 0.2842(4) & 0.2907(3) & 0.2951(2) & 0.2825(6) & \textcolor{red}{\textbf{0.2986(1)}} \\

scene & 0.7834(8) & 0.7860(4) & 0.7855(6) & 0.7893(2) & 0.7846(7) & 0.7859(5) & 0.7861(3) & \textcolor{red}{\textbf{0.7931(1)}}
& 0.7829(5) & 0.7840(2) & 0.7828(6) & 0.7815(8) & 0.7836(4) & 0.7825(7) & \textcolor{red}{\textbf{0.7890(1)}} & 0.7838(3)
& 0.7818(4) & 0.7877(2) & 0.7770(6) & 0.5300(8) & \textcolor{red}{\textbf{0.7883(1)}} & 0.7765(7) & 0.7813(5) & 0.7862(3) \\

yeast & 0.4251(4) & 0.4193(7) & 0.4263(3) & 0.4234(5) & 0.4195(6) & 0.4176(8) & 0.4266(2) & \textcolor{red}{\textbf{0.4507(1)}}
& 0.3607(7) & 0.3661(4) & 0.3749(2) & 0.3557(8) & 0.3621(6) & 0.3679(3) & 0.3650(5) & \textcolor{red}{\textbf{0.3784(1)}}
& 0.3876(3) & 0.3870(4) & 0.3734(8) & 0.3811(6) & 0.3844(5) & 0.4007(2) & 0.3806(7) & \textcolor{red}{\textbf{0.4015(1)}} \\

Corel5k & 0.0653(7) & 0.0661(6) & 0.0693(2) & \textcolor{red}{\textbf{0.0711(1)}} & 0.0670(5) & 0.0649(8) & 0.0685(4) & 0.0693(2)
& 0.0715(8) & 0.0753(4) & 0.0762(3) & 0.0738(6) & 0.0732(7) & 0.0739(5) & 0.0763(2) & \textcolor{red}{\textbf{0.0789(1)}}
& 0.0390(8) & 0.0391(7) & 0.0401(6) & 0.0402(5) & 0.0463(2) & 0.0463(2) & 0.0432(4) & \textcolor{red}{\textbf{0.0511(1)}} \\

rcvsubset1 & 0.2659(8) & 0.2663(6) & 0.2670(5) & 0.2773(2) & 0.2688(4) & 0.2661(7) & 0.2689(3) & \textcolor{red}{\textbf{0.2836(1)}}
& 0.2789(8) & 0.2803(7) & 0.2847(5) & 0.2882(2) & 0.2833(6) & 0.2856(3) & 0.2851(4) & \textcolor{red}{\textbf{0.2903(1)}}
& 0.2004(7) & 0.1982(8) & 0.2085(5) & 0.2215(4) & \textcolor{red}{\textbf{0.2407(1)}} & 0.2006(6) & 0.2385(3) & 0.2396(2) \\

rcvsubset2 & 0.2503(8) & 0.2509(7) & 0.2667(4) & 0.2703(2) & 0.2538(6) & 0.2584(5) & 0.2670(3) & \textcolor{red}{\textbf{0.2768(1)}}
& 0.2701(8) & 0.2728(6) & 0.2784(2) & 0.2775(3) & 0.2752(5) & 0.2772(4) & 0.2719(7) & \textcolor{red}{\textbf{0.2824(1)}}
& 0.1907(7) & 0.1982(6) & 0.2067(5) & 0.2105(4) & 0.2170(3) & 0.1844(8) & 0.2183(2) & \textcolor{red}{\textbf{0.2195(1)}} \\

rcvsubset3 & 0.2367(8) & 0.2446(3) & 0.2369(7) & 0.2513(2) & 0.2380(5) & 0.2373(6) & 0.2381(4) & \textcolor{red}{\textbf{0.2568(1)}}
& 0.2517(5) & 0.2513(6) & 0.2502(7) & 0.2528(4) & 0.2481(8) & 0.2585(2) & 0.2534(3) & \textcolor{red}{\textbf{0.2663(1)}}
& 0.1942(5) & 0.1907(8) & 0.1941(6) & 0.1976(4) & 0.1989(3) & 0.1938(7) & \textcolor{red}{\textbf{0.2600(1)}} & 0.2325(2) \\

bibtex & 0.2341(8) & 0.2410(6) & 0.2460(4) & 0.2445(5) & 0.2391(7) & \textcolor{red}{\textbf{0.2598(1)}} & 0.2466(3) & 0.2500(2)
& 0.2888(8) & 0.2903(7) & 0.2932(5) & 0.2928(6) & 0.2952(2) & 0.2936(4) & \textcolor{red}{\textbf{0.2974(1)}} & 0.2951(3)
& 0.3066(8) & \textcolor{red}{\textbf{0.3166(1)}} & 0.3148(4) & 0.3095(6) & 0.3105(5) & 0.3079(7) & 0.3161(3) & 0.3165(2) \\

yahoo-Arts1 & 0.2801(7) & 0.2788(8) & 0.2821(4) & \textcolor{red}{\textbf{0.2831(1)}} & 0.2820(5) & 0.2824(3) & 0.2806(6) & 0.2825(2)
& 0.2387(7) & 0.2487(2) & \textcolor{red}{\textbf{0.2594(1)}} & 0.2217(8) & 0.2418(3) & 0.2411(4) & 0.2400(6) & 0.2401(5)
& \textcolor{red}{\textbf{0.2573(1)}} & 0.2511(2) & 0.2451(5) & 0.2330(8) & 0.2373(6) & 0.2476(3) & 0.2332(7) & 0.2468(4) \\

yahoo-Business1 & 0.3599(6) & 0.3535(7) & 0.3679(3) & 0.3535(7) & 0.3680(2) & 0.3658(4) & \textcolor{red}{\textbf{0.3682(1)}} & 0.3654(5)
& 0.2751(8) & 0.2977(2) & 0.2805(6) & 0.2762(7) & 0.2824(5) & 0.2912(3) & 0.2832(4) & \textcolor{red}{\textbf{0.3028(1)}}
& 0.2843(7) & 0.2894(6) & 0.3033(4) & 0.2825(8) & 0.2906(5) & 0.3101(3) & 0.3149(2) & \textcolor{red}{\textbf{0.3163(1)}} \\

mediamill & 0.0482(5) & 0.0486(3) & 0.0485(4) & 0.0504(2) & 0.0482(5) & 0.0481(7) & 0.0461(8) & \textcolor{red}{\textbf{0.0506(1)}}
& 0.1012(5) & 0.1053(3) & \textcolor{red}{\textbf{0.1102(1)}} & 0.0939(8) & 0.0979(7) & 0.0995(6) & 0.1049(4) & 0.1082(2)
& 0.1463(3) & 0.1395(7) & 0.1405(6) & 0.1451(4) & 0.1432(5) & \textcolor{red}{\textbf{0.1550(1)}} & 0.1318(8) & 0.1479(2) \\ 
\midrule
Avg(Rank) & 6.62 & 5.08 & 4.69 & 3.00 & 5.31 & 5.31 & 4.00 & \textcolor{red}{\textbf{1.69}} 
& 6.46 & 4.77 & 3.23 & 5.92 & 5.38 & 4.00 & 4.23 & \textcolor{red}{\textbf{1.92}} 
& 5.54 & 5.54 & 5.69 & 5.38 & 3.77 & 4.00 & 4.31 & \textcolor{red}{\textbf{1.69}} \\
\bottomrule
\end{tabular}}
\caption{The Macro-F1 results of three base models using various batch selection methods on multi-label tabular datasets.
}  
\label{ta:Macro-F1}
\end{table*}


\begin{table*}[!t]
\centering
\resizebox{\textwidth}{!}{
\begin{tabular}{ccccccccc|cccccccc|cccccccc}
\toprule
\multirow{2}{*}{Dataset} & \multicolumn{8}{c}{CLIF} & \multicolumn{8}{c}{DELA} & \multicolumn{8}{c}{PACA} \\ 
\cmidrule(lr){2-9} \cmidrule(lr){10-17} \cmidrule(lr){18-25}
& Random & Active & Recent & DIHCL & Balance & Hard-Imb & ML-Unc & \textbf{D2ACE} & Random & Active & Recent & DIHCL & Balance & Hard-Imb & ML-Unc & \textbf{D2ACE} & Random & Active & Recent & DIHCL & Balance & Hard-Imb & ML-Unc  & \textbf{D2ACE} \\ 
\midrule
cal500 & 0.2318(6) & 0.2313(4) & 0.2317(5) &  \textcolor{red}{\textbf{0.2300(1)}} & 0.2310(2) & 0.2318(6) & 0.2311(3) & 0.2318(6)
& 0.2349(5) & 0.2351(6) & 0.2356(8) & 0.2353(7) & \textcolor{red}{\textbf{0.2339(1)}} & 0.2345(3) & 0.2346(4) & \textcolor{red}{\textbf{0.2339(1)}}
& 0.2353(8) & 0.2324(2) & 0.2342(6) & 0.2339(5) & 0.2347(7) & 0.2336(4) & 0.2330(3) & \textcolor{red}{\textbf{0.2322(1)}} \\

birds & 0.1777(7) & 0.1710(2) & 0.1822(8) & 0.1773(6) & 0.1728(4) & 0.1739(5) &  \textcolor{red}{\textbf{0.1705(1)}} & 0.1720(3)
& \textcolor{red}{\textbf{0.2774(1)}} & 0.2857(5) & 0.2862(6) & 0.2820(3) & 0.2914(8) & 0.2820(3) & 0.2901(7) & 0.2792(2)
& 0.3166(3) & 0.3291(8) & 0.3228(5) & 0.3233(7) & 0.3167(4) & 0.3160(2) & \textcolor{red}{\textbf{0.3153(1)}} & 0.3229(6) \\

enron & 0.1205(5) & 0.1213(7) & 0.1209(6) &  \textcolor{red}{\textbf{0.1146(1)}} & 0.1242(8) & 0.1166(2) & 0.1186(4) & 0.1170(3)
& 0.1161(6) & 0.1138(4) & 0.1185(7) & 0.1127(2) & 0.1187(8) & 0.1130(3) & 0.1159(5) & \textcolor{red}{\textbf{0.1126(1)}}
& 0.1319(7) & 0.1314(6) & 0.1307(4) & 0.1276(2) & 0.1280(3) & 0.1308(5) & 0.1346(8) & \textcolor{red}{\textbf{0.1261(1)}} \\

scene & 0.0671(7) & 0.0669(6) & 0.0665(5) & 0.0650(2) & 0.0661(3) & 0.0671(7) & 0.0664(4) & \textcolor{red}{\textbf{0.0629(1)}}
& 0.0650(6) & 0.0655(7) & 0.0647(5) & 0.0641(3) & 0.0670(8) & 0.0644(4) & 0.0633(2) & \textcolor{red}{\textbf{0.0630(1)}}
& 0.0697(8) & 0.0663(7) & 0.0645(2) & 0.0645(2) & \textcolor{red}{\textbf{0.0636(1)}} & 0.0647(5) & 0.0659(6) & 0.0646(4) \\

yeast & 0.1654(5) & 0.1648(4) & 0.1670(7) & 0.1682(8) & 0.1646(3) & 0.1658(6) & 0.1645(2) & \textcolor{red}{\textbf{0.1616(1)}}
& 0.1647(5) & 0.1639(3) & 0.1636(2) & 0.1670(8) & 0.1639(3) & 0.1656(6) & 0.1658(7) & \textcolor{red}{\textbf{0.1628(1)}}
& 0.1650(5) & 0.1683(7) & 0.1685(8) & 0.1649(4) & 0.1664(6) & 0.1644(2) & 0.1645(3) & \textcolor{red}{\textbf{0.1638(1)}} \\

Corel5k & 0.1590(8) & 0.1584(5) &  0.1564(2) & 0.1589(6) & 0.1576(4) & 0.1589(6) & 0.1568(3) & \textcolor{red}{\textbf{0.1544(1)}}
& 0.1572(5) & 0.1554(2) & 0.1587(8) & 0.1568(4) & 0.1582(7) & 0.1561(3) & 0.1574(6) & \textcolor{red}{\textbf{0.1543(1)}}
& 0.1663(8) & 0.1627(5) & 0.1638(7) & 0.1635(6) & 0.1625(3) & 0.1626(4) & 0.1618(2) & \textcolor{red}{\textbf{0.1595(1)}} \\

rcvsubset1 & 0.0668(8) & 0.0664(6) & 0.0655(3) &  0.0641(2) & 0.0667(7) & 0.0662(5) & 0.0655(3) & \textcolor{red}{\textbf{0.0633(1)}}
& 0.0555(6) & 0.0553(4) & 0.0551(3) & 0.0569(8) & 0.0557(7) & 0.0554(5) & 0.0549(2) & \textcolor{red}{\textbf{0.0542(1)}}
& 0.0469(6) & 0.0469(6) & 0.0466(3) & 0.4662(8) & 0.0467(5) & \textcolor{red}{\textbf{0.0458(1)}} & 0.0466(3) & 0.0460(2) \\

rcvsubset2 & 0.0719(8) & 0.0709(7) & 0.0694(2) & 0.0696(4) & 0.0696(4) & 0.0698(6) & 0.0694(2) & \textcolor{red}{\textbf{0.0675(1)}}
& 0.0621(8) & 0.0615(6) & 0.0596(2) & \textcolor{red}{\textbf{0.0593(1)}} & 0.0615(6) & 0.0600(3) & 0.0614(5) & 0.0603(4)
& 0.0523(2) & 0.0525(3) & 0.0551(8) & 0.0532(6) & 0.0544(7) & 0.0526(4) & 0.0528(5) & \textcolor{red}{\textbf{0.0505(1)}} \\

rcvsubset3 & 0.0738(6) & 0.0743(8) & 0.0739(7) & 0.0705(2) & 0.0728(3) & 0.0736(5) & 0.0733(4) & \textcolor{red}{\textbf{0.0697(1)}}
& 0.0675(5) & 0.0679(6) & 0.0687(7) & 0.0673(4) & 0.0663(3) & 0.0661(2) & 0.0695(8) & \textcolor{red}{\textbf{0.0656(1)}}
& 0.0544(7) & 0.0536(5) & 0.0558(8) & 0.0530(3) & 0.0541(6) & 0.0515(2) & 0.0533(4) & \textcolor{red}{\textbf{0.0498(1)}} \\

bibtex & 0.0879(8) & 0.0862(7) & 0.0861(6) & 0.0857(3) & 0.0858(5) & 0.0857(3) & 0.0855(2) & \textcolor{red}{\textbf{0.0839(1)}}
& 0.0789(8) & 0.0786(6) & 0.0787(7) & 0.0784(5) & \textcolor{red}{\textbf{0.0767(1)}} & 0.0775(3) & 0.0778(4) & 0.0773(2)
& 0.0795(8) & 0.0780(2) & 0.0786(4) & 0.0781(3) & 0.0794(6) & 0.0787(5) & 0.0794(6) & \textcolor{red}{\textbf{0.0773(1)}} \\

yahoo-Arts1 & 0.1578(8) & 0.1572(7) & 0.1569(5) & 0.1565(4) & 0.1564(3) & 0.1569(5) & 0.1563(2) & \textcolor{red}{\textbf{0.1528(1)}}
& 0.1525(6) & 0.1512(5) & 0.1504(4) & 0.1572(8) & 0.1495(2) & 0.1527(7) & 0.1495(2) & \textcolor{red}{\textbf{0.1490(1)}}
& \textcolor{red}{\textbf{0.1621(1)}} & 0.1669(6) & 0.1686(8) & 0.1635(4) & 0.1682(7) & 0.1631(3) & 0.1646(5) & 0.1628(2) \\

yahoo-Business1 & 0.1339(6) & 0.1334(4) & 0.1336(5) & 0.1332(2) & 0.1346(8) & 0.1332(2) & 0.1343(7) & \textcolor{red}{\textbf{0.1267(1)}}
& 0.1404(8) & 0.1380(5) & 0.1382(6) & 0.1372(3) & 0.1401(7) & 0.1372(3) & 0.1345(2) & \textcolor{red}{\textbf{0.1327(1)}}
& 0.1583(4) & 0.1688(8) & 0.1566(3) & 0.1611(7) & 0.1608(6) & 0.1599(5) & \textcolor{red}{\textbf{0.1527(1)}} & 0.1530(2) \\

mediamill & 0.0441(7) & 0.0442(8) &  0.0438(2) & 0.0439(5) &  0.0438(2) & 0.0440(6) & 0.0438(2) & \textcolor{red}{\textbf{0.0429(1)}}
& 0.0366(6) & 0.0365(4) & 0.0366(6) & 0.0375(8) & \textcolor{red}{\textbf{0.0363(1)}} & 0.0364(2) & 0.0365(4) & 0.0364(2)
& 0.0358(7) & 0.0352(4) & 0.0354(6) & 0.0358(7) & \textcolor{red}{\textbf{0.0347(1)}} & 0.0350(2) & 0.0351(3) & 0.0352(4) \\ 
\midrule
Avg(Rank) & 6.85 & 5.77 & 4.85 & 3.54 & 4.31 & 4.92 & 3.00 & \textcolor{red}{\textbf{1.69}}
& 5.77 & 4.85 & 5.46 & 4.92 & 4.77 & 3.62 & 4.46 & \textcolor{red}{\textbf{1.46}} 
& 5.69 & 5.31 & 5.54 & 4.92 & 4.77 & 3.38 & 3.85 & \textcolor{red}{\textbf{2.08}} \\
\bottomrule
\end{tabular}}
\caption{The Ranking-loss results of three base models using various batch selection methods on multi-label tabular datasets.
}  
\label{ta:Ranking-loss}
\end{table*}

\begin{table*}[!t]
\centering
\resizebox{\textwidth}{!}{
\begin{tabular}{ccccccccc}
\toprule
Metric & Base Model & Random & Active & Recent & DIHCL & Balance & Hard-Imb & ML-Unc \\
\midrule
\multirow{3}{*}{Macro-AUC}
& CLIF     & \textbf{win} [0.0007] & \textbf{win} [0.0007] & \textbf{win} [0.0007] & \textbf{win} [0.0092] & \textbf{win} [0.0007] & \textbf{win} [0.0020] & \textbf{win} [0.0009] \\
& DELA     & \textbf{win} [0.0007] & \textbf{win} [0.0014] & \textbf{win} [0.0014] & \textbf{win} [0.0018] & \textbf{win} [0.0196] & \textbf{win} [0.0139] & \textbf{win} [0.0139] \\
& PACA     & \textbf{win} [0.0196] & \textbf{win} [0.0016] & \textbf{win} [0.0065] & \textbf{win} [0.0053] & \textbf{win} [0.0196] & \textbf{win} [0.0095] & tie [0.0732] \\
\midrule
\multirow{3}{*}{Macro-F1}
& CLIF     & \textbf{win} [0.0007] & \textbf{win} [0.0007] & \textbf{win} [0.0023] & \textbf{win} [0.0053] & \textbf{win} [0.0018] & \textbf{win} [0.0165] & \textbf{win} [0.0014] \\
& DELA     & \textbf{win} [0.0009] & \textbf{win} [0.0065] & tie [0.0664] & \textbf{win} [0.0007] & \textbf{win} [0.0018] & \textbf{win} [0.0011] & \textbf{win} [0.0059] \\
& PACA     & \textbf{win} [0.0023] & \textbf{win} [0.0028] & \textbf{win} [0.0007] & \textbf{win} [0.0007] & \textbf{win} [0.0014] & \textbf{win} [0.0065] & \textbf{win} [0.0115] \\
\midrule
\multirow{3}{*}{Ranking-Loss}
& CLIF     & \textbf{win} [0.0011] & \textbf{win} [0.0014] & \textbf{win} [0.0009] & \textbf{win} [0.0105] & \textbf{win} [0.0009] & \textbf{win} [0.0014] & \textbf{win} [0.0018] \\
& DELA     & \textbf{win} [0.0028] & \textbf{win} [0.0007] & \textbf{win} [0.0011] & \textbf{win} [0.0011] & \textbf{win} [0.0030] & \textbf{win} [0.0018] & \textbf{win} [0.0007] \\
& PACA     & \textbf{win} [0.0138] & \textbf{win} [0.0011] & \textbf{win} [0.0014] & \textbf{win} [0.0009] & \textbf{win} [0.0196] & \textbf{win} [0.0272] & \textbf{win} [0.0164] \\
\bottomrule
\end{tabular}}
\caption{Statistical significance of performance differences between D2ACE and compared batch selection approaches, evaluated using the Wilcoxon signed-ranks test ($p < 0.05$) across all tabular datasets. Results are shown for each base model and evaluation metric.}
\label{ta:statistical test}
\end{table*}

\section{Supplementary Analysis}

\subsection{Parameter Analysis}

\begin{figure}[t]
  \centering
  \subfigure[cal500]{\includegraphics[width=0.46\columnwidth]{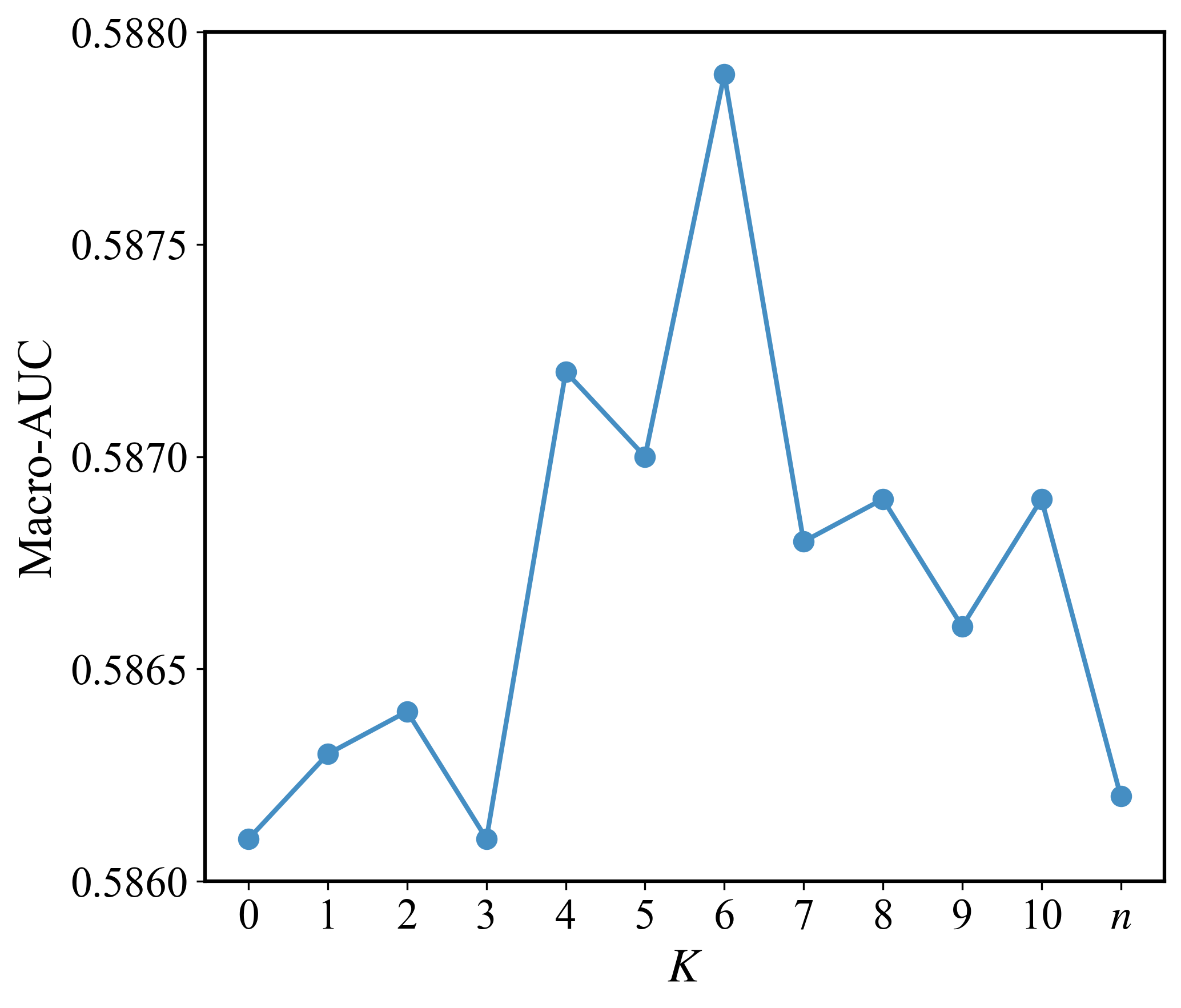}}
  \hspace{0.02\columnwidth}
  \subfigure[bibtex]{\includegraphics[width=0.46\columnwidth]{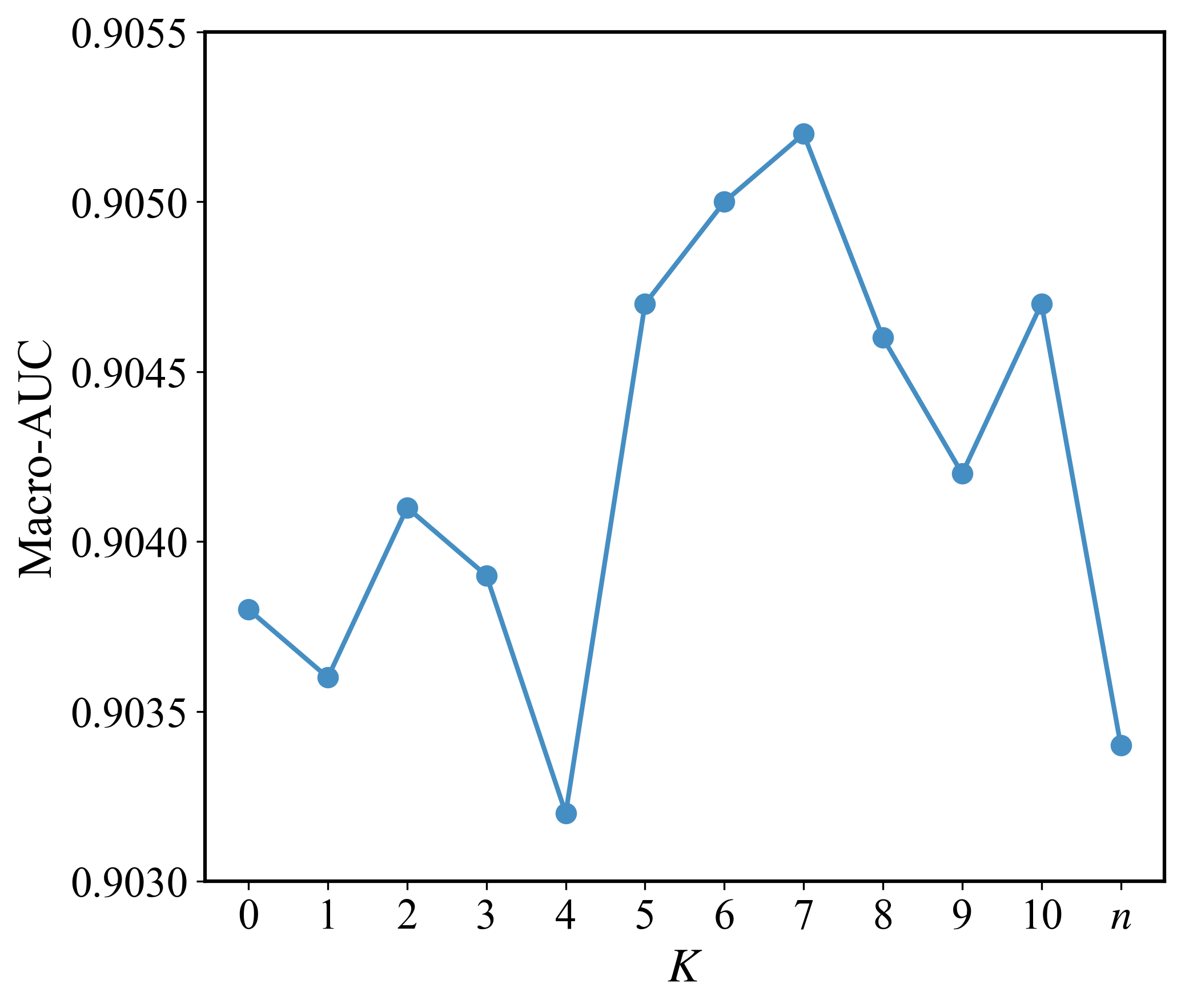}}
  \caption{The Macro-AUC results of CLIF with D2ACE under various neighborhood size $K$.}
  \label{fig:para_k}
\end{figure}

Figure \ref{fig:para_k} presents the sensitivity analysis of the neighborhood size \(K\) on the cal500 and bibtex datasets using CLIF base model.
The results show that performance improves as \(K\) grows from zero, since incorporating more neighbors helps capture a richer set of label correlations within the local context. 
Performance reaches its peak when \(K\) is around 6 to 7, suggesting a good balance between exploiting informative label dependencies and preserving locality. 
However, further increasing \(K\) causes performance degradation, because excessive neighbors cause 
$\mathbf{Z}$ to activate many irrelevant labels, weakening the locality constraint and propagating noisy or less relevant correlations.
Moreover, when $K=n$, meaning the full dataset is treated as the neighborhood (equivalent to the global setting in ML-Unc), performance drops substantially, highlighting the critical role of local context in effective label correlation enhancement.



\begin{figure}[!t]
  \centering
  \subfigure[cal500]{\includegraphics[width=0.46\columnwidth]{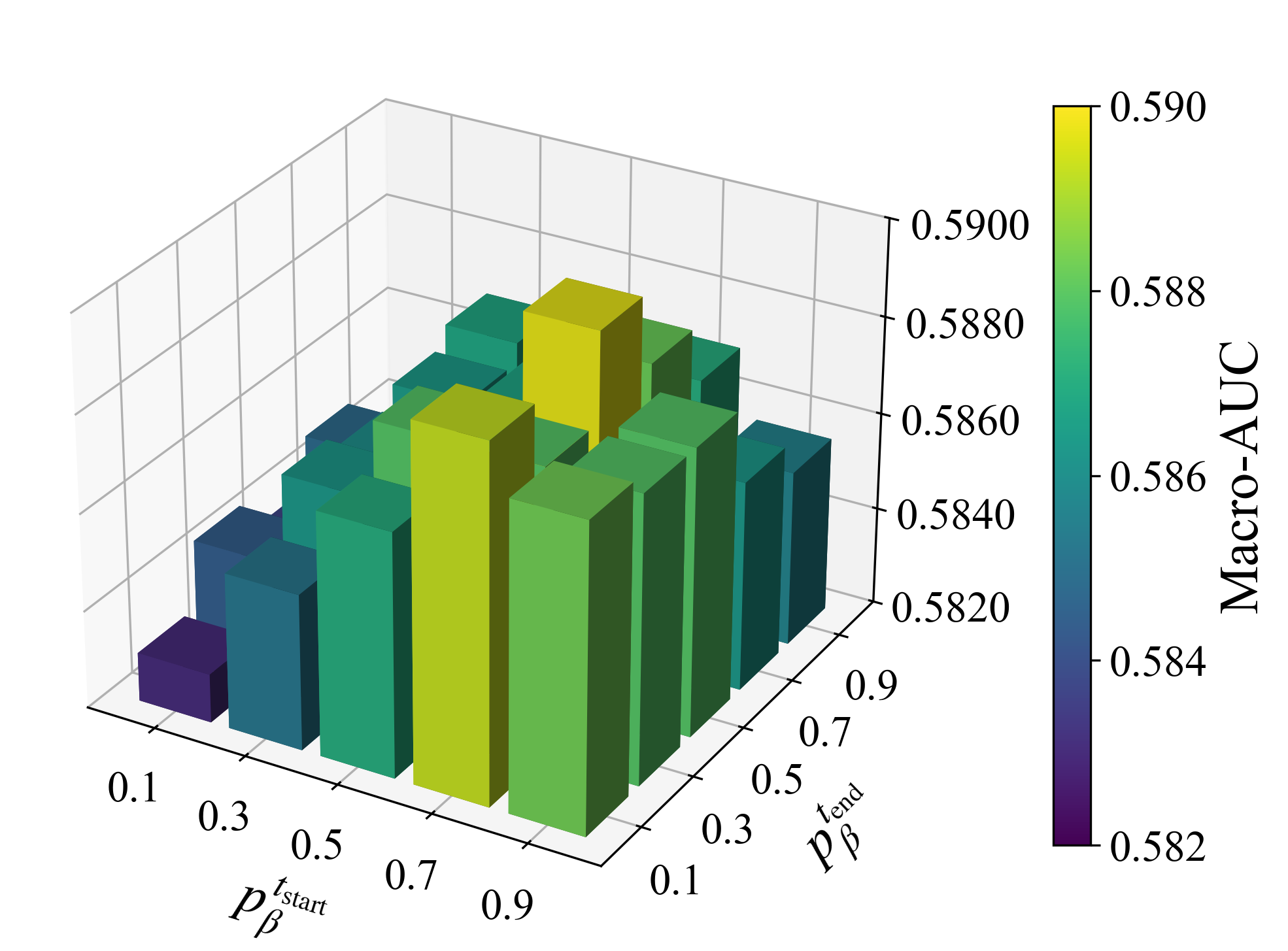}}
  \hspace{0.02\columnwidth}
  \subfigure[bibtex]{\includegraphics[width=0.46\columnwidth]{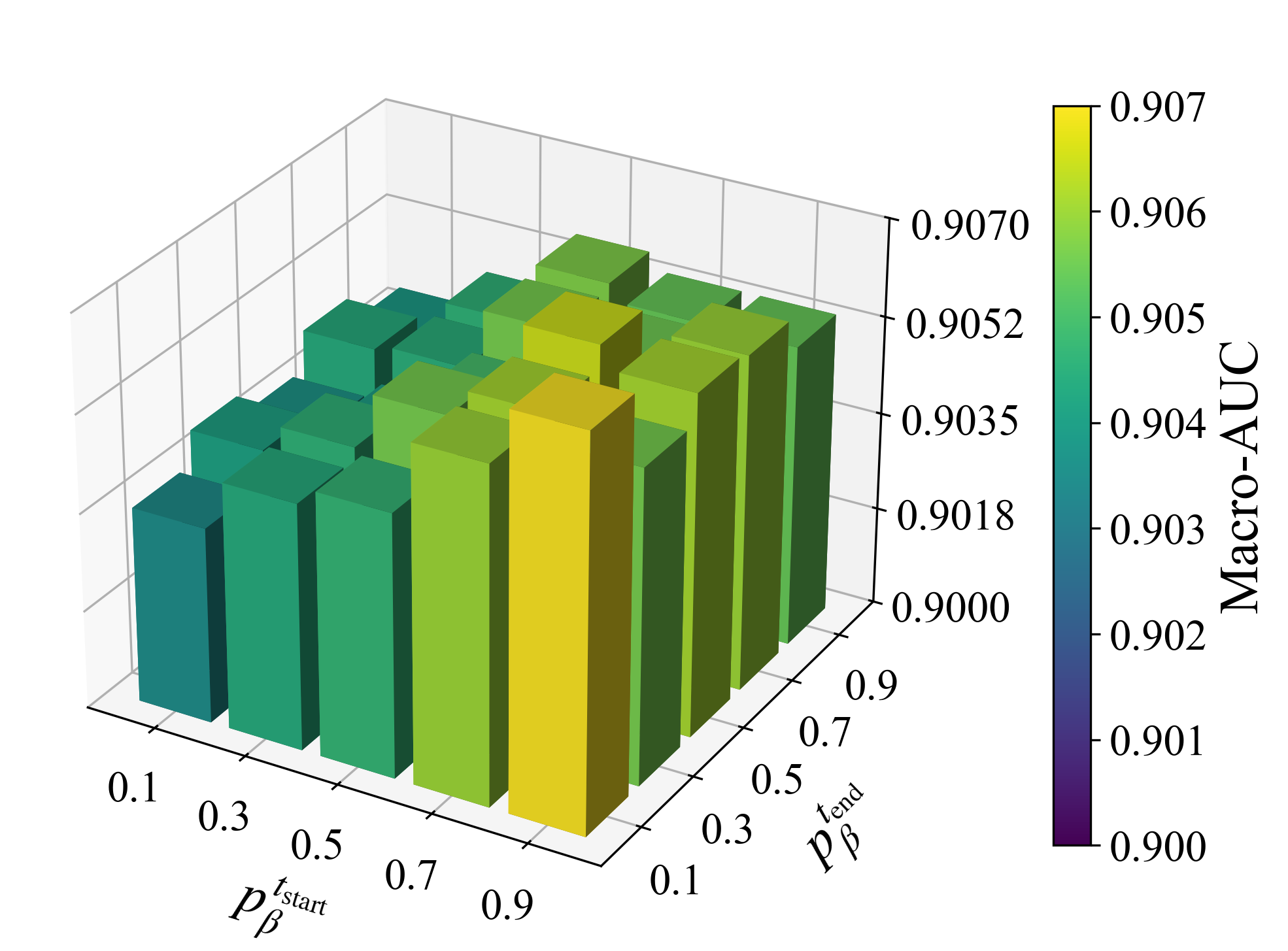}}
  \caption{The Macro-AUC results of CLIP with D2ACE under various uncertainty–hardness mixing coefficient decay schedules (from $p_{\beta}^{t_{\text{start}}}$ to $p_{\beta}^{t_{\text{end}}}$). }
  \label{fig:para_bili}
\end{figure}

\begin{figure}[!t]
  \centering
  \includegraphics[width=0.8\linewidth]{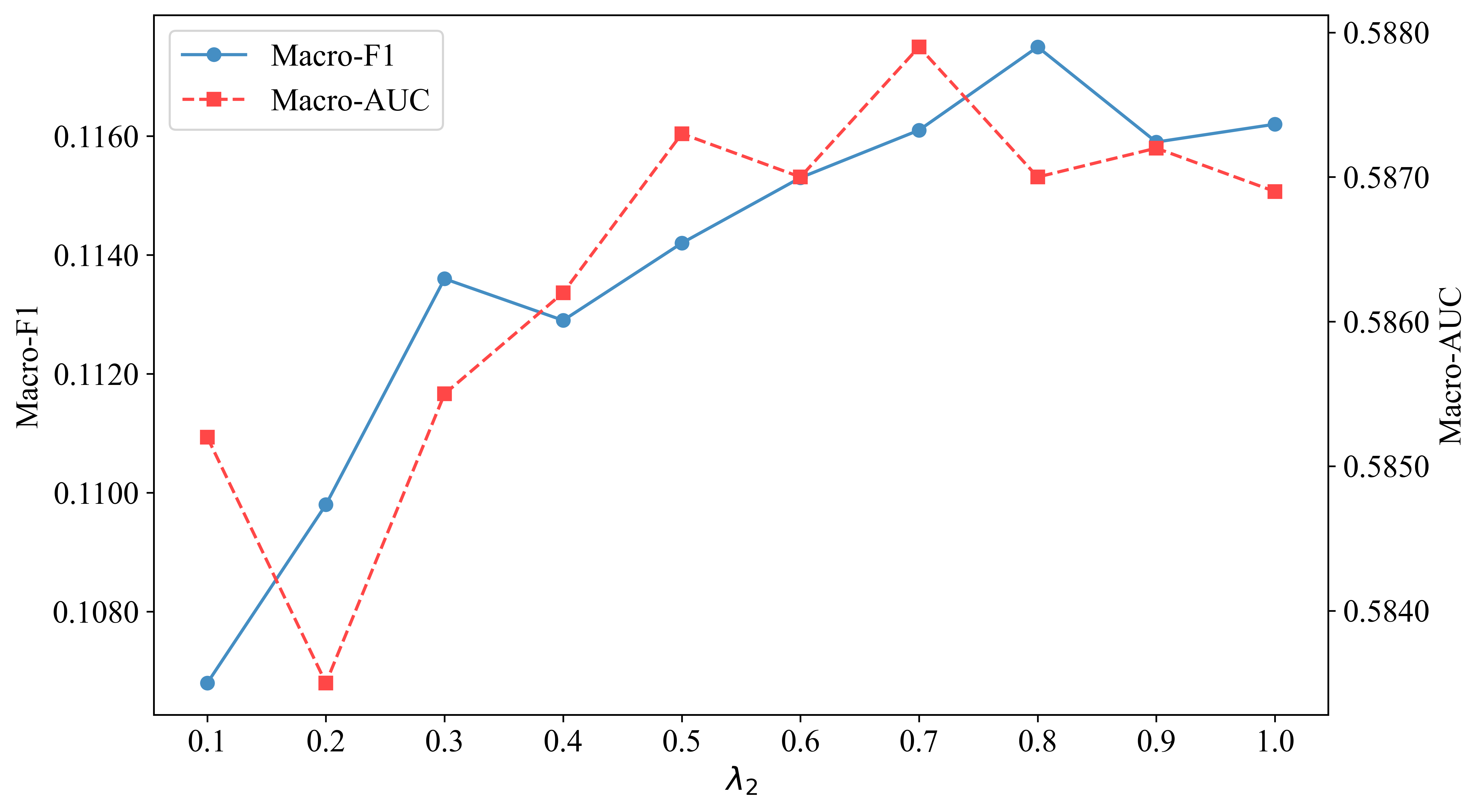}
  \caption{Performance of CLIF with D2ACE under various $\lambda_2$ on the \textit{cal500 }dataset.}
  \label{fig:para_lambda_2}
\end{figure}

Figure \ref{fig:para_bili} shows the effect of different of uncertainty–hardness mixing coefficient decay schedules (from $p_{\beta}^{t_{\text{start}}}$ to $p_{\beta}^{t_{\text{end}}}$) on the cal500 and bibtex datasets using CLIF base model. Each configuration corresponds to a linear decay of the uncertainty sampling probability across training epochs.
Overall, the results show a clear preference for schedules that assign a relatively high initial weight to uncertainty-based sampling ($p_{\beta}^{t_{\text{start}}} = 0.7-0.9$)  and gradually decay it to a lower value in later stages ($p_{\beta}^{t_{\text{end}}}=0.1-0.5$). Such schedules consistently yield higher Macro-AUC on both datasets, while overly low initial uncertainty weights or insufficient decay lead to inferior performance.
This trend aligns with the intention of the Bernoulli mixture sampling strategy. In early training epochs, model predictions are highly unstable, and uncertainty-based sampling encourages exploration of ambiguous regions in the label space, helping the model avoid premature overfitting and improving representation learning. As training progresses, prediction confidence increases, and uncertainty becomes less informative. Thus, gradually shifting emphasis toward hardness-based sampling allows the model to focus on persistently misclassified difficult instances.

Figure \ref{fig:para_lambda_2} further analyzes the EMA smoothing factor $\lambda_2$, where both Macro-F1 and Macro-AUC remain stable and achieve strong performance within $\lambda_2 \in [0.7, 0.9]$, indicating robustness to parameter variations. 

\subsection{Effectiveness of Stage-Wise Bernoulli Mixture Sampling}

\begin{figure*}[!t]
  \centering
    
  \subfigure[Hard-Imb (epoch 20)]{
    \includegraphics[width=0.46\columnwidth]{figure/outlier.png}
    \label{fig:ours_prob_a}
    }
    \hspace{0.02\columnwidth}
  \subfigure[D2ACE (epoch 20)]{
    \includegraphics[width=0.46\columnwidth]{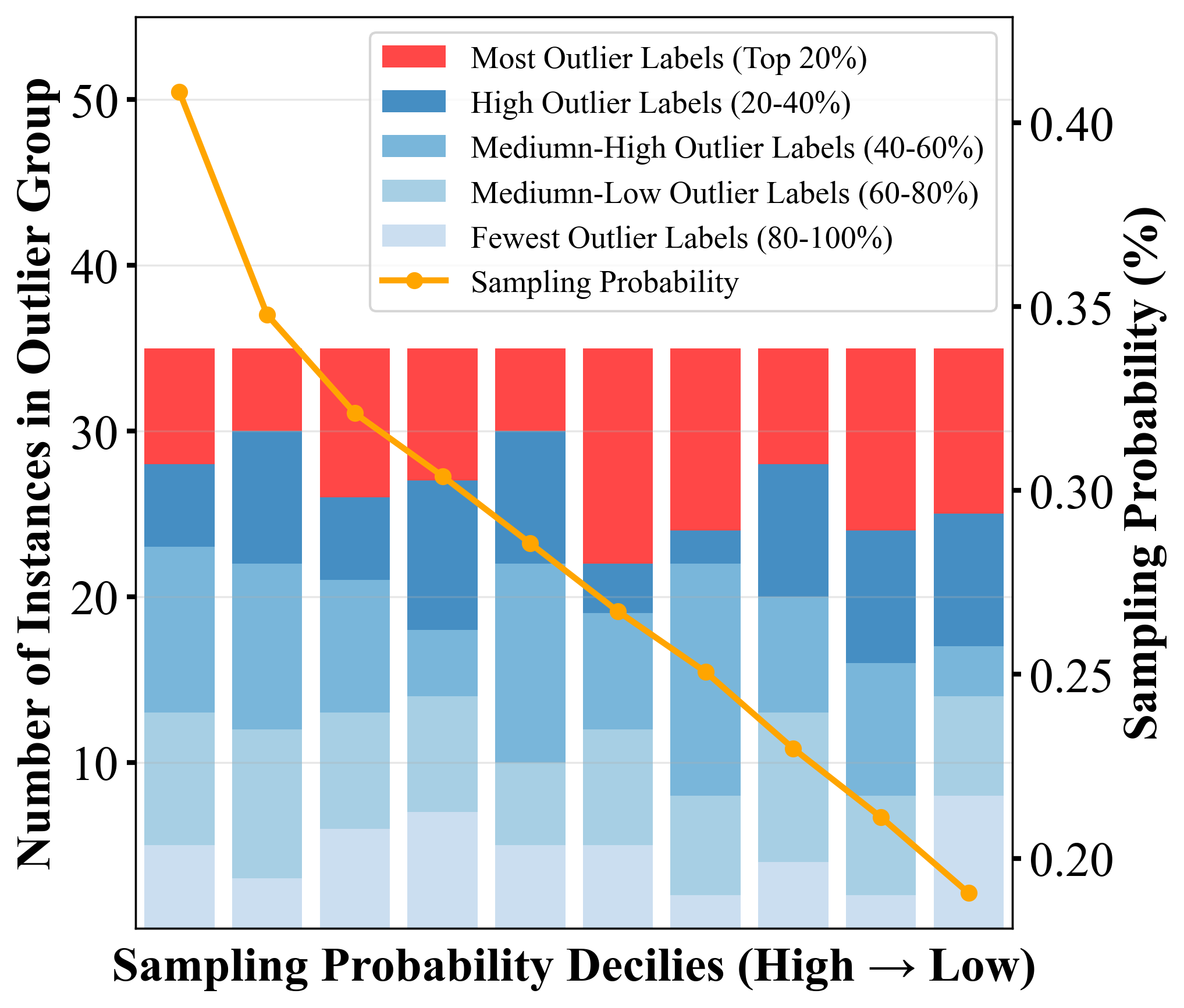}
    \label{fig:ours_prob_b}
  }
    \hspace{0.02\columnwidth}
    \subfigure[ML-Unc (epoch 80)]{
    \includegraphics[width=0.46\columnwidth]{figure/loss_prob.png}
    \label{fig:ours_prob_c}
  }
    \hspace{0.02\columnwidth}
  \subfigure[D2ACE (epoch 80)]{
    \includegraphics[width=0.46\columnwidth]{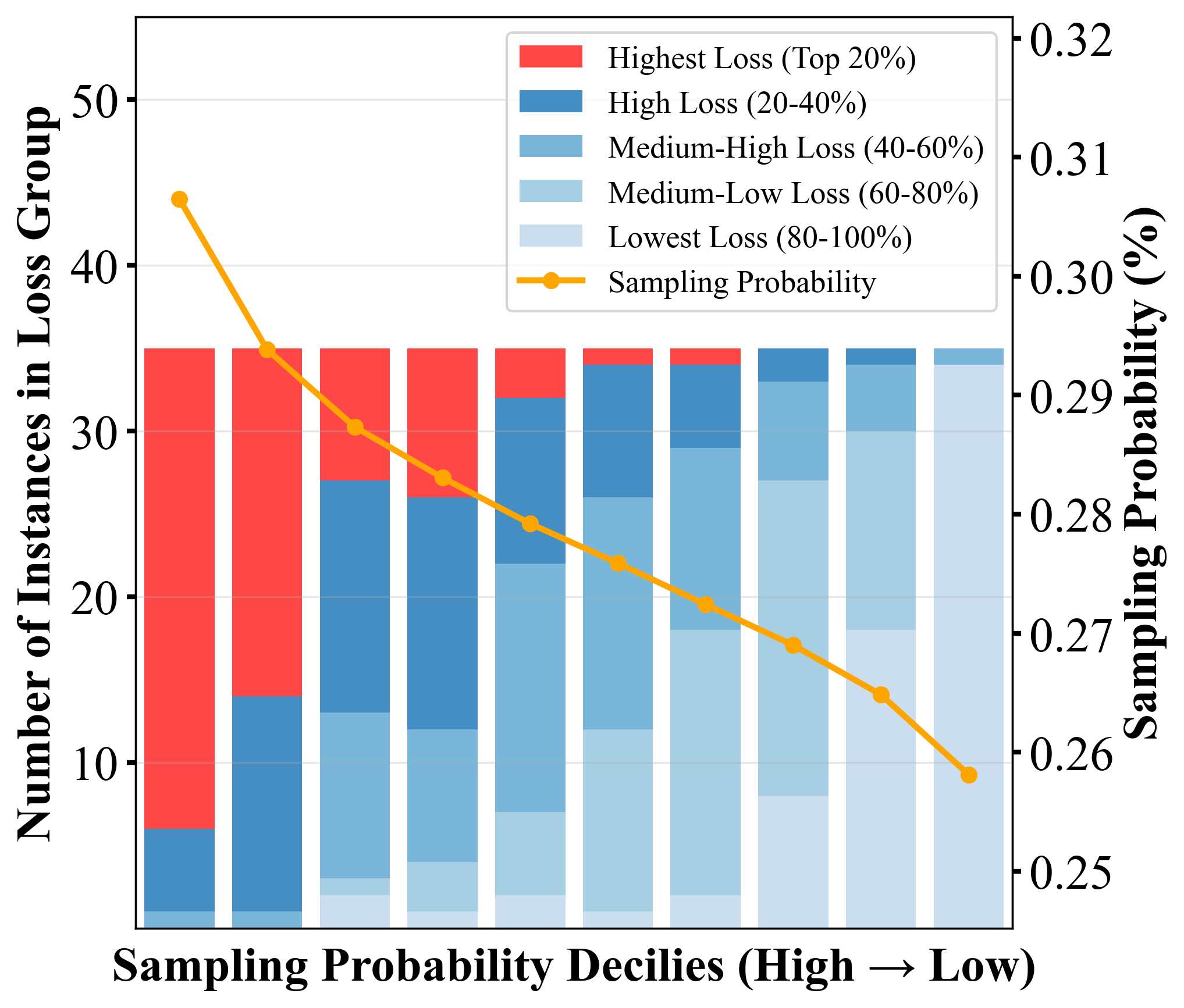}
    \label{fig:ours_prob_d}
  }
    
  \caption{Distribution of instance properties across sampling probability deciles for Hard-Imb, ML-Unc, and D2ACE using CLIF base model on the CAL500 dataset. X-axis shows deciles of instances sorted by descending sampling probability, and color bars indicate counts from each loss/outlier tier within each decile.}
  \label{fig:ours_prob}
\end{figure*}

Figure~\ref{fig:ours_prob} illustrates the distribution of instance properties across sampling probability deciles for Hard-Imb, ML-Unc, and D2ACE at different training stages. Figures~A\ref{fig:ours_prob_a} and A\ref{fig:ours_prob_c} correspond to Figures~1(a) and (b) in the Introduction Section, respectively.

In comparison to Figures~A\ref{fig:ours_prob_a} and A\ref{fig:ours_prob_b}, D2ACE mitigates the excessive focus on outlier samples during early training, thereby reducing the overfitting observed in Hard-Imb. 
In later stages (Figures~A\ref{fig:ours_prob_c} and A\ref{fig:ours_prob_d}), D2ACE successfully prioritizes persistently misclassified hard (high loss) instances, enabling more reliable refinement of complex decision boundaries and correction of systematic errors compared to ML-Unc.


\subsection{Effectiveness of Dynamic Label Weighting}

\begin{figure*}[t]
  \centering
  \subfigure[Balance]{
  \includegraphics[width=0.65\columnwidth]{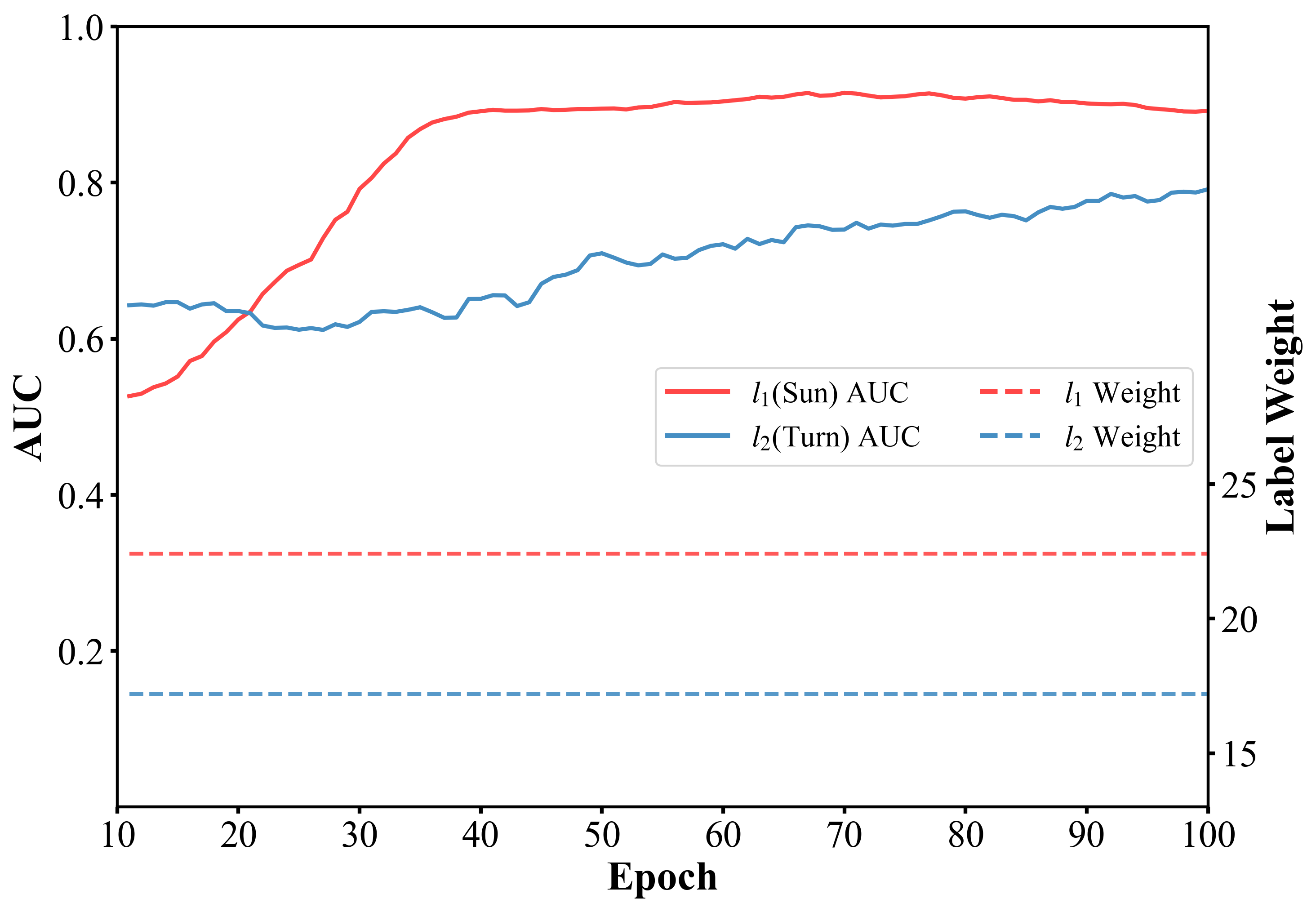}
  \label{fig:labelweight_auc_a}
  }
 \hspace{0.01\columnwidth}
  \subfigure[Hard-Imb]{
  \includegraphics[width=0.65\columnwidth]{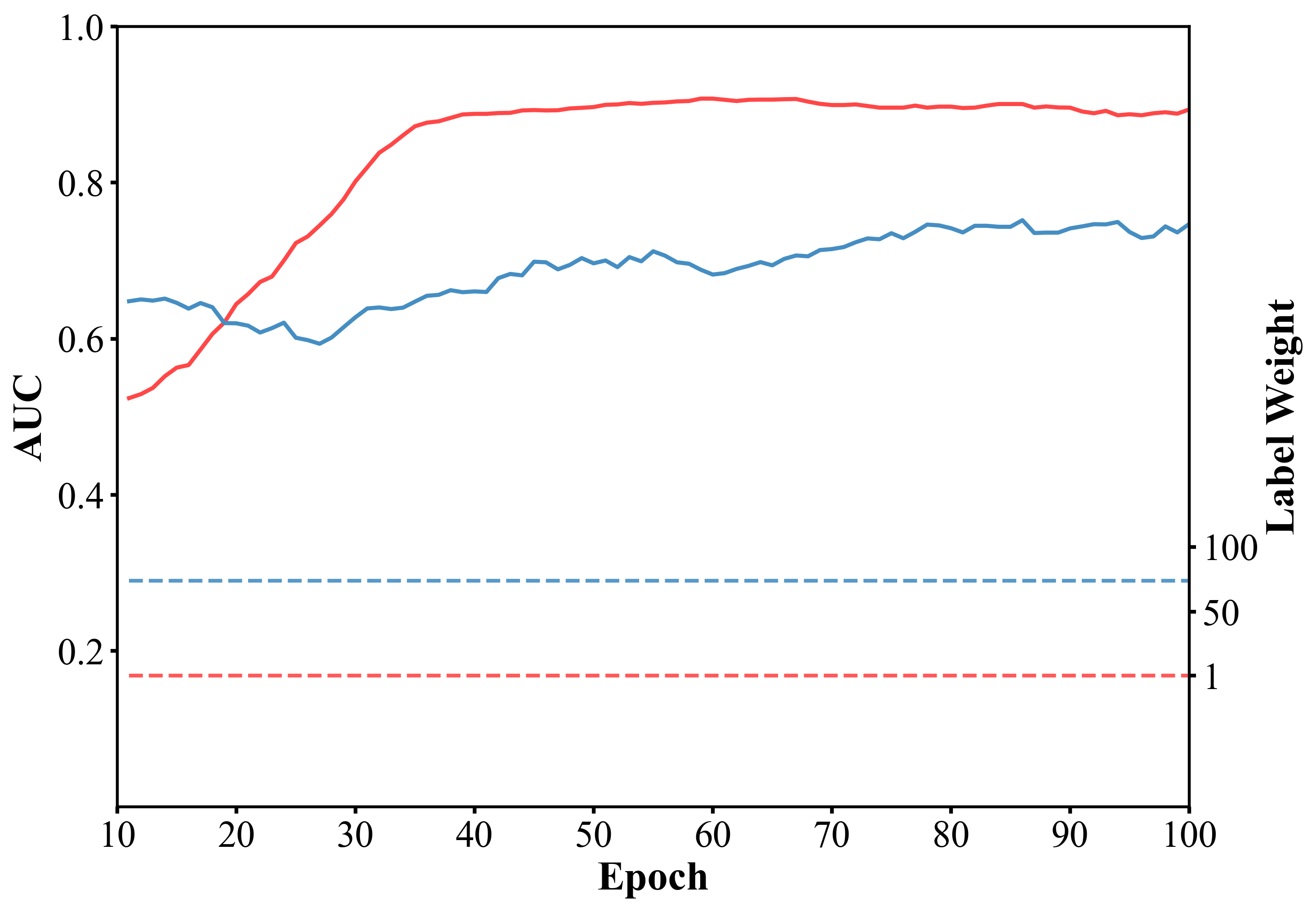}
  \label{fig:labelweight_auc_b}
  }
   \hspace{0.01\columnwidth}
   \subfigure[D2ACE]{
   \includegraphics[width=0.65\columnwidth]{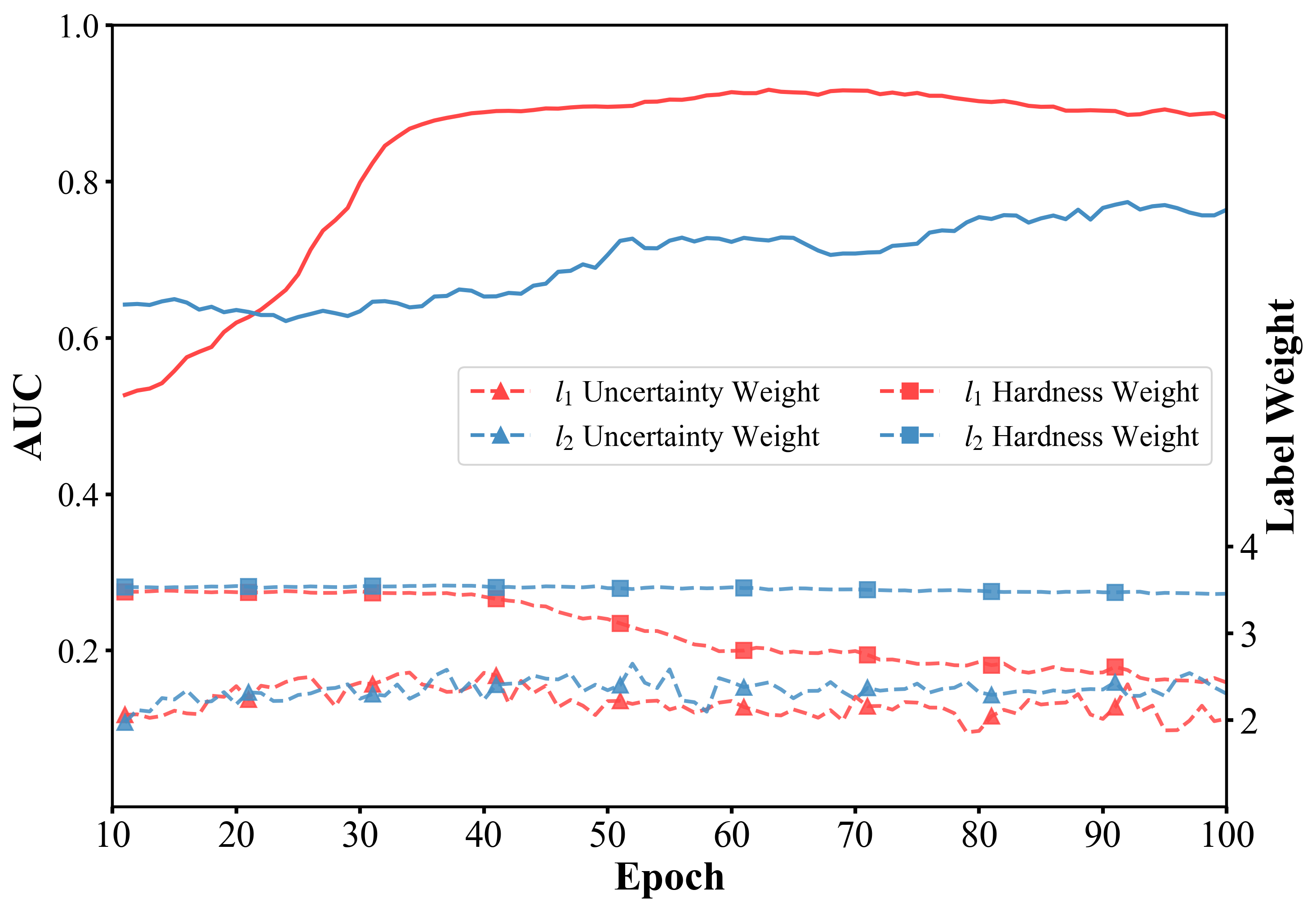}
  \label{fig:labelweight_auc_c}
  }
  \caption{AUC of two labels from the Corel5k dataset and their corresponding label weights across various epochs given by Balance, Hard-Imb, and D2ACE with CLIF base model.}
  \label{fig:labelweight_auc}
\end{figure*}

Figure~\ref{fig:labelweight_auc} shows the AUC scores of two labels ($l_1$ Sun and $l_2$ Turn) throughout training, alongside the label weights assigned by Balance, Hard-Imb, and D2ACE. Figures~A\ref{fig:labelweight_auc_a} and A\ref{fig:labelweight_auc_b} correspond to Figures~2(a) and (b) in the Introduction Section, respectively.

Unlike Balance and Hard-Imb that use static label priority based on label statistics, D2ACE dynamically adjusts label weights during training. Specifically, the hardness-based weight for $l_1$ decreases as its AUC increases, whereas the hardness-based weight for $l_2$ declines only slightly due to its slower AUC growth. The uncertainty-based weights exhibit larger fluctuations than hardness-based weights, reflecting the more variable nature of prediction probability compared to loss. Notably, in the later stages, the uncertainty weight for $l_1$ becomes lower than that for $l_2$, further illustrating that D2ACE adapts label weights inversely with AUC scores to emphasize less well-learned labels.

\bibliographystyle{named}
\bibliography{ijcai26.bib}